\documentclass[
10pt, 
english, 
onehalfspacing, 
 nolistspacing, 
parskip, 
headsepline, 
]{TUThesis} 

\usepackage[utf8]{inputenc} 
\usepackage[T1]{fontenc}    
\usepackage{hyperref}       
\usepackage{wrapfig} 
\usepackage{textcomp}
\usepackage{mathtools}
\usepackage{cleveref}
\usepackage{siunitx}
\usepackage{pifont}
\usepackage{lipsum}
\SetLipsumParListSurrounders{\begingroup\color{gray}}{\endgroup}	

\usepackage{pgf, tikz} 
\usetikzlibrary{arrows, automata}
\usepackage{tabularx, array}
\usepackage{mathrsfs}  

\usepackage{lmodern}
\usepackage{dsfont}

\usepackage[sortcites,doi=false,isbn=true,url=false, style=numeric-comp, natbib=true,maxbibnames=99,giveninits=true, backend=biber]{biblatex}
\usepackage[autostyle=true]{csquotes} 
\AtBeginEnvironment{enquote}{\itshape} 
\usepackage{rotating} 
\usepackage[normalem]{ulem} 
\usepackage{svg}

\usepackage{xargs}                  
\usepackage[colorinlistoftodos,prependcaption,textsize=tiny]{todonotes}
\newcommandx{\unsure}[2][1=]{\todo[linecolor=red,backgroundcolor=red!25,bordercolor=red,#1]{#2}}
\newcommandx{\change}[2][1=]{\todo[linecolor=blue,backgroundcolor=blue!25,bordercolor=blue,#1]{#2}}
\newcommandx{\info}[2][1=]{\todo[linecolor=orange,backgroundcolor=orange!25,bordercolor=orange,#1]{#2}}
\newcommandx{\improvement}[2][1=]{\todo[linecolor=violet,backgroundcolor=violet!25,bordercolor=violet,#1]{#2}}
\newcommandx{\thiswillnotshow}[2][1=]{\todo[disable,#1]{#2}}
\newcommandx{\draft}[2][1=]{\todo[inline,#1]{#2}}

\newcolumntype{P}[1]{>{\centering\arraybackslash}p{#1}}
\newcolumntype{M}[1]{>{\centering\arraybackslash}m{#1}}

\newcolumntype{Y}{>{\centering\arraybackslash}X}
\newcolumntype{C}{>{\centering\let\newline\\\arraybackslash\hspace{0pt}}c}

\usepackage{makecell}

\thesistitle{Explainability-aided Domain Generalization for Image Classification}

\supervisor{Dr. Massimiliano \textsc{Mancini}}

\cosupervisor{Prof. Dr. Zeynep \textsc{Akata}}

\reviewer{Prof. Dr. Philipp \textsc{Hennig}}

\examiner{}

\degree{Master of Science (M.Sc.)}

\author{Robin M. \textsc{Schmidt}}

\addresses{}

\subject{}


\university{University of Tübingen}

\bsuniversity{Cooperative State University Stuttgart}
\msuniversity{University of Tübingen}

\department{Department of Computer Science}

\group{Explainable Machine Learning Group}

\faculty{}

\AtBeginDocument{
\hypersetup{pdftitle=\ttitle} 
\hypersetup{pdfauthor=\authorname} 
\hypersetup{pdfkeywords=\keywordnames} 
}
\newcommand*{\eg}{e.g.\@\xspace}
\newcommand*{\ie}{i.e.\@\xspace}

\newcommand{\adam}{\textsc{Adam}\xspace}

\newcommand{\sgd}{\textsc{SGD}\xspace}

\DeclarePairedDelimiterX{\norm}[1]{\lVert}{\rVert}{#1}
\newcommand{\plus}{\raisebox{.4\height}{\scalebox{.6}{+}}}
\newcommand{\minus}{\raisebox{.4\height}{\scalebox{.8}{-}}}
\newcommand{\imagequadsize}{1cm}
\newcommand{\imagequadsizecams}{2cm}
\newcommand{\domainsize}[1]{\scriptsize{#1}}
\newcommand{\domainbed}{\textsc{DomainBed}\xspace}
\newcommand{\lime}{\textsc{LIME}}

\newcommand{\xx}{\mathbf{x}}
\newcommand{\yy}{\mathbf{y}}
\newcommand{\zz}{\mathbf{z}}
\newcommand{\ypred}{\hat{\yy}}
\newcommand{\ys}{\mathcal{Y}}
\newcommand{\xs}{\mathcal{X}}
\newcommand{\zs}{\mathcal{Z}}
\newcommand{\data}{D}
\newcommand{\dist}{\mathcal{D}}
\newcommand{\xrandom}{X}
\newcommand{\yrandom}{Y}
\newcommand{\p}{\boldsymbol{\theta}}
\newcommand{\pg}{\boldsymbol{\theta}^\prime}
\newcommand{\ps}{\Theta}
\newcommand{\modelf}{f_{\p}}
\newcommand{\modelg}{g_{\pg}}
\newcommand{\gs}{\mathcal{G}}
\newcommand{\featureex}{\phi}
\newcommand{\classifier}{w}
\newcommand{\f}[1]{\modelf(#1)}
\newcommand{\risk}{R(\modelf)}
\newcommand{\riske}{R_{\mathrm{erm}}(\modelf)}
\newcommand{\xxi}{\xx_i}
\newcommand{\xxic}{\xx_{i,c}}
\newcommand{\xxq}{\xx_{q}}
\newcommand{\yyi}{\yy_i}
\newcommand{\lterm}{\mathcal{L}}
\newcommand{\lossi}[2]{\lterm\left(#1, #2\right)}
\newcommand{\sample}{(\xxi, \yyi)}
\newcommand{\yoh}{\dot{\yy}}
\newcommand{\yioh}{\dot{\yy}_i}
\newcommand{\envs}{\Xi}
\newcommand{\env}{\xi}
\newcommand{\tenvs}{\Phi}
\newcommand{\meta}{\mathfrak{D}}
\newcommand{\com}{\Pi}
\newcommand{\compl}[1]{\com(#1)}
\newcommand{\proti}{\mathbf{p}}
\newcommand{\prot}{\proti_{j}}
\newcommand{\protc}{\proti_{j,c}}
\newcommand{\prots}{\mathcal{P}}
\newcommand{\unit}{g_{\prot}}

\newcommand{\player}{g_\proti}
\newcommand{\mplayer}{\Tilde{g}_\proti}
\newcommand{\simmap}{\Psi_j}
\newcommand{\zpatch}{\Bar{\zz}}
\newcommand{\stability}{\epsilon}
\newcommand{\cdistance}{\varrho}

\newcommand{\featureg}{\mathbf{g}_\mathbf{z}}
\newcommand{\featuregl}{\mathbf{g}_{\mathbf{z},i,j}}
\newcommand{\featurega}{\Tilde{\mathbf{g}}_\mathbf{z}}
\newcommand{\featuregal}{\Tilde{\mathbf{g}}_{\mathbf{z}, i, j}}

\newcommand{\loguni}[2]{\mathcal{L}\mathcal{U}_{10}(#1, #2)}
\newcommand{\logunitwo}[2]{\mathcal{L}\mathcal{U}_{2}(#1, #2)}
\newcommand{\logunix}[2]{\mathcal{L}\mathcal{U}_{x}(#1, #2)}
\newcommand{\uni}[2]{\mathcal{U}(#1, #2)}

\newcommand{\rsc}{\textsc{RSC}\xspace}

\newcommand{\dtransformers}{\textsc{D-Transformers}\xspace}
\newcommand{\tdtransformers}{\textsc{D-Transformers}*\xspace}

\newcommand{\divcam}{\textsc{DivCAM}\xspace}
\newcommand{\divcams}{\textsc{\divcam-S}\xspace}
\newcommand{\divcamc}{\textsc{\divcam-C}\xspace}
\newcommand{\divcamt}{\textsc{\divcam-T}\xspace}
\newcommand{\divcamd}{\textsc{\divcam-D}\xspace}
\newcommand{\divcamds}{\textsc{\divcam-DS}\xspace}
\newcommand{\divcamdc}{\textsc{\divcam-DC}\xspace}
\newcommand{\divcamdcs}{\textsc{\divcam-DCS}\xspace}
\newcommand{\divcamdt}{\textsc{\divcam-DT}\xspace}
\newcommand{\divcamdts}{\textsc{\divcam-DTS}\xspace}
\newcommand{\divcamcs}{\textsc{\divcam-CS}\xspace}
\newcommand{\divcamts}{\textsc{\divcam-TS}\xspace}
\newcommand{\tdivcam}{\textsc{DivCAM}*\xspace}
\newcommand{\tdivcams}{\textsc{\divcam-S}*\xspace}
\newcommand{\tdivcamc}{\textsc{\divcam-C}*\xspace}
\newcommand{\tdivcamt}{\textsc{\divcam-T}*\xspace}
\newcommand{\tdivcamd}{\textsc{\divcam-D}*\xspace}
\newcommand{\tdivcamds}{\textsc{\divcam-DS}*\xspace}
\newcommand{\tdivcamdc}{\textsc{\divcam-DC}*\xspace}
\newcommand{\tdivcamdcs}{\textsc{\divcam-DCS}*\xspace}
\newcommand{\tdivcamdt}{\textsc{\divcam-DT}*\xspace}
\newcommand{\tdivcamdts}{\textsc{\divcam-DTS}*\xspace}
\newcommand{\tdivcamcs}{\textsc{\divcam-CS}*\xspace}
\newcommand{\tdivcamts}{\textsc{\divcam-TS}*\xspace}

\newcommand{\prodrop}{\textsc{ProDrop}\xspace}

\newcommand{\tabtop}[1]{\textcolor{darkred}{\mathbf{#1}}}

\DeclarePairedDelimiter\floor{\lfloor}{\rfloor}

\newcommand{\support}{S}
\newcommand{\supportc}{\support_c}
\newcommand{\keyh}{\Gamma}
\newcommand{\valueh}{\Lambda}
\newcommand{\queryh}{\Omega}
\newcommand{\dimkey}{d_{k}}
\newcommand{\dimvalue}{d_{v}}
\newcommand{\keys}{\mathbf{k}_{i,m,c}^\env}
\newcommand{\values}{\mathbf{v}_{i,m,c}^\env}
\newcommand{\valuesq}{\mathbf{w}_{p}^\env}

\newcommand{\queries}{\mathbf{q}^\env_{p}}
\newcommand{\protipc}{\proti_{p, c}^\env}

\newcommand{\attentionw}{\alpha_{i,m,c,p}^\env}
\newcommand{\attentionwf}{\Tilde{\alpha}_{i,m,c,p}^\env}
\newcommand{\scaler}{\tau}

\begin{document}

\frontmatter 
\pagenumbering{Roman} 
\pagestyle{plain} 


\begin{titlepage}
\addtocounter{page}{-1}
\begin{center}


\vspace{0.5cm}
\textsc{\Large Master Thesis}\\[0.5cm] 

\rule[0.4cm]{16cm}{0.1pt}\\
{\huge \bfseries \ttitle\par}\vspace{0.4cm} 
\rule{16cm}{0.1pt}\\ \vspace{1.5cm}
 
\begin{minipage}[t]{0.5\textwidth}
\begin{flushleft} \large
\emph{Author:}\\
\href{https://scholar.google.de/citations?user=20vb63kAAAAJ&hl=de}{\authorname} 
\end{flushleft}
\end{minipage}
\begin{minipage}[t]{0.4\textwidth}
\begin{flushright} \large
\emph{Supervisor:} \\
\href{https://scholar.google.it/citations?user=bqTPA8kAAAAJ&hl=en}{\supname} \\[0.3cm]
\emph{Co-Supervisor:} \\
\href{https://scholar.google.com/citations?user=jQl9RtkAAAAJ&hl=de}{\cosupname} \\[0.3cm]
\emph{Reviewer:} \\
\href{https://scholar.google.de/citations?user=UeG5w08AAAAJ&hl=en}{\reviewname}
\end{flushright}
\end{minipage}\\[1.5cm]

\includegraphics[width= 230px]{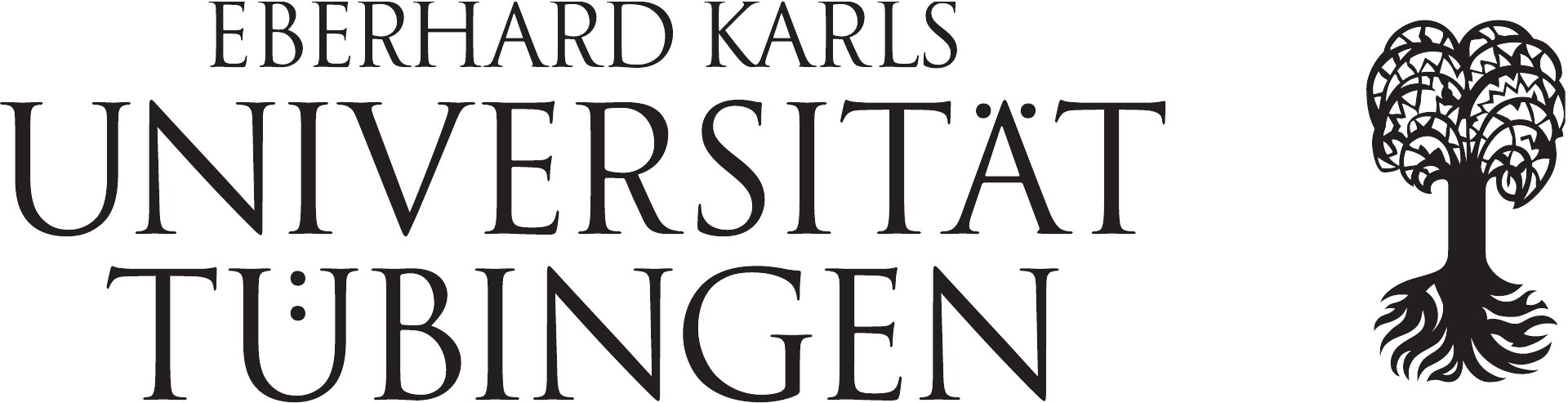} 
 \vspace{1.2cm}
\vfill

\large \textit{A thesis submitted in fulfillment of the requirements\\ for the degree of \degreename{} in Computer Science}\\[0.3cm] 
\textit{in the}\\[0.4cm]
\href{https://uni-tuebingen.de/en/faculties/faculty-of-science/departments/computer-science/department/}{\deptname} \\ \href{https://eml-unitue.de}{\groupname}\\[1cm] 
 
\vfill

{\large \usdate\today}\\

\vfill
\end{center}

\end{titlepage}

\blankpage
\addtocounter{page}{-1}
\newgeometry{top=2.5cm,bottom=2.5cm,right=2.5cm,left=2.5cm, bindingoffset=.5cm,}


\begin{abstract}

\noindent Traditionally, for most machine learning settings, gaining some degree of explainability that tries to give users more insights into \emph{how} and \emph{why} the network arrives at its predictions, restricts the underlying model and hinders performance to a certain degree. For example, decision trees are thought of as being more explainable than deep neural networks but they lack performance on visual tasks. In this work, we empirically demonstrate that applying methods and architectures from the explainability literature can, in fact, achieve state-of-the-art performance for the challenging task of domain generalization while offering a framework for more insights into the prediction and training process. For that, we develop a set of novel algorithms including \divcam, an approach where the network receives guidance during training via gradient based class activation maps to focus on a diverse set of discriminative features, as well as \prodrop and \dtransformers which apply prototypical networks to the domain generalization task, either with self-challenging or attention alignment. Since these methods offer competitive performance on top of explainability, we argue that the proposed methods can be used as a tool to improve the robustness of deep neural network architectures. 

\end{abstract}

{
\let\cleardoublepage\clearpage
\newpage
\thispagestyle{empty}
\addtocounter{page}{-1}
\vspace*{\fill}
\scshape \noindent Copyright \copyright 2021, by \authorname \\
\noindent all rights reserved.
\vspace*{\fill}
\newpage
\rm
\begin{declaration}
\setcounter{page}{1}

\vspace{0.6cm}
\noindent I, \authorname, declare that this thesis titled \enquote{\ttitle} which is submitted in fulfillment of the requirements for the degree of Master of Science in Computer Science represents my own work except where acknowledgements have been made. I further declare that this work has not been previously included as a whole or in parts in a thesis, dissertation, or report submitted to this university or to any other institution for a degree, diploma or other qualifications.

\vspace{2cm} 
\begin{flushright}
\hfill Signed: \underline{\hspace{5cm}}\\[2em] 
\hfill Date: \hspace{0.12cm} \underline{\hspace{1.34cm} \usdate\today \hspace{1.34cm}}\\ 
\end{flushright}

\begin{tikzpicture}[remember picture,overlay]
\node[xshift=-5cm,yshift=-17.3cm] at (current page.north east){%
\includegraphics[width=2.2cm]{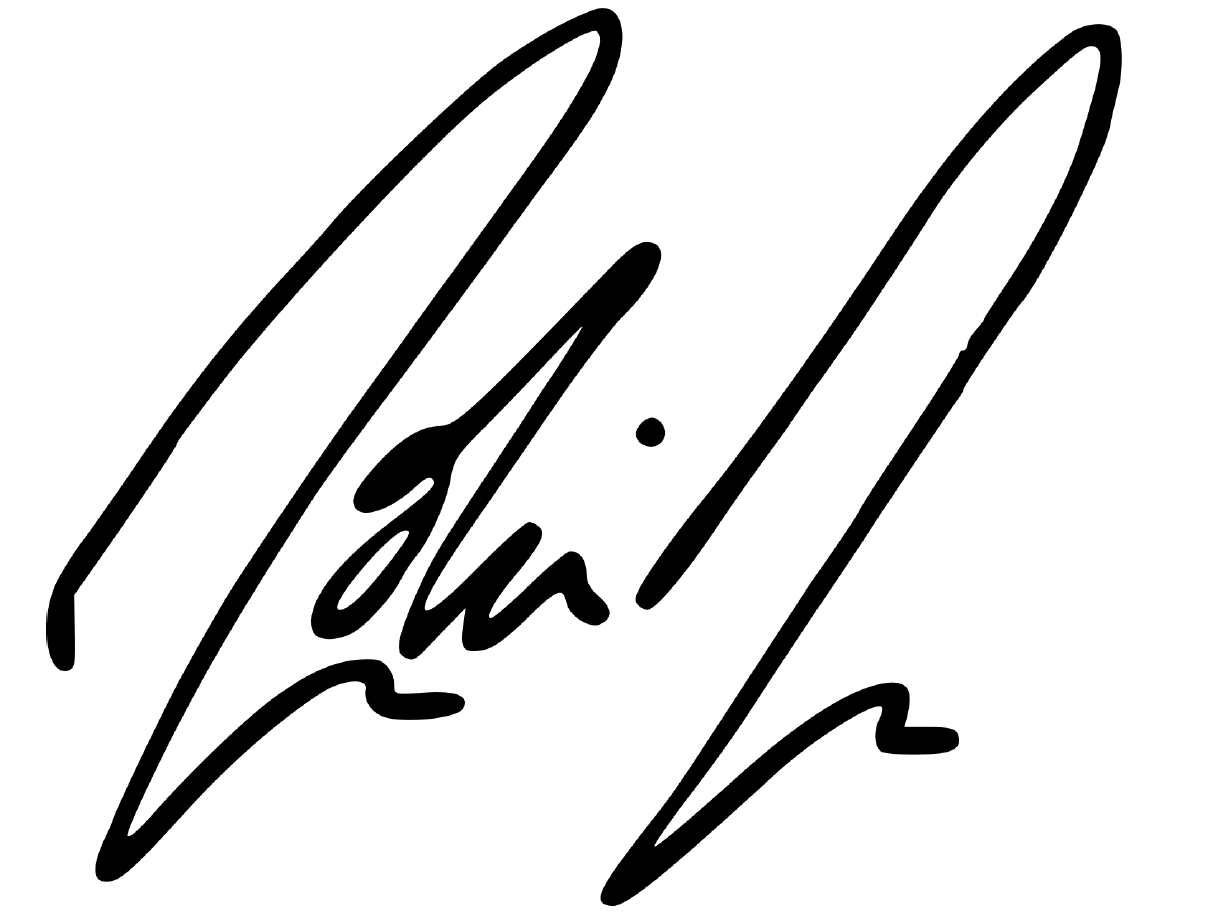}};
\end{tikzpicture}
\end{declaration}

{
  \hypersetup{linkcolor=black}
  \tableofcontents
  \listoffigures 
  \listoftables
  \listofalgorithms
}
\begin{abbreviations}{ll} 
\textbf{ADAM} & \textbf{ADA}ptive \textbf{M}oment Estimation \\
\textbf{AUC} & \textbf{A}rea \textbf{U}nder the \textbf{C}urve \\
\textbf{CAM} & \textbf{C}lass \textbf{A}ctivation \textbf{M}aps \\
\textbf{CDANN} & \textbf{C}onditional \textbf{D}omain \textbf{A}dversarial \textbf{N}eural \textbf{N}etwork \\
\textbf{CE} & \textbf{C}ross-\textbf{E}ntropy \\
\textbf{CNN} & \textbf{C}onvolutional \textbf{N}eural \textbf{N}etwork \\
\textbf{CMNIST} & \textbf{C}olored \textbf{MNIST}  \\ 
\textbf{DA} & \textbf{D}omain \textbf{A}daptation \\
\textbf{DANN} & \textbf{D}omain \textbf{A}dversarial \textbf{N}eural \textbf{N}etwork \\
\textbf{DeepLIFT} & \textbf{Deep} \textbf{L}earning \textbf{I}mportant \textbf{F}ea\textbf{t}ures \\
\textbf{DG} & \textbf{D}omain \textbf{G}eneralization \\
\textbf{DivCAM} & \textbf{Div}ersified \textbf{C}lass \textbf{A}ctivation \textbf{M}aps \\
\textbf{DNN} & \textbf{D}eep \textbf{N}eural \textbf{N}etwork \\
\textbf{ERM} & \textbf{E}mpirical \textbf{R}isk \textbf{M}inimization \\
\textbf{FC} & \textbf{F}ully \textbf{C}onnected Layer \\
\textbf{GAN} & \textbf{G}enerative \textbf{A}dversarial \textbf{N}etwork \\
\textbf{GAP} & \textbf{G}lobal \textbf{A}verage \textbf{P}ooling \\
\textbf{Grad-CAM} &  \textbf{Grad}ient-weighted \textbf{C}lass \textbf{A}ctivation \textbf{M}aps \\
\textbf{HNC} &  \textbf{H}omogeneous \textbf{N}egative \textbf{C}lass Activation Maps \\
\textbf{I.I.D.} & \textbf{I}ndependent and \textbf{I}dentically \textbf{D}istributed \\
\textbf{KL} & \textbf{K}ullback–\textbf{L}eibler \\
\textbf{LIME} & \textbf{L}ocal \textbf{I}nterpretable \textbf{M}odel-agnostic \textbf{E}xplanations \\
\textbf{MAML} & \textbf{M}odel-\textbf{A}gnostic \textbf{M}eta-\textbf{L}earning \\
\textbf{MMD} & \textbf{M}aximum \textbf{M}ean \textbf{D}iscrepancy \\
\textbf{MTAE} & \textbf{M}ulti-\textbf{T}ask \textbf{A}uto\textbf{e}ncoder \\
\textbf{NLP} & \textbf{N}atural \textbf{L}anguage \textbf{P}rocessing \\
\textbf{ProDrop} & \textbf{Pro}totype \textbf{Drop}ping \\
\textbf{ReLU} & \textbf{R}ectified \textbf{L}inear \textbf{U}nit \\
\textbf{RSC} & \textbf{R}epresentation \textbf{S}elf-\textbf{C}hallenging \\
\textbf{RKHS} & \textbf{R}eproducing \textbf{K}ernel
\textbf{H}ilbert \textbf{S}pace \\
\textbf{RMNIST} & \textbf{R}otated \textbf{MNIST} \\
\textbf{SGD} & \textbf{S}tochastic \textbf{G}radient \textbf{D}escent \\
\textbf{SVM} & \textbf{S}upport \textbf{V}ector \textbf{M}achine \\
\textbf{TAP} & \textbf{T}hreshold \textbf{A}verage \textbf{P}ooling \\
\textbf{UML} & \textbf{U}nbiased \textbf{M}etric \textbf{L}earning \\

\end{abbreviations}
\begin{symbols}{p{0.15\textwidth}p{0.7\textwidth}l} 

\multicolumn{3}{l}{\symboltitle{Chapter~2}}\\ \\
$\xxi$ & instance of input features & --- \\
$\yyi$ & corresponding label for input features & --- \\
$\yioh$ & instance of label one-hot encoding & --- \\
$\sample$ & sample of input and label pair & --- \\
$\ypred$ & predicted output label & --- \\
$\xrandom$ & random variable for input features & --- \\
$\yrandom$ & random variable for output labels & --- \\
$C$ & number of classes & --- \\
$\data$ & training dataset & --- \\
$\dist$ & training distribution & --- \\
$\zz$ & feature representation & --- \\
$\Tilde{\mathbf{z}}$ & masked features & --- \\
$\xs$ & input space & --- \\
$\ys$ & output space & --- \\
$\zs$ & latent space & --- \\
$\ps$ & parameter space & --- \\
$\p$ & model parameters & --- \\
$\modelf$ & model predictor & --- \\
$\featureex$ & feature extractor & --- \\
$\classifier$ & classifier & --- \\
$K$ & number of feature maps of the last convolutional layer & --- \\
$\risk$ & model risk & --- \\
$\riske$ & empirical model risk & --- \\
$\lterm$ & loss term & --- \\
$\envs$ & set of source environments & --- \\
$\tenvs$ & set of test environments & --- \\
$\env$ & environment & --- \\
$\meta$ & meta distribution & --- \\
$U$ & unlabeled dataset & --- \\
$L$ & labeled dataset & --- \\
$\mathcal{H}$ & reproducing kernel Hilbert space & --- \\
$\varphi$ & feature map induced by a kernel & --- \\
$\Delta_\env$ & local envinronment bias & --- \\
$\p_\env$ & environment parameters & --- \\
$\mathbf{M}_c$ & class activation map for class $c$ & --- \\
$\featureg$ & gradient with respect to the features & --- \\
$\featurega$ & average pooled gradient values & --- \\
$H_\mathbf{z}$ & feature map height & --- \\
$W_\mathbf{z}$ & feature map width & --- \\
$y_c$ & logit for class $c$ & --- \\
$\mathbf{m}_{i,j}$ & feature mask at spatial location $(i,j)$ & --- \\
$\mathbf{c}$ & change vector after applying the mask & --- \\
$q_p$ & feature percentile threshold & --- \\
$b_p$ & batch percentile threshold & --- \\[1cm]

\multicolumn{3}{l}{\symboltitle{Chapter~3}}\\ \\
$\modelg$ & interpretable model & --- \\
$\gs$ & set of interpretable models & --- \\
$\com$ & complexity measure & --- \\
$\prots$ & set of prototypes & --- \\
$\prot$ & $j$-th prototype & --- \\
$H_\proti$ & height of the prototypes & --- \\
$W_\proti$ & width of the prototypes & --- \\
$\unit$ & prototype unit & --- \\
$\player$ & prototype layer & --- \\
$\zpatch$ & latent patch & --- \\
$\simmap$ & similarity map between $j$-th prototype and latent representation & --- \\
$\stability$ & numerical stability factor & --- \\
$w_{c,j}$ & classifier weight connecting the $j$-th prototype unit and class $c$ logit & --- \\
$\p_\featureex$ & parameters of the feature extractor & --- \\
$\p_\classifier$ & parameters of the classifier & --- \\
$\ell_2$ & euclidean distance & --- \\
$\lambda$ & loss term weighting factor & --- \\\\[1cm]

\multicolumn{3}{l}{\symboltitle{Chapter~4}} \\ \\

$\tau_{t a p}$ & threshold for average pooling & --- \\
$J^{m}_>$ & Set of Top-$m$ negative classes & --- \\ 
$\boldsymbol{U}$ & uniform probability matrix & --- \\
$\boldsymbol{M}_{c}^{\prime}$ & probability map & --- \\
$\omega$ & domain predictor & --- \\
$\mathbf{d}$ & domain ground truth & --- \\
$\eta$ & $\ell^2$ regularization weighting factor & --- \\
$k$ & kernel function & --- \\
$\mathfrak{P}$ & set of source domain pairs & --- \\
$\cdistance$ & cosine distance & --- \\
$\supportc$ & support set for class $c$ & --- \\
$\keyh$ & key head & --- \\
$\valueh$ & value head & --- \\
$\queryh$ & query head & --- \\
$\mathbf{k}$ & keys of the support set & --- \\
$\mathbf{q}$ & queries & --- \\
$\mathbf{v}$ & support-set values & --- \\
$\mathbf{w}$ & query image values & --- \\
$\alpha$ & dot similarity between keys and queries & --- \\
$\Tilde{\alpha}$ & attention weights & --- \\
$\mathcal{B}$ & batch size & --- \\
$\alpha$ & learning rate & --- \\
$\gamma$ & weight decay factor & --- \\

\end{symbols}
}


\mainmatter 

\pagestyle{thesis} 

\setlength{\parindent}{0pt}

\chapter{Introduction} 
\label{sec:Introduction}

Modern machine learning solutions commonly rely on supervised deep neural networks as their default approach and one of the tasks that is commonly required and implemented in practice is \emph{image classification}. In its simplest form, the used networks combine multiple layers of linear transformations coupled with non-linear activation functions to \emph{classify} an input image based on its pixel values into a discrete set of classes. The chosen architecture, also known as the model, is then able to \emph{learn} good parameters  that represent how to combine the information extracted from the individual pixel values using additional labeled information. That is, for a set of images we know the correct class and can automatically guide the network towards well-working parameters by determining how wrong the current prediction is and in which direction we need to update the individual parameters, for our network to give a more accurate class prediction. 

Obtaining this labeled information, however, is very tedious in practice and either requires a lot of manual human labeling or sufficient human quality assurance for any automatic labeling system. A commonly used approach to overcome this impediment is to combine multiple sources of data, that might have been collected in different settings but represent the same set of classes and have already been labeled a priori. Since the training distribution described by the obtained labeled data is then often different from the testing distribution imposed by the images we observe once we deploy our system, we commonly observe a \emph{distribution shift} when testing and the network generally needs to do \emph{out-of-distribution} predictions. Similar behavior can be observed when the model encounters irregular properties during testing such as weather or lighting conditions which have not been captured well by the training data. For many computer vision neural network models, this poses an interesting and challenging task which is known as \emph{out-of-distribution generalization} where researchers try to improve the predictions under these alternating circumstances for more robust machine learning models.

In this work, we pick up on this challenge and try to improve the out-of-distribution generalization capabilities with models and techniques that have been proposed to make deep neural networks more explainable. For most machine learning settings, gaining a degree of explainability that tries to give humans more insights into \emph{how} and \emph{why} the network arrives at its predictions, restricts the underlying model and hinders performance to a certain degree. For example, decision trees are thought of as being more explainable than deep neural networks but lack performance on visual tasks. In this work, we investigate if these properties also hold for the out-of-distribution generalization task or if we can deploy explainability methods during the training procedure and gain, both, better performance as well as a framework that enables more explainability for the users. In particular, we develop a regularization technique based on class activation maps that visualize parts of an image that led to certain predictions (\divcam) as well as prototypical representations that serve as a number of class or attribute centroids which the network uses to make its predictions (\prodrop and \dtransformers). Specifically, we deploy these methods for the \emph{domain generalization} task where the model has access to images from multiple training domains, each imposing a different distribution, but without access to images from the immediate testing distribution.

From our experiments, we observe that especially \divcam offers state-of-the-art performance on some datasets while providing a framework that enables additional insights into the training and prediction procedure. For us, this is a property that is highly desirable, especially in safety-critical scenarios such as self-driving cars, any application in the medical field such as cancer or tumor prediction, or any other automation robot that needs to operate in a diverse set of environments. Hopefully, some of the methods presented in this work can find application in such scenarios and establish additional trust and confidence into the machine learning system to work reliable. All of our experiments have been conducted within the \domainbed domain generalization benchmarking framework and the respective code has been open-sourced.\footnote{All implementation details can be found here: \url{https://github.com/SirRob1997/DomainBed/}}

\chapter{Domain Generalization} 
\label{DomainGeneralization} 

Machine learning systems often lack \emph{out-of-distribution generalization} which causes models to heavily rely on the training distribution and as a result don't perform very well when presented with a different input distribution during testing. Examples are application scenarios where intelligent systems don't generalize well across health centers if the training data was only collected in a single hospital \citep{Castro_2020, AlBadawy2018, PeroneBBC19} or when self-driving cars struggle under alternative lighting or weather conditions \citep{DaiG18, VolkMBH019}. Properties that are often interpreted falsely as part of the relevant feature set include backgrounds \citep{BeeryHP18}, textures \citep{GeirhosRMBWB19}, or racial biases \citep{StockC18}. Not only can a failure in capturing this domain shift lead to poor performance, but in safety-critical scenarios this can correspond to a large impact on people's lives. Due to the prevalence of this challenge for the wide-spread deployment of machine learning systems in diverse environments, many researchers tried to tackle this task with different approaches. In this chapter, we want to give a broad overview of the literature in \emph{domain generalization} and prepare the fundamentals for the following chapters. If you are already familiar with the field, you can safely skip this chapter and only familiarize yourself with the used notation.

\section{Problem formulation}
\label{sec:domain_gen_problem}

\paragraph{Supervised Learning}
In supervised learning we are aiming to optimize the predictions $\ypred$ for the values $\yy \in \ys$ of a random variable $\yrandom$ when presented with values $\xx \in \xs$ of a random variable $\xrandom$. These predictions are generated with a model predictor $\f{\cdot}: \xs \rightarrow \ys$ that is parameterized by $\p \in \ps$, usually the weights of a neural network,  and is assigning the predictions as $\ypred=\f{\cdot}$. To improve our predictions, we utilize a training dataset containing $n$ input-output pairs denoted as $\data=\left\{\left(\xxi, \yyi \right)\right\}_{i=1}^{n}$ where each sample $\sample$ is ideally drawn identically and independently distributed (i.i.d.) from a single joint probability distribution $\dist$. By using a loss term $\lossi{\ypred}{\yy}: \ys \times \ys \rightarrow \mathbb{R}^{+}$, which quantifies how different the prediction $\ypred$ is from the ground truth $\yy$, we would like to minimize the risk,
\begin{equation}
    \risk = \mathbb{E}_{\sample \sim \dist}\left[\lossi{\f{\xxi}}{\yyi}\right],
\end{equation}
of our model. Since we only have access to the distribution $\dist$ through a proxy in the form of the dataset $\data$, we are instead using Empirical Risk Minimization (ERM):
\begin{equation}
    \riske =\frac{1}{n} \sum_{i=1}^n \lossi{\f{\xxi}}{\yyi},
\end{equation}
by adding up the loss terms of each sample. One common choice for this loss term is the Cross-Entropy (CE) loss which is shown in \Cref{eq:cross_entropy}.
\begin{equation}
\label{eq:cross_entropy}
    \lterm_{\mathrm{ce}}(\ypred_i, \yioh)=-\sum_{c=1}^{C} y_{i, c} \cdot \log \left(\hat{y}_{i, c}\right)
\end{equation}
Here, $\yioh$ is the one-hot vector representing the ground truth class, $\ypred_i$ is the \emph{softmax} output of the model, $\hat{y}_{i, c}$ and $y_{i, c}$ are the $c$-th dimension of $\ypred_i$ and $\yioh$ respectively.

The occurring minimization problem is then often solved through iterative gradient-based optimization algorithms \eg \sgd{} \citep{Robbins1951} or \adam \citep{Kingma2015} which perform on-par with recent methods for the non-convex, continuous loss surfaces produced by modern machine learning problems and architectures \citep{schmidt2020descending}.

On top of that, the model predictor $\modelf$ can be decomposed into two functions as $\modelf = \classifier \circ \featureex$ where $\featureex: \xs \rightarrow \zs$ is an embedding into a feature space, hence sometimes called the feature extractor, and $\classifier: \zs \rightarrow \ys$ which is called the classifier since it is a prediction from the feature space to the output space \citep{gulrajani2020search, MotiianPAD17}. This often allows for a more concise mathematical notation.

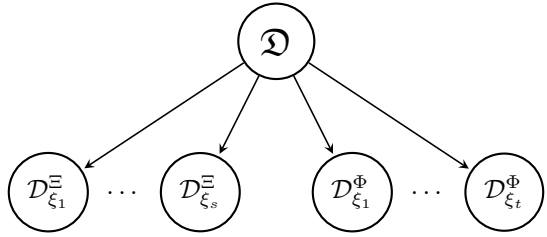
\begin{figure}[htbp]
    \centering
    \begin{tikzpicture}[
            > = stealth, 
            shorten > = 1pt, 
            auto,
            node distance = 3cm, 
            semithick 
        ]
    \tikzstyle{every state}=[
            draw = black,
            thick,
            fill = white,
            minimum size = 1cm
        ]

        \node[state] (meta) at (0,0) {\Large{$\meta$}};
        \node[state] (d1) at (-3,-2) {$\dist_{\env_1}^\envs$};
        \node[state] (d2) at (-1,-2) {$\dist_{\env_s}^\envs$};
        \node[state] (d3) at (1,-2) {$\dist_{\env_1}^\tenvs$};
        \node[state] (d4) at (3,-2) {$\dist_{\env_t}^\tenvs$};

        \path[->] (meta) edge node {} (d1);
        \path[->] (meta) edge node {} (d2);
        \path[->] (meta) edge node {} (d3);
        \path[->] (meta) edge node {} (d4);
        \path (d1) -- node[auto=false]{\ldots} (d2);
        \path (d3) -- node[auto=false]{\ldots} (d4);
    \end{tikzpicture}
    \caption[Meta-distribution $\meta$ generating source and unseen domain distributions]{Meta-distribution $\meta$ generating source environment distributions (left) and unseen environment distributions (right), adapted from: \citep{albuquerque2019generalizing}}
    \label{fig:meta_domain}
\end{figure}

\paragraph{Domain Generalization}
The problem of \emph{Domain generalization} (DG) builds on top of this framework, where now we have a set of training environments $\envs = \left\{\env_{1}, \ldots, \env_{s} \right\}$, also known as source domains, where each environment $\env$ has an associated dataset $\data_{\env}=\left\{\left(\xxi^{\env}, \yyi^{\env}\right)\right\}_{i=1}^{n_{\env}}$ containing $n_\env$ i.i.d. samples from individual data distributions $\dist_\env$. Note that, while related, the environments have different joint distributions, \ie $\dist_{\env_i}\neq \dist_{\env_j} \;\forall i\neq j$. Here, $\xxi^\env \in \mathbb{R}^{m}$ is the $i$-th sample for environment $\env$ representing an $m$-dimensional feature vector (\ie an image in our case) and $\yyi^\env \in \ys$ is the corresponding ground truth class label over the $C$ possible classes. The one-hot vector representing the ground truth is denoted as $\yioh^\env$. To clear up some notation, we sometimes omit $\env$ where it is obvious. From these source domains, we try to learn generic feature representations agnostic to domain changes to improve model performance \citep{seo2019learning}. Simply, we try to do \emph{out-of-distribution generalization} where our model aims to achieve good performance for an unseen test environment $\env_t$ sampled from the set of unseen environments $\tenvs = \left\{\env_{1}, \ldots, \env_{t}\right\}$ with $\envs \cap \tenvs =\emptyset$ based on statistical invariances across the observed training (source) and testing (target) domains \citep{gulrajani2020search, huang2020selfchallenging}. For that, we try to minimize the expected target risk of our model as:
\begin{equation}
\label{eq:domain_risk}
    \risk = \mathbb{E}_{(\xxi^{\env_t}, \yyi^{\env_t}) \sim \dist_{\env_t}}\left[\lossi{\f{\xxi^{\env_t}}} {\yyi^{\env_t}}\right].
\end{equation}
Since we don't have access to $\dist_{\env_{t}}$ during training, one simple approach is to assume that minimizing the risk over all source domains in $\envs$ achieves good generalization to the target domain. That is, we disregard the separated environments:
\begin{equation}
\label{eq:erm}
    \risk = \mathbb{E}_{(\xxi, \yyi) \sim \cup_{\env \in \envs} \dist_\env} \left[\lossi{\f{\xxi}} {\yyi} \right].
\end{equation}
Again, this can be written with the empirical risk as a simple sum over all the environments and their corresponding samples:
\begin{equation}
\label{eq:domain_risk_emp}
    \riske = \frac{1}{s} \sum_{\env \in \envs} \frac{1}{n_\env} \sum_{i=1}^{n_\env} \lossi{\f{\xxi^{\env}}} {\yyi^{\env}}.
\end{equation}
The difference of this approach when compared to ordinary supervised learning is shown on a high-level in \Cref{tab:learning_setups}. It may also be helpful to think about a meta-distribution $\meta$ (real-world) generating source environment distributions $\dist_\env^\envs$ and unseen testing domain distributions $\mathcal{D}_\env^\tenvs$ as shown in \Cref{fig:meta_domain}.

\begin{table}[t]
    \centering
    \begin{tabular}{lll}
    \toprule
    \textbf{Setup} & \textbf{Training inputs}  & \textbf{Testing inputs} \\
    \midrule
        Generative learning & $U_{\env_1}$ & $\emptyset$ \\ 
        Unsupervised learning & $U_{\env_1}$ & $U_{\env_1}$  \\ 
        Supervised learning & $L_{\env_1}$ & $U_{\env_1}$ \\ 
        Semi-supervised learning & $L_{\env_1}, U_{\env_1}$ & $U_{\env_1}$ \\ 
        Multitask learning & $L_{\env_1}, \ldots, L_{\env_s}$ & $U_{\env_1}, \ldots, U_{\env_s}$ \\ 
        Continual (lifelong) learning & $L_{\env_1}, \ldots, L_{\env_\infty}$ & $U_{\env_1}, \ldots, U_{\env_\infty}$ \\ 
        Domain Adaptation & $L_{\env_1}, \ldots, L_{\env_s}, U_{\env_t}$ & $U_{\env_t}$ \\ 
        Transfer learning & $U_{\env_1}, \ldots, U_{\env_s}, L_{\env_t}$ & $U_{\env_t}$ \\ 
        \emph{Domain generalization} & $L_{\env_1}, \ldots, L_{\env_s}$ & $U_{\env_t}$ \\ 
    \bottomrule
    \end{tabular}
    \caption[Differences in learning setups]{Differences in learning setups adapted from: \citep{gulrajani2020search}. For each environment $\env$ the labeled and unlabeled datasets are denoted as $L_\env$ and $U_\env$ respectively.}
    \label{tab:learning_setups}
\end{table}

\paragraph{Homogeneous and Heterogeneous}
Sometimes, domain generalization is also divided into \emph{homogeneous} and \emph{heterogeneous} subtasks. In homogeneous DG we assume that all domains share the same label space $\ys^{\env_i} = \ys^{\env_j} = \ys^{\env_t}$, $\forall \env_i \neq \env_j \in \envs$. On the contrary, the more challenging heterogeneous DG allows for different label spaces $\ys^{\env_i} \neq \ys^{\env_j} \neq \ys^{\env_t}$, $\forall \env_i \neq \env_j \in \envs$ which can even be completely disjoint \citep{LiZYLSH19}. For this work, we assume a homogeneous setting.

\paragraph{Single- and Multi-Source}
There exist subtle differences within this task which are called \emph{single-} or \emph{multi-source domain generalization}. While multi-source domain generalization refers to the standard-setting we have just outlined, single-source domain generalization is a more generic formulation \citep{zunino2020explainable}. Instead of relying on multiple training domains to learn models which generalize better, single-source domain generalization aims at learning these representations with access to only one source distribution. Hence, our training domains are restricted to $\envs = \left\{\env_1\right\}$, described by one dataset $\data_{\env_1}=\left\{\left(\xxi^{\env_1}, \yyi^{\env_1}\right)\right\}_{i=1}^{n_{\env_1}}$ and modeling a single source distribution $\dist_{\env_1}$. For example, this can be achieved by combining the different datasets or distributions similar to \Cref{eq:erm} or mathematically $\cup_{\env \in \envs} \data_\env$. This is different from the ordinary supervised learning setup since we want to analyze the performance of the model under a clear domain-shift (\ie out-of-distribution generalization). Keep in mind, that strong regularization methods will also perform well on this subtask. These cross-over and related techniques are described in the following section.

\section{Related concepts and their differences}

As already introduced, members of the causality community might know the task of domain generalization under the term \emph{learning from multiple environments} \citep{arjovsky2019invariant, gulrajani2020search, PetBuhMei15} and researchers coming from deep learning might know it under \emph{learning from multiple domains}. While these two concepts refer to the same task, there exist quite a few related techniques which we want to highlight here and distinguish in their scope. In particular, we focus on ``Generic neural network regularization'' and ``Domain Adaptation'' since each of these are very closely related to domain generalization and sometimes hard to distinguish, if at all. The overview in \Cref{tab:learning_setups}, however, includes even more learning setups to properly position this concept into the machine learning landscape.

\subsection{Generic Neural Network Regularization}

In theory, generic model regularization which aims to prevent neural networks from overfitting on the source domain could also improve the domain generalization performance \citep{huang2020selfchallenging}. As such, methods like dropout \citep{SrivastavaHKSS14}, early stopping \citep{CaruanaLG00}, or weight decay \citep{NowlanH92} can have a positive effect on this task when deployed properly. Apart from regular dropout, where we randomly disable neurons in the training phase to stop them from co-adapting too much, a few alternative methods exist. These include dropping random  patches of input images (Cutout \& HaS) \citep{devries2017improved, SinghL17} or channels of the feature map (SpatialDropout) \citep{TompsonGJLB15}, dropping contiguous regions of the feature maps (DropBlock) \citep{GhiasiLL18}, dropping features of high activations across feature maps and channels (MaxDrop) \citep{ParkK16}, or generalizing the traditional dropout of single units to entire layers during training (DropPath) \citep{LarssonMS17}. There even exist methods like curriculum dropout \citep{MorerioCVVM17} that deploy scheduling for the dropout probability and therefore softly increase the number of units to be suppressed layerwise during training. 

Generally, deploying some of these methods when aiming for out-of-distribution generalization can be a good idea and should definitely be considered for the task of domain generalization.

\subsection{Domain Adaptation}

\emph{Domain Adaptation} (DA) is often mentioned as a closely related task in domain generalization literature \citep{MotiianPAD17, VolpiM19, QiaoZP20}. When compared, domain adaptation has additional access to an unlabeled dataset from the target domain \citep{mancini2020, Csurka17}. Formally, aside from the set of source domains $\envs$ and the domain datasets $\data_\env$, as outlined in \Cref{sec:domain_gen_problem}, we have access to target samples $U_{\env_t} = \left\{\xx_1^{\env_t},\dots,\xx_{n_{\env_t}}^{\env_t}\right\}$ that are from the target domain $\xxi^{\env_t} \sim \dist_{\env_t}$ but their labels remain unknown during training since we want to predict them during testing. As a result, domain generalization is considered to be the harder problem of the two. This difference is also shown in \Cref{tab:learning_setups}.

Earlier methods in this space deploy hand-crafted features to reduce the difference between the source and the target domains \citep{ManciniPBC018}. Like that, \emph{instance-based methods} try to re-weight source samples according to target similarity \citep{GongGS13, HuangSGBS06, YamadaSR12}, or \emph{feature-based methods} try to learn a common subspace \citep{FernandoHST13, GongSSG12, LongD0SGY13, BaktashmotlaghHLS13}. More recent works focus on \emph{deep domain adaptation} based on deep architectures where domain invariant features are learned utilizing supervised neural networks \citep{BousmalisTSKE16, CarlucciPCRB17, GaninL15, GhifaryKZBL16}, autoencoders \citep{ZengOWW14}, or generative adversarial networks (GANs) \citep{BousmalisSDEK17, ShrivastavaPTSW17, TzengHSD17}. These deep NN-based architectures significantly outperform the approaches for hand-crafted features \citep{ManciniPBC018}.

Even though domain adaptation and domain generalization both try to reduce the dataset bias, they are not compatible with each other \citep{GhifaryBKZ17}. Hence, domain adaptation methods often cannot be directly used for domain generalization or vice versa \citep{GhifaryBKZ17}. For this work, we don't rely on the simplifying assumptions of domain adaptation, but instead, tackle the more challenging task of DG.

\section{Previous Works}

Most commonly, work in \emph{domain generalization} can be divided into methods that try to \emph{learn invariant features}, combine domain-specific models in a process called \emph{model ensembling}, pursue \emph{meta-learning}, or utilize \emph{data augmentation} to generate new domains or more robust representations.

Since literature in the domain generalization space is broad, we utilized \citet[Appendix A]{gulrajani2020search} for an overview and to identify relevant literature while individually adding additional works and more detailed information where necessary.  

\subsection{Learning invariant features}
\label{sec:invariant_features}

Methods that try to learn invariant features typically minimize the difference between source domains. They assume that with this approach the features will be domain-invariant and therefore will have good performance for unseen testing domains \citep{huang2020selfchallenging}.

Some of the earliest works on learning invariant features were \emph{kernel methods} applied by \citet{MuandetBS13}  where they experimented with a feature transformation that minimizes the across-domain dissimilarity between transformed feature distributions while preserving the functional relationship between original features and targets. In recent years, there have been approaches following a similar kernel-based approach \citep{LiGTLT18, LiTGLLZT18}, sometimes while maximizing class separability \citep{Hu0CC19, GhifaryBKZ17}. As an early method, \citet{FangXR13} introduce Unbiased Metric Learning (UML) with an SVM metric that enforces the neighborhood of samples to contain samples with the same class label but from other training domains.

After that, \citet{GaninUAGLLML16} introduced Domain Adversarial Neural Networks (DANNs) using neural network architectures to learn domain-invariant feature representations by adding a gradient reversal layer. Recently, their approach got extended to support statistical dependence between domains and class labels \citep{AkuzawaIM19} or considering one-versus-all adversaries to minimize pairwise divergences between source distributions \citep{albuquerque2019generalizing}. \citet{MotiianPAD17} use a siamese architecture to learn a feature transformation that tries to achieve semantical alignment of visual domains while maximally separating them. Other methods are also matching the feature covariance across source domains \citep{RahmanFBS20} or take a causal interpretation to match representations of features \citep{mahajan2020domain}. \citet{huang2020selfchallenging} have also shown that self-challenging (\ie dropping features with high gradient values at each epoch) works very well. 

\citet{MatsuuraH20} use clustering techniques to split single-source domain generalization into different domains and then train a domain-invariant feature extractor via adversarial
learning. Other works have also deployed similar approaches based on adversarial strategies \citep{deng2020representation, Jia_2020_CVPR_SSDG}.

\citet{LiPWK18} deploy adversarial autoencoders with maximum mean discrepancy (MMD) \citep{GrettonBRSS12} to align the source distributions, \ie for distributions $\dist_{\env_1}$, $\dist_{\env_2}$ and a feature map $\varphi: \mathcal{X} \rightarrow \mathcal{H}$ where $\mathcal{H}$ is a reproducing kernel Hilbert space (RKHS) this measure is defined as \Cref{eq:mmd}.
\begin{equation}
\label{eq:mmd}
    \operatorname{MMD}(\dist_{\env_1}, \dist_{\env_2})=\left\|\mathbb{E}_{\sample \sim \dist_{\env_1}}[\varphi(\xxi^{\env_1}))]-\mathbb{E}_{\sample \sim \dist_{\env_2}}[\varphi(\xxi^{\env_2})]\right\|_{\mathcal{H}}
\end{equation}
\citet{ilse2019diva} extend the variational autoencoder \citep{KingmaW13} by introducing latent representations for environments $\zs_\env$, classes $\zs_\yy$, and residual variations $\zs_\xx$. Further, \citet{LiZYLSH19} use episodic training \ie they train a domain agnostic feature extractor $\featureex$ and classifier $\classifier$ by mismatching them with an equivalent trained on a specific domain $\featureex_\env$ and $\classifier_\env$ in combinations $(\featureex_{\env_1}, \classifier_{\env_2}, \xx^{\env_2}_i)$ and $(\featureex_{\env_2}, \classifier_{\env_1}, \xx^{\env_2}_i)$ and letting them predict data outside of the trained domain $\env_1 \neq \env_2$. \citet{piratla2020efficient} also learn domain-specific and common components but the domain-specific parts are discarded after training. \citet{li2020sequential} deploy a lifelong sequential learning strategy.

\subsection{Model ensembling}
\label{sec:model_ensembling}

Some methods try to associate model parameters with each of the training domains and combine them, often together with shared parameters, in a meaningful matter to improve generalization to the test domain. Commonly, the number of models in these type of architectures grow linearly with the number of source domains. 

The first work to pose the problem of domain generalization and analyze it was \citet{BlanchardLS11}. In there, they use classifiers for each sample $\xx^\env$  denoted as $\f{\xx^\env, \mu^\env}$ where $\mu^\env$ corresponds to a kernel mean embedding \citep{MuandetFSS17}. For theoretical analysis on such methods please see \citet{an2019generalization} and \citet{blanchard2017domain}. Later on, \citet{KhoslaZMET12} combine global weights $\p$ with local domain biases $\Delta_\env$ to learn one max-margin linear classifier (SVM) per domain as $\p_\env = \p + \Delta_\env$ and finally combine them, which has recently been extended to neural network settings by adding an additional dimension describing the training domains to the parameter tensors \citep{LiYSH17}. \citet{GhifaryKZB15} propose a Multi-task Autoencoder (MTAE) with shared parameters to the hidden state and domain-specific parameters for each of the training domains. Further, \citet{ManciniBCR18} use domain-specific batch-normalization \citep{IoffeS15} layers and then linearly combine them using a softmax domain classifier. Other works utilize other domain-specific normalization techniques \citep{seo2019learning}, linearly combine domain-specific predictors \citep{ManciniBC018}, or use more elaborate aggregation strategies \citep{DInnocenteC18}. \citet{DingF18} use multiple domain-specific deep neural networks with a structured low-rank constraint and a domain-invariant deep neural network to generalize to the target domain. There have also been works that assign weights to mini-batches depending on their respective error to the training distributions \citep{HuNSS18, sagawa2019distributionally}. \citet{jin2020feature} use attention mechanisms to align the features of the different training domains.

\subsection{Meta-learning}
Meta-learning approaches provide algorithms that tackle the problem of  \emph{learning  to learn} \citep{1998TP, SchmidhuberZW97}.
 As such, \citet{FinnAL17} propose a Model-Agnostic Meta-Learning (MAML) algorithm that can quickly learn new tasks with fine-tuning. \citet{LiYSH18} adapt this algorithm for domain generalization (no fine-tuning) such that we can adapt to new domains by utilizing the meta-optimization objective which ensures that steps to improve training domain performance should also improve testing domain performance. Both approaches are not bound to a specific architecture and can therefore be deployed for a wide variety of learning tasks. These approaches recently got extended by two regularizers that encourage general knowledge about inter-class relationships and domain-independent class-specific cohesion \citep{DouCKG19}, to heterogeneous domain generalization \citep{LiYZH19}, or via meta-learning a regularizer that encourages across-domain performance \citep{BalajiSC18}.

\subsection{Data Augmentation}
Data Augmentation remains a competitive method for generalizing to unseen domains \citep{zhang2019unseen}. Works in this segment try to extend the source environments to a wider range of domains by augmenting the available training environments. However, to deploy an efficient procedure for that, human experts need to consider the data at hand to develop a useful routine \citep{gulrajani2020search}.

Several works have used the \textsc{mixup} \citep{ZhangCDL18} algorithm as a method to merge samples from different domains \citep{XuZNLWTZ20, yan2020improve, WangLK20, mancini2020}. Other works have also tried removing textural information from images \citep{WangHLX19} or shifting it more towards shapes \citep{nam2019reducing, asadi2019shape}. \citet{CarlucciDBCT19} used jigsaw puzzles of image patches as a classification task to show that this improves domain generalization while \citet{VolpiNSDMS18} demonstrate that adversarial data augmentation on a single domain is sufficient. Further, \citet{VolpiM19} use popular image transformations (\eg brightness, contrast, sharpness) with different intensity levels to train a more robust model, or \citet{somavarapu2020frustratingly} use other stylizing techniques. Several methods also use GANs to augment the available training data  \citep{RahmanFBS19, ZhouYHX20, ShankarPCCJS18} or use other methods to generate synthetic domains \citep{zhou2020learning}. \citet{QiaoZP20} deploy an adversarial domain augmentation approach using a Wasserstein Auto-Encoder \citep{TolstikhinBGS18}.

\section{Common Datasets}
There exist several datasets that are commonly used in domain generalization research. Here, we want to introduce the most popular choices as well as interesting datasets to consider. We give an overview over Rotated MNIST, Colored MNIST, Office-Home, VLCS, PACS, Terra Incognita, DomainNet, and ImageNet-C. Currently, the most popular choices include PACS, VLCS, and Office-Home.

\subsection{Rotated MNIST}
The Rotated MNIST (RMNIST) dataset \citep{GhifaryKZB15} is a variation of the original MNIST dataset \citep{lecun-mnisthandwrittendigit-2010} where each digit got rotated by degrees $\{0, 15, 30, 45, 60, 75\}$. Each rotation angle represents one domain as shown in \Cref{tab:common_examples} for classes ``2'' and ``4''. The overall dataset in \citet{gulrajani2020search} includes \num{70000} images from $10$ homogeneous classes ($0-9$) each with dimension $1\times28\times28$.

\begin{table}[htbp]
    \centering
    \begin{tabularx}{\textwidth}{lcYYYYYYYY}
    \toprule
    \textbf{Dataset}  & \textbf{Reference} & $\env_1$   &  $\env_2$ & $\env_3$ & $\env_4$ & $\env_5$ & $\env_6$ & $\dots$ & $\env_{75}$ \\
    \midrule
    \multirow{7}{*}{\textbf{ImageNet-C}}  & \multirow{7}{*}{\textbf{\citep{HendrycksD19}}} & \domainsize{Shot Noise} &     \domainsize{Impulse Noise} & \domainsize{Defocus Blur} & \domainsize{Snow} & \domainsize{Fog} & \domainsize{Brightness}  & &  \domainsize{Pixelate}   \\
    & & \includegraphics[height=\imagequadsize, width=\imagequadsize]{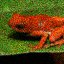} &  \includegraphics[height=\imagequadsize, width=\imagequadsize]{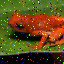} &  \includegraphics[height=\imagequadsize, width=\imagequadsize]{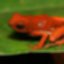} & \includegraphics[height=\imagequadsize, width=\imagequadsize]{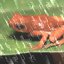} &  \includegraphics[height=\imagequadsize, width=\imagequadsize]{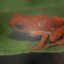} & 
       \includegraphics[height=\imagequadsize, width=\imagequadsize]{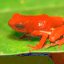} & $\dots$ & \includegraphics[height=\imagequadsize, width=\imagequadsize]{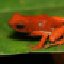} \\
        & & \includegraphics[height=\imagequadsize, width=\imagequadsize]{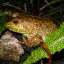} &\includegraphics[height=\imagequadsize, width=\imagequadsize]{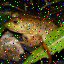} & \includegraphics[height=\imagequadsize, width=\imagequadsize]{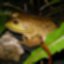} & \includegraphics[height=\imagequadsize, width=\imagequadsize]{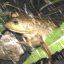} & \includegraphics[height=\imagequadsize, width=\imagequadsize]{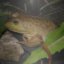} &
       \includegraphics[height=\imagequadsize, width=\imagequadsize]{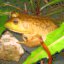} & $\dots$ & \includegraphics[height=\imagequadsize, width=\imagequadsize]{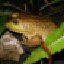} \\
       \addlinespace
       \multirow{7}{*}{\textbf{Rotated MNIST}}  & \multirow{7}{*}{\textbf{\citep{GhifaryKZB15}}} & \domainsize{$0^\circ$} & \domainsize{$15^\circ$} & \domainsize{$30^\circ$} & \domainsize{$45^\circ$} & \domainsize{$60^\circ$} & \domainsize{$75^\circ$} && \\
       & & \includegraphics[height=\imagequadsize, width=\imagequadsize]{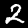} &  \includegraphics[height=\imagequadsize, width=\imagequadsize]{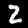} &  \includegraphics[height=\imagequadsize, width=\imagequadsize]{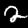} & \includegraphics[height=\imagequadsize, width=\imagequadsize]{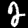} &  \includegraphics[height=\imagequadsize, width=\imagequadsize]{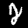} & 
       \includegraphics[height=\imagequadsize, width=\imagequadsize]{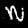} && \\
        & & \includegraphics[height=\imagequadsize, width=\imagequadsize]{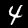} &\includegraphics[height=\imagequadsize, width=\imagequadsize]{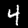} & \includegraphics[height=\imagequadsize, width=\imagequadsize]{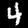} & \includegraphics[height=\imagequadsize, width=\imagequadsize]{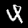} & \includegraphics[height=\imagequadsize, width=\imagequadsize]{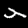} &
       \includegraphics[height=\imagequadsize, width=\imagequadsize]{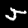} & & \\
       \addlinespace
       \multirow{7}{*}{\textbf{DomainNet}} &  \multirow{7}{*}{\textbf{\citep{PengBXHSW19}}} & \domainsize{Clipart} & \domainsize{Graphic} & \domainsize{Painting} & \domainsize{Draw} & \domainsize{Photo} & \domainsize{Sketch}  & & \\
       & & \includegraphics[height=\imagequadsize, width=\imagequadsize]{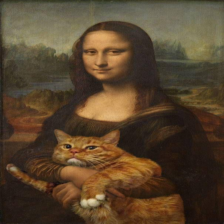} & \includegraphics[height=\imagequadsize, width=\imagequadsize]{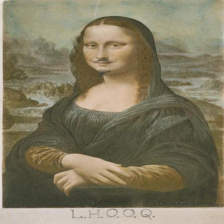} & \includegraphics[height=\imagequadsize, width=\imagequadsize]{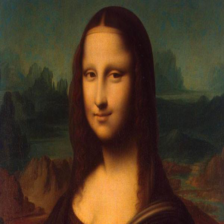} &  \includegraphics[height=\imagequadsize, width=\imagequadsize]{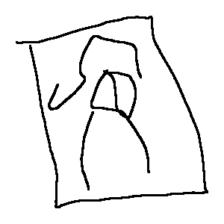} &  \includegraphics[height=\imagequadsize, width=\imagequadsize]{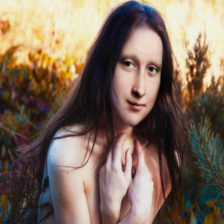} &  \includegraphics[height=\imagequadsize, width=\imagequadsize]{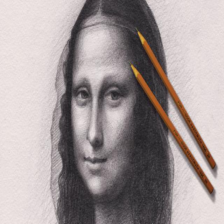}  & & \\
       &  & \includegraphics[height=\imagequadsize, width=\imagequadsize]{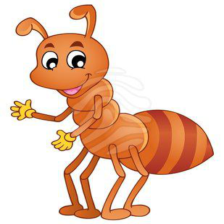} & \includegraphics[height=\imagequadsize, width=\imagequadsize]{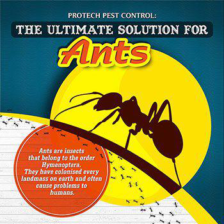} & \includegraphics[height=\imagequadsize, width=\imagequadsize]{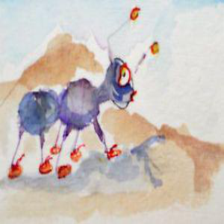} & \includegraphics[height=\imagequadsize, width=\imagequadsize]{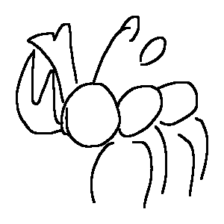} & \includegraphics[height=\imagequadsize, width=\imagequadsize]{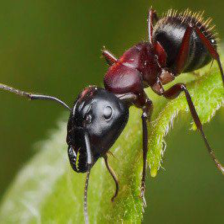} & \includegraphics[height=\imagequadsize, width=\imagequadsize]{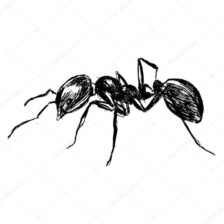}  & & \\
       \addlinespace
       \multirow{7}{*}{\textbf{Office-Home}} & \multirow{7}{*}{\textbf{\citep{VenkateswaraECP17}}} & \domainsize{Art} & \domainsize{Clipart} & \domainsize{Product} & \domainsize{Real} & &  & & \\
       & & \includegraphics[height=\imagequadsize, width=\imagequadsize]{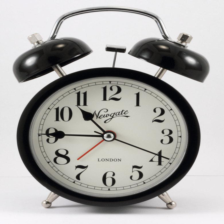} & \includegraphics[height=\imagequadsize, width=\imagequadsize]{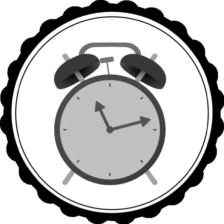} & \includegraphics[height=\imagequadsize, width=\imagequadsize]{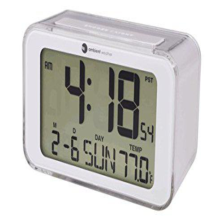} & \includegraphics[height=\imagequadsize, width=\imagequadsize]{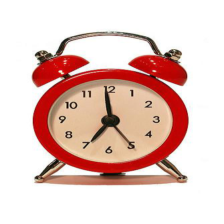} & &  & & \\
        & & \includegraphics[height=\imagequadsize, width=\imagequadsize]{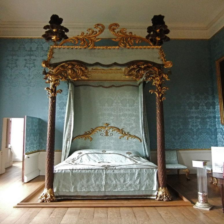} & \includegraphics[height=\imagequadsize, width=\imagequadsize]{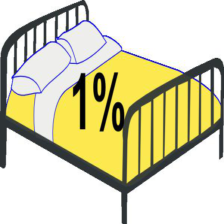} & \includegraphics[height=\imagequadsize, width=\imagequadsize]{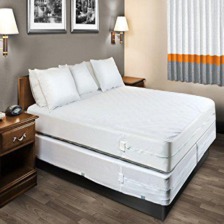} & \includegraphics[height=\imagequadsize, width=\imagequadsize]{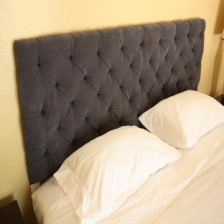} & &  & & \\
        \addlinespace
       \multirow{7}{*}{\textbf{VLCS}} & \multirow{7}{*}{\textbf{\citep{FangXR13}}} & \domainsize{V} & \domainsize{L} & \domainsize{C} & \domainsize{S} & &  & & \\ 
       & & \includegraphics[height=\imagequadsize, width=\imagequadsize]{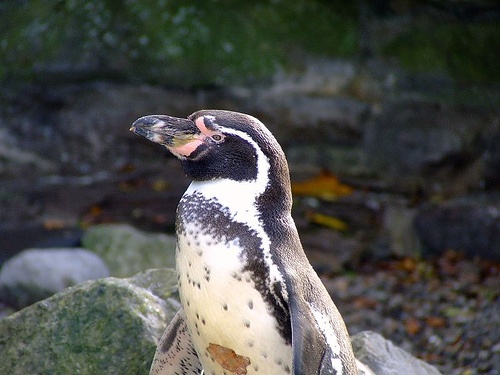}  &  \includegraphics[height=\imagequadsize, width=\imagequadsize]{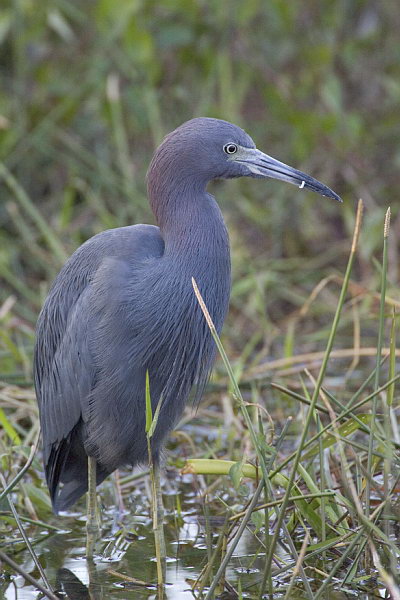} & \includegraphics[height=\imagequadsize, width=\imagequadsize]{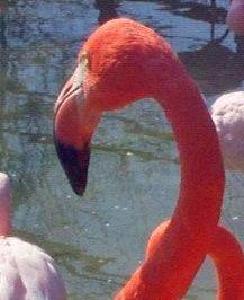} & \includegraphics[height=\imagequadsize, width=\imagequadsize]{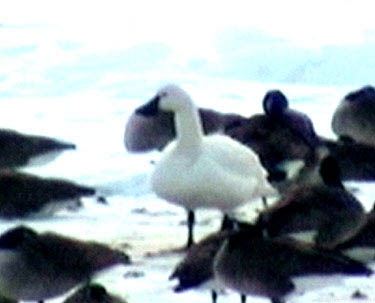} & &  & & \\
       & & \includegraphics[height=\imagequadsize, width=\imagequadsize]{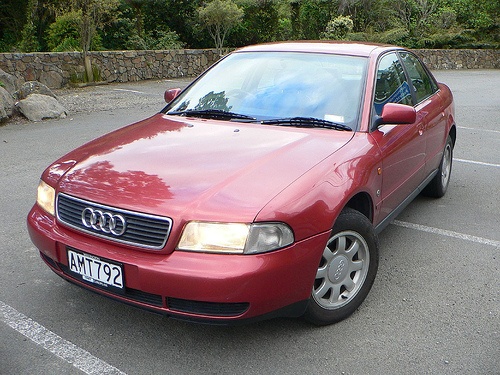} & \includegraphics[height=\imagequadsize, width=\imagequadsize]{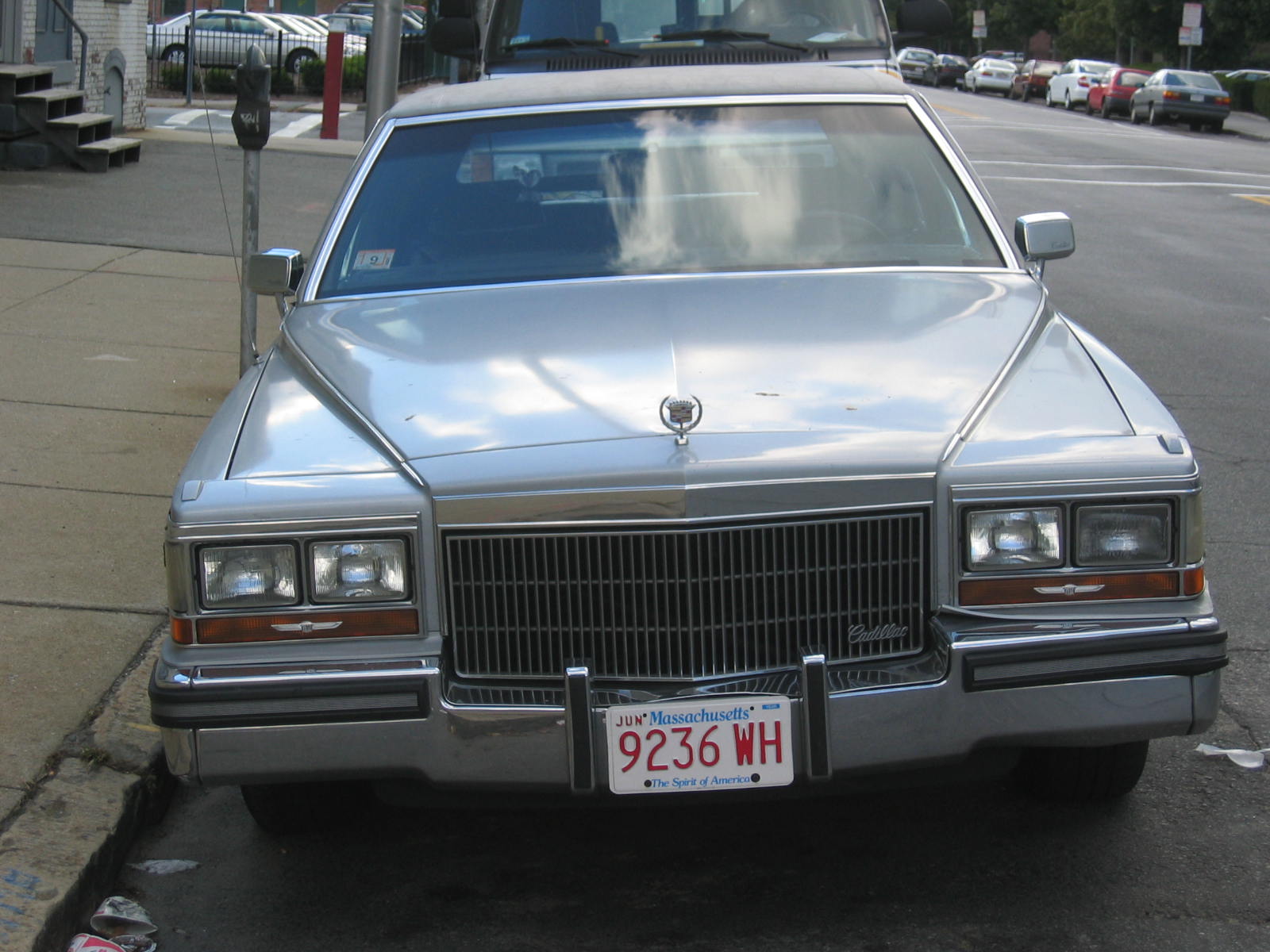} & \includegraphics[height=\imagequadsize, width=\imagequadsize]{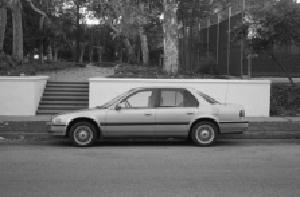} & \includegraphics[height=\imagequadsize, width=\imagequadsize]{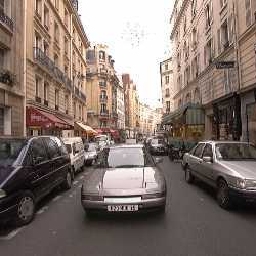} & &  & & \\
       \addlinespace
       \multirow{7}{*}{\textbf{Terra Incognita}} & \multirow{7}{*}{\textbf{\citep{BeeryHP18}}} & \domainsize{L100} & \domainsize{L38} & \domainsize{L43} & \domainsize{L46} & &  & & \\ 
       & & \includegraphics[height=\imagequadsize, width=\imagequadsize]{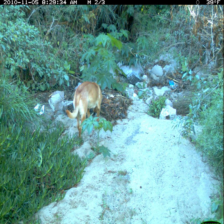} &  \includegraphics[height=\imagequadsize, width=\imagequadsize]{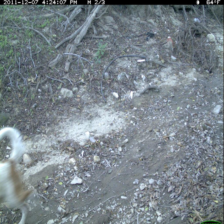} &
       \includegraphics[height=\imagequadsize, width=\imagequadsize]{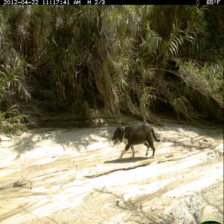} &  \includegraphics[height=\imagequadsize, width=\imagequadsize]{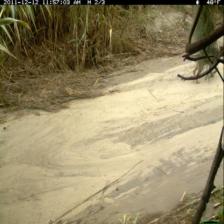} & &  & & \\
       & & \includegraphics[height=\imagequadsize, width=\imagequadsize]{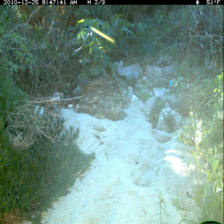} & \includegraphics[height=\imagequadsize, width=\imagequadsize]{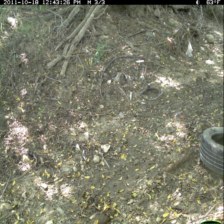} & \includegraphics[height=\imagequadsize, width=\imagequadsize]{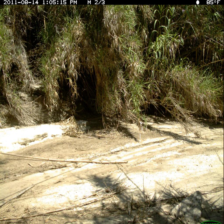} & \includegraphics[height=\imagequadsize, width=\imagequadsize]{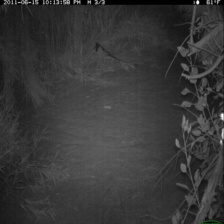} & &  & & \\
       \addlinespace
       \multirow{7}{*}{\textbf{PACS}} & \multirow{7}{*}{\textbf{\citep{LiYSH17}}}  & \domainsize{Art} & \domainsize{Cartoon} & \domainsize{Photo} & \domainsize{Sketch} & &  & & \\
       & & \includegraphics[height=\imagequadsize, width=\imagequadsize]{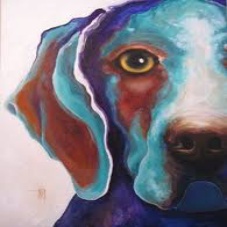} &  \includegraphics[height=\imagequadsize, width=\imagequadsize]{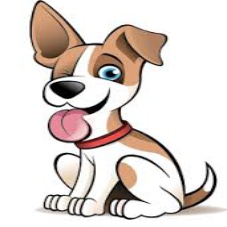} &  \includegraphics[height=\imagequadsize, width=\imagequadsize]{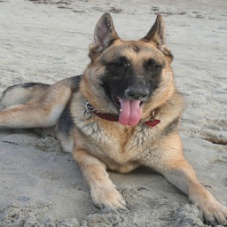} &  \includegraphics[height=\imagequadsize, width=\imagequadsize]{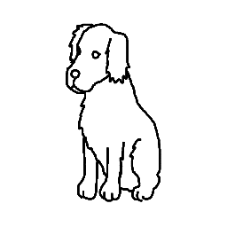} & &  & & \\
       & & \includegraphics[height=\imagequadsize, width=\imagequadsize]{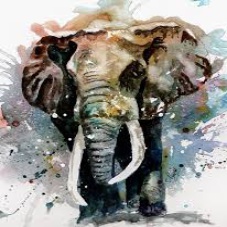} & \includegraphics[height=\imagequadsize, width=\imagequadsize]{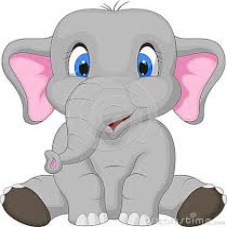} & \includegraphics[height=\imagequadsize, width=\imagequadsize]{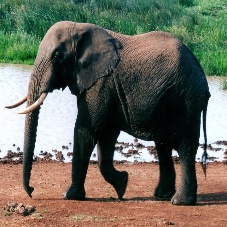} & \includegraphics[height=\imagequadsize, width=\imagequadsize]{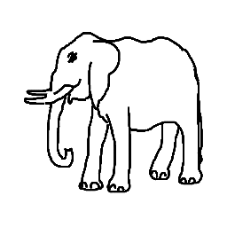} & &  & & \\
       \addlinespace
       \multirow{7}{*}{\textbf{Colored MNIST}} & \multirow{7}{*}{\textbf{\citep{arjovsky2019invariant}}} & \domainsize{$\plus90\%$} & \domainsize{$\plus80\%$} & \domainsize{$\minus90\%$} & & &  & & \\
       & & \includegraphics[height=\imagequadsize, width=\imagequadsize]{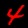} &  \includegraphics[height=\imagequadsize, width=\imagequadsize]{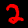} & \includegraphics[height=\imagequadsize, width=\imagequadsize]{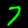} & & &  & & \\
        & & \includegraphics[height=\imagequadsize, width=\imagequadsize]{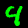} & \includegraphics[height=\imagequadsize, width=\imagequadsize]{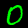} & \includegraphics[height=\imagequadsize, width=\imagequadsize]{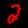}  & & &  & & \\
        \addlinespace
    \bottomrule
    \end{tabularx}
    \caption{Samples for two different classes across domains for popular datasets}
    \label{tab:common_examples}
\end{table}

\subsection{Colored MNIST}

The Colored MNIST (CMNIST) dataset \citep{arjovsky2019invariant} is another variation of the original MNIST dataset \citep{lecun-mnisthandwrittendigit-2010}. The grayscale images of MNIST got colored in red and green. The respective label corresponds to a combination of digit and color where the color correlates to the class label with factors $\{0.1,0.2,0.9\}$ as domains and the digit has a constant correlation of $0.75$. Since this is a synthetic dataset, the factors can easily be adapted or extended to more domains. We report these numbers since they are used in \citet{gulrajani2020search, arjovsky2019invariant}. Since the correlation factor between color and label varies between domains, this dataset is well-suited for determining a models capability of removing color as a predictive feature \citep{arjovsky2019invariant}.

To construct the dataset \citet{arjovsky2019invariant} first assign an initial label $\Tilde{y} = 0$ for digit $0-4$ and $\Tilde{y} = 1$ for digit $5-9$. This initial label is then flipped with a probability of $25\%$ to obtain the final label $y$. Finally, we obtain the color $z$ by flipping the label $y$ with probabilities $p^d \in \{0.1,0.2,0.9\}$ depending on the domain. The image is then colored red for $z=1$ or green for $z=0$ \citep{arjovsky2019invariant}. Samples for both classes across domains can be seen in \Cref{tab:common_examples}.

Overall, the dataset in \citet{gulrajani2020search} contains \num{70000} images from $2$ homogeneous classes (green \& red) of dimension $2 \times 28 \times 28$.

\subsection{Office-Home}
The Office-Home dataset \citep{VenkateswaraECP17}  provides \num{15588} images from $65$ categories across $4$ domains. The domains include Art, Clipart, Products (objects without a background), and Real-World (captured with a regular camera). Samples from these domains for the classes ``alarm-clock'' and ``bed'' can be seen in \Cref{tab:common_examples}. On average, each class contains around $70$ images with a maximum of $99$ images in a category \citep{VenkateswaraECP17}. In \citet{gulrajani2020search} they use dimension $3 \times 224 \times 224$ for each image.

\subsection{VLCS}

The VLCS dataset \citep{FangXR13} is a dataset that utilizes photographic datasets as individual domains. As such, it contains the domains PASCAL VOC (V) \citep{EveringhamGWWZ10}, LabelMe (L) \citep{RussellTMF08}, Caltech101 (C) \citep{Fei-FeiFP07}, and SUN09 (S) \citep{ChoiLTW10}. In total, there are \num{10729} images from $5$ classes. Samples for the classes ``bird'' and ``car'' can be seen in \Cref{tab:common_examples}. In \citet{gulrajani2020search} they use the dimension $3 \times 224 \times 224$ for each image. 

\subsection{PACS}
The PACS dataset \citep{LiYSH17} consists of images from different domains including Photo (P), Art (A), Cartoon (C), and Sketch (S) as individual domains. As such, it extends the previously photo-dominated data sets in domain generalization \citep{LiYSH17}. It includes seven homogeneous classes (dog, elephant, giraffe, guitar, horse, house, person) across the four previously mentioned domains. \Cref{tab:common_examples} shows samples from the ``dog'' and ``elephant'' classes across all domains. 

In total, PACS contains \num{9991} images which got obtained by intersecting classes from Caltech256 (Photo), Sketchy (Sketch) \citep{SangkloyBHH16}, TU-Berlin (Sketch) \citep{EitzHA12}, and Google Images (all but Sketch) \citep{LiYSH17}.

\subsection{Terra Incognita}
The Terra Incognita dataset is a subset of the initial Caltech camera traps dataset proposed by \citet{BeeryHP18}. It contains photographs of wild animals taken by camera traps at different locations (IDs: $38$, $43$, $46$, $100$) which represent the domains. As such, the version used by \citet{gulrajani2020search} contains \num{24788} images of $10$ classes, each with size $3\times 224 \times 224$. Samples from two different classes can be seen in \Cref{tab:common_examples}. The chosen locations represent the Top-4 locations with the largest number of images, each with more than \num{4000} images. 

The main data challenges that arise in this dataset include illumination, motion blur, size of the region of interest, occlusion, camouflage, and perspective \citep{BeeryHP18}. This includes animals not always being salient or them being small or far from the camera which results in only partial views of the animals' body being available \citep{BeeryHP18}. 

\subsection{DomainNet}

The DomainNet dataset \citep{PengBXHSW19} contains six domains: clipart (\num{48129} images), infographic (\num{51605} images), painting (\num{72266} images), quickdraw (\num{172500} images), real (\num{172947} images), and sketch (\num{69128} images) for $345$ classes. In total, it contains \num{586575} images that got accumulated by searching a category with a domain name in multiple image search engines and, as an exception, players of the game ``Quick Draw!'' for the quickdraw domain \citep{PengBXHSW19}.

To secure the quality of the dataset, they hired $20$ annotators for a total of \num{2500} hours to filter out falsely labeled images \citep{PengBXHSW19}. Each category has an average of $150$ images for the domains clipart and infographic, $220$ images for painting and sketch, and $510$ for the real domain \citep{PengBXHSW19}.

\subsection{ImageNet-C}
The ImageNet-C dataset \citep{HendrycksD19} contains images out of ImageNet \citep{RussakovskyDSKS15} permutated according to $15$ corruption types each with $5$ levels of severity which results in $75$ domains. The types of corruptions are out of the categories ``noise'', ``blur'', ``weather'', and ``digital'' \citep{HendrycksD19}. \Cref{tab:common_examples} shows a few of the available corruptions at severity $3$ for the same image sample of two different classes. As their corruptions, they provide Gaussian Noise, Shot Noise, Impulse Noise, Defocus Blur, Frosted Glass Blur, Motion Blur, Zoom Blur, Snow, Frost, Fog, Brightness, Contrast, Elastic, and Pixelate \citep{HendrycksD19}. Overall, the dataset provides all \num{1000} ImageNet classes where each image has the standard dimension of $3 \times 224 \times 224$. Currently, this dataset is not implemented by \citet{gulrajani2020search}.

\section{Considerations regarding model validation}
\label{sec:considerations}

In traditional supervised learning setups, we train our model on the training dataset, validate its hyperparameters (\eg number of layers or hidden units in a neural network) on a separate validation dataset, and finally evaluate our model on an unused test dataset. Notice, that the validation dataset should be distributed identically to the test data to properly fit the hyperparameters of our architecture. This is not a straightforward process for domain generalization, as we lack a proper validation dataset with the needed statistical properties. There exist several approaches to this problem, some of them being more grounded than others. Here we give an overview of the three approaches outlined by \citet{gulrajani2020search} that are respectively used in their DG framework \domainbed. 

\paragraph{Training-domain validation set} 
In this approach, each training domain gets further split into training and validation subsets where all validation subsets across domains get pooled into one global validation set. We can then maximize the model's performance on that global validation set to set the hyperparameters. This approach assumes the similarity of training and test distributions.

\paragraph{Leave-one-domain-out cross-validation}
We can train $s$ models with equal hyperparameters based on the $s$ training domains where we each hold one of the domains out of training. This allows us to validate on the held-out domain and average among them to calculate the global held-out domain accuracy. Based on that, we can choose a model and re-train it on all of the training domains. 

\paragraph{Test-domain validation set (oracle)}
A rather statistically biased way of validating the model's hyperparameters is incorporating the test dataset as a validation dataset. Because of this, it is considered bad style and should be avoided or at least explicitly marked. However, one method that is possible is to restrict the test dataset access as done by \citet{gulrajani2020search} where they prohibit early stopping and only use the last checkpoint.   

Other works have also come up with alternative methods to choose the hyperparameters. For example, \citet{krueger2020outofdistribution} validate the hyperparameters on all domains of the VLCS dataset and then apply the settings to PACS while \citet{DInnocenteC18} use a validation technique that combines probabilities specific to their method. 

\section{Deep-Dive into Representation Self-Challenging}
\label{sec:RSC}
Since some of our proposed methods use ideas from Representation Self-Challenging (RSC) \citep{huang2020selfchallenging}, we explain their approach more in detail here. They deploy two RSC variants called Spatial-Wise RSC and Channel-Wise RSC which they randomly alternate between. Generally, these are shown in \Cref{alg:SpatialRSC} and operate on features after the last convolutional layer. 

First, RSC calculates the gradient of the upper layer with respect to the latent feature representation according to \Cref{eq:gz}. Here, $\odot$ is the element-wise product and $\yoh$ is the one-hot encoding of the ground truth.
\begin{equation}
    \featureg = \frac{\partial( \classifier (\zz) \odot \yoh)}{\partial \zz}
    \label{eq:gz}
\end{equation}
Afterward, they average-pool the gradients to obtain $\featurega$. The key difference between the Spatial-Wise and Channel-Wise RSC lies in the average pooling done to compute $\featurega$ in line $5$ and the duplication in line $6$ in \Cref{alg:SpatialRSC}. While for Spatial-Wise RSC average pooling is done on the channel dimension according to \Cref{eq:SpatialRSCavg} yielding $\featuregal \in \mathbb{R}^{H_\zz \times W_\zz \times 1}$ for spatial location $(i,j)$, in Channel-Wise RSC the same computation is done on the spatial dimension with \Cref{eq:ChannelRSCavg} yielding $\featurega \in \mathbb{R}^{1 \times 1 \times K}$, a vector with the size of the feature map count.
\begin{equation}
\label{eq:SpatialRSCavg}
     \featuregal = \frac{1}{K} \sum_{k=1}^K \featuregl^k
\end{equation}
\begin{equation}
\label{eq:ChannelRSCavg}
     \featurega = \frac{1}{H_\mathbf{z}W_\mathbf{z}} \sum_{i=1}^{H_\mathbf{z}} \sum_{j=1}^{W_\mathbf{z}} \featuregl
\end{equation}
Depending on which dimensions are missing to get back to the original size of $\mathbf{z}, \featureg \in \mathbb{R}^{H_\mathbf{z} \times W_\mathbf{z} \times K}$, the computed values get duplicated along these dimensions. In the case of the Spatial-Wise RSC these are the channels, while for the Channel-Wise RSC these are the spatial dimensions. 

Next, \citet{huang2020selfchallenging} compute the $(100-p)\mathrm{th}$ percentile with the threshold value as $q_p$ and compute the mask $\mathbf{m}_{i,j}$ for spatial location $(i,j)$ based on \Cref{eq:Masking}. This mask is set to $0$ for the corresponding Top-$p$ percentage elements in $\featurega$ and therefore has the same shape. 
\begin{equation}
\mathbf{m}_{i,j}=\left\{\begin{array}{ll}
0, & \text { if } \quad \featuregal \geq q_{p} \\
1, & \text { otherwise }
\end{array}\right.
\label{eq:Masking}
\end{equation}
\citet{huang2020selfchallenging} apply the computed mask on the feature representation to yield $\Tilde{\mathbf{z}}_p = \mathbf{z} \odot \mathbf{m}$ which they validate using \Cref{eq:changeval}. This computes the difference with and without the masking in the correct class probabilities and yields a difference score for each sample in the vector $\mathbf{c}$.  
\begin{equation}
   \mathbf{c} = \sum_{c=1}^C  \left(\mathtt{softmax}(w(\mathbf{z})) \odot \yoh - \mathtt{softmax}(w(\Tilde{\mathbf{z}})) \odot \yoh \right)_c
   \label{eq:changeval}
\end{equation}
A positive value represents that the masking for that sample made the classifier \emph{less} certain about the correct class while a negative value represents the opposite and made the classifier \emph{more} certain about the correct class. Similar to previously, \citet{huang2020selfchallenging} calculate Top-$p$ of the positive values with the threshold as $b_p$. They revert the whole masking for all Top-$p$ samples inside each batch according to \Cref{eq:Masking-Reversion-RSC} where each spatial location $(i,j)$ of the mask associated with sample $n$ gets set back to $1$ if the condition applies, otherwise the mask values remain unchanged.
\begin{equation}
    \mathbf{m}^n_{i,j}=\left\{\begin{array}{ll}
1, & \text { if } \quad \mathbf{c}_n \leq b_{p} \\
-, & \text { otherwise }
\end{array}\right.
\label{eq:Masking-Reversion-RSC}
\end{equation}
Finally, we mask the features with the obtained final mask to obtain $\Tilde{\mathbf{z}} = \mathbf{z} \odot \mathbf{m}$, compute the loss $\mathcal{L}(w(\Tilde{\mathbf{z}}), \mathbf{y})$ and backpropagate to the whole network.

\begin{algorithm}[t]
    \SetAlgoLined
    \SetNoFillComment
    \SetKwInOut{Input}{Input}
    \Input{Data $\mathbf{X}, \mathbf{Y}$ with $\mathbf{x}_i \in \mathbb{R}^{H \times W \times 3}$, drop factor $p$, epochs $T$}
    \BlankLine
    \While{$epoch \leq T$}{
    \For{every sample (or batch) $\mathbf{x}, \mathbf{y}$}{
    Extract features $\mathbf{z} = \phi(\mathbf{x})$ \tcp*[r]{$\mathbf{z}$ has shape  $\mathbb{R}^{H_\mathbf{z} \times W_\mathbf{z} \times K} $}
    Compute gradient $\featureg$ w.r.t features according to \Cref{eq:gz} \;
    Compute $\featuregal$ by avg. pooling using $50\%$ \Cref{eq:SpatialRSCavg} or $50\%$ \Cref{eq:ChannelRSCavg} \;
    Duplicate $\featurega$ along channel/spatial dimension for initial shape\;
    Compute mask $\mathbf{m}_{i,j}$ according to \Cref{eq:Masking}\;
    Mask features to obtain $\Tilde{\mathbf{z}}_p = \mathbf{m} \odot \mathbf{z}$ \tcp*[r]{Evaluate effect of preliminary mask $\downarrow$}
    Compute change $\mathbf{c}$ according to \Cref{eq:changeval} \;
    Revert masking for specific samples according \Cref{eq:Masking-Reversion-RSC} \;
    Mask features $\Tilde{\mathbf{z}} = \mathbf{m} \odot \mathbf{z}$ \;
    Compute loss $\mathcal{L}(w(\Tilde{\mathbf{z}}), \mathbf{y})$ and backpropagate to whole network
    }
}
\caption{Spatial- and Channel-Wise RSC}
\label{alg:SpatialRSC}
\end{algorithm}

\paragraph{Problems}
Interestingly, since many architectures like ResNet-18/ResNet-50 deploy average pooling in their forward pass after the last convolutional layer, na\"ive Spatial-Wise RSC doesn't make sense since average pooling is done along the channel dimension and such architectures additionally average pool on the spatial dimension. This results in feature values getting spread evenly across the image regardless of the masking. Even though this isn't mentioned in their paper, they address this issue in their official repository and propose an alternative computation. For that, they calculate the mean $\featuregal$ from \Cref{eq:SpatialRSCavg} on the gradients of features from the previous convolutional layer, instead of the last one, and downsample it by factor $0.5$ to match the size.


\paragraph{Our Results}
In an effort to provide somewhat fair results which aren't too optimistic and neither too penalizing, we run the original \rsc code five times for each of the testing environments and compute the average performance in \Cref{tab:reproduced-RSC}. 

\begin{table}[hb]
    \centering
    \begin{tabular}{lccccc}
    \toprule
    \textbf{Run}   &  \textbf{P} & \textbf{A} & \textbf{C} & \textbf{S} \\
    \midrule
    1    & $93.23$  & $81.69$ & $78.11$ & $81.14$ \\
    2    & $93.41$  & $79.44$ & $77.38$ & $80.55$ \\
    3    & $94.37$  & $80.08$ & $76.58$ & $79.18$ \\
    4    & $93.71$  & $81.49$ & $78.84$ & $81.90$ \\
    5    & $93.95$  & $79.39$ & $76.75$ & $81.19$ \\
    \midrule
    Average & 93.73 & 80.41 & 77.53 & $80.79$ \\
    Reported & $95.99$ & $83.43$ & $80.31$ & $80.85$ \\
    \bottomrule
    \end{tabular}
    \caption[Reproduced results for Representation Self-Challenging using the official code base]{Reproduced results for Representation Self-Challenging using the official code base on the PACS dataset and with a ResNet-18 backbone.} 
    \label{tab:reproduced-RSC}
\end{table}

Given these observations, we follow other works such as \citet{nuriel2020permuted} and report our reproduced results whenever comparing to \rsc.

\chapter{Explainability in Deep Learning} 
\label{sec:Explainability} 

Machine Learning systems and especially deep neural networks have the characteristic that they are often seen as ``black-boxes'' \ie they are hard to interpret and pinpointing as a user how and why they converge to their prediction is often very difficult, if not impossible. Neural networks commonly lack transparency for human understanding \citep{sun2020fixing}. This property becomes a prominent impediment for intelligent systems deployed in impactful sectors like, for example, employment \citep{QinZXZJCX18, CaiSJLQXZ20, ZhaoHCFZ18}, jurisdiction \citep{GuoHQX019}, healthcare \citep{Pasa2019}, or banking loans where users would like to know the deciding factors for decisions. As such, we would like systems that are easily interpretable, relatable to the user, provide contextual information about the choice, and reflect the intermediate thinking of the user for a decision \citep{xie2020explainable}. Since these properties are very broad, it is not surprising that researchers in this field have very different approaches. For this chapter, we used the field guide by \citet{xie2020explainable} to paint an appropriate overview and properly introduce the different approaches. Commonly, methods try to provide better solutions with respect to \emph{Confidence}, \emph{Trust}, \emph{Safety}, and \emph{Ethics} to improve the overall explainability of the model \citep{xie2020explainable}:

\paragraph{Confidence}
The confidence of a machine learning system is high when the ``reasoning'' behind a decision between the model and the user matches often. For example, saliency attention maps \citep{ParkHARSDR18, HudsonM18} ensure that semantically relevant parts of an image get considered and therefore increase confidence.

\paragraph{Trust} 
Trust is established when the decision of an intelligent system doesn't need to be validated anymore. Recently, many works have studied the problem of whether a model can safely be adopted \citep{GharibLBADB18, VarshneyA17, JiangKGG18}. To be able to trust a model, we need to ensure satisfactory testing of the model and users need experience with it to ensure that the results commonly match the expectation \citep{xie2020explainable}.

\paragraph{Safety}
Safety needs to be high for machine learning systems that have an impact on people's lives in any form. As such, the model should perform \emph{consistently} as expected, prevent choices that may hurt the user or society, have high reliability under all operating conditions, and provide feedback on how the operating conditions influence the behavior.

\paragraph{Ethics}
The ethics are defined differently depending on the moral principles of each user. In general, though, one can create an ``ethics code'' on which a system's decisions are based off \citep{xie2020explainable}. Any sensitive characteristic \eg  religion, gender, disability, or sexual orientation are features that should be handled with great care. Similarly, we try to reduce the effect of any features that serve as a proxy for any type of discrimination process \eg living in a specific part of a city, such as New York City's Chinatown, can be a proxy for the ethical background or income. 

Since this chapter gives a high-level overview of recent advances in explainability for neural networks, also concerning domain generalization, it is up to the reader's background if this is necessary. 

\section{Related topics}
There exist several concepts which are related to explainable deep learning. Here, we explicitly cover \emph{model debugging} which tries to identify aspects that hinder training inference, and \emph{fairness and bias} which especially tackles the ethics trait to search for differences in regular and irregular activation patterns to promote robust and trustworthy systems \citep{xie2020explainable}. 

\subsection{Model Debugging}
Model debugging, similar to traditional software debugging, tries to pinpoint aspects of the architecture, data-processing, or training process which cause errors \citep{xie2020explainable}. It aims at giving more insights into the model, allowing easier solving of faulty behavior. While such approaches help to open the black-box of neural network architectures, we handle them distinctly from the other literature here. 

\citet{amershi2015modeltracker} propose \textsc{ModelTracker} which is a debugging framework and interactive visualization that displays traditional statistics like Area Under the Curve (AUC) or confusion matrices. It also shows how close samples are in the feature space and allows users to expand the visualization to show the raw data or annotate them. \citet{AlainB17} deploy linear classifiers to predict the information content in intermediate layers where the features of every layer serve as input to a separate classifier. They show that using features from deeper layers improves prediction accuracy and  that level of linear separability increases monotonically. \citet{fuchs2018scrutinizing} introduce \emph{neural stethoscopes} as a framework for analyzing factors of influence and interactively promoting and suppressing information. They extend the ordinary DNN architecture via a parallel two-layer perceptron at different locations where the input are the features from any layer from the main architecture. This stethoscope is then trained on a supplemental task and the loss is back-propagated to the main model with weighting factor $\lambda$ \citep{fuchs2018scrutinizing}. This factor controls if the stethoscope functions analytical ($\lambda = 0$), auxiliary ($\lambda > 0$), or adversarial ($\lambda < 0$) \citep{fuchs2018scrutinizing}. Further, \citet{KangRBZ20} use \emph{model assertions} which are functions for a model's input and output that indicate when errors may be occurring. They show that with these they can solve problems where car detection in successive frames disappears and reappears \citep{KangRBZ20}. Their model debugging is therefore implemented through a verification system \citep{xie2020explainable}.


\subsection{Fairness and Bias}
To secure model fairness, there exist several definitions which have emerged in the literature in recent years. \emph{Group fairness} \citep{CaldersKP09}, also known as  demographic parity or statistical parity, aims at equalizing benefits across groups with respect to protected characteristics (\eg religion, gender, etc.). By definition, if group $A$ has twice as many members as group $B$, twice as many people in group $A$ should receive the benefit when compared to $B$ \citep{xie2020explainable}. On the other hand, \emph{individual fairness} \citep{DworkHPRZ12} tries to secure that similar feature inputs get treated similarly. There also exist other notions of fairness such as \emph{equal opportunity} \citep{HardtPNS16}, \emph{disparate mistreatment} \citep{ZafarVGG17}, or other variations \citep{HeidariFGK18, WoodworthGOS17}.

Methods that try to ensure fairness in machine learning systems can be classified into three approaches which operate during different steps called \emph{pre-processing}, \emph{in-processing}, \emph{post-processing}:
\begin{enumerate}
    \item \textbf{Pre-processing} methods adapt the input data beforehand to remove features correlated to protected characteristics. As such, they try to learn an alternative feature representation without relying on these types of attributes \citep{GordalizaBGL19, CalmonWVRV17, LouizosSLWZ15, ZemelWSPD13}. 
    \item \textbf{In-processing} approaches add adjustments for fairness during the model learning process. This way, they punish decisions that are not aligned with certain fairness constraints \citep{DworkIKL18, DoniniOBSP18, AgarwalBD0W18}.
    \item \textbf{Post-processing} methods adjust the model predictions after training to account for fairness. It is the reassignment of class labels after classification to minimize classification errors subject to a particular fairness constraint \citep{HardtPNS16,PleissRWKW17, FeldmanFMSV15}.
\end{enumerate}

\section{Previous Works}



 
 Generally, we can divide methods for explainable deep neural networks in \emph{visualization}, \emph{model distillation}, and \emph{intrinsic} methods \citep{xie2020explainable}. While visualization methods try to highlight features that strongly correlate with the output of the DNN, model distillation builds upon a jointly trained ``white-box'' model, following the input-output behavior of the original architecture and aiming to identify its decision rules \citep{xie2020explainable}. Finally, intrinsic methods are networks designed to explain their output, hence they aim to optimize both, its performance and the respective explanations \citep{xie2020explainable}.

\subsection{Visualization}
Commonly, visualization methods use saliency maps to display the saliency values of the features \ie to which degree the features influence the model's prediction \citep{xie2020explainable}. We can further divide visualization methods into \emph{back-propagation} and \emph{perturbation}-based approaches where they respectively determine these values based on the volume of the gradient or between modified versions of the input \citep{xie2020explainable}. 

\subsubsection{Back-Propagation}
\label{sec:CAMs}
These approaches stick to the gradient passed through the network to determine the relevant features. As a simplistic baseline, one can display the partial derivative with respect to each input feature multiplied by its value \citep{xie2020explainable}. This way,  \citet{SimonyanVZ13} and \citet{SpringenbergDBR14} assess the sensitivity of the model for input changes \citep{xie2020explainable}. This can also be done for collections of intermediate layers \citep{ShrikumarGK17, MontavonLBSM17, Bach2015, ZeilerF14}.

\begin{figure}[htbp]
    \centering
    \includegraphics[width=\textwidth]{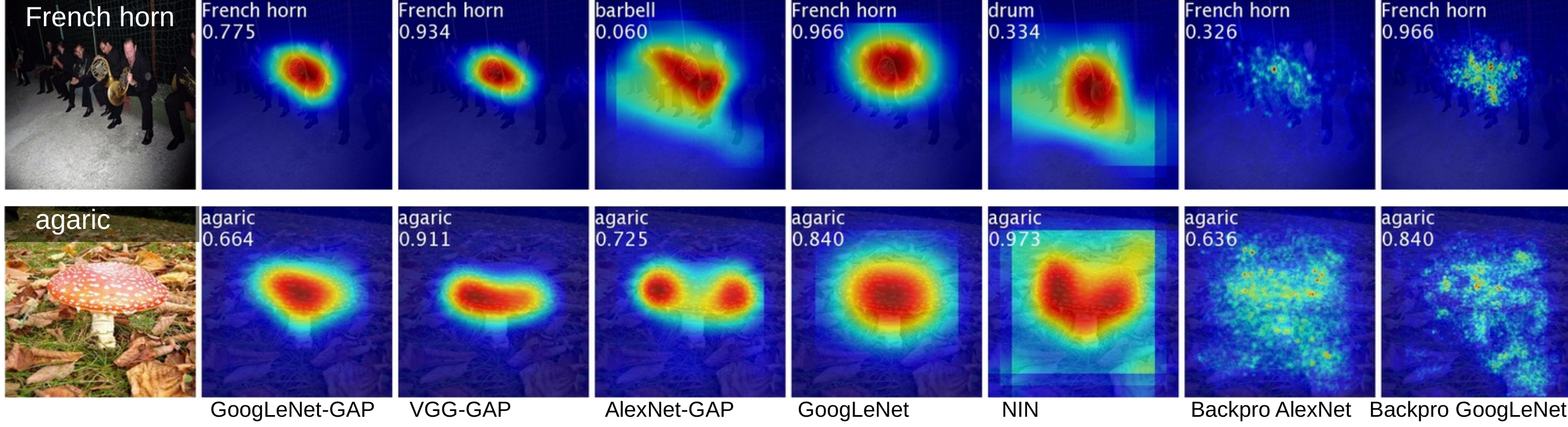}
    \caption[Class activation maps across different architectures]{Class activation maps across different architectures: \citep{ZhouKLOT16}}
    \label{fig:cams}
\end{figure}

\citet{ZhouKLOT16} introduce class activation maps (CAMs) which are shown in \Cref{fig:cams} based on global average pooling (GAP) \citep{LinCY13}. With GAP, they deploy the following CNN structure at the end of the network: \texttt{GAP(Convs)} $\rightarrow$ \texttt{Fully Connected Layer (FC)} $\rightarrow$ \texttt{softmax} where CAMs $\mathbf{M}_{c}$ for each class $c$ are then calculated according to \Cref{eq:cam}. Here, $K$ are the number of convolutional filters, $\mathbf{z}$ are the activations of the last convolutional layer, and $ w_{k, c}$ indicate the weights from the feature map $k$ of the last \texttt{Convolutional Layer} to logit for class $c$ of the \texttt{FC} \citep{ZhouKLOT16}.
\begin{equation}
\label{eq:cam}
    \mathbf{M}_{c}=\sum_{k}^{K} \mathbf{z}_k w_{k, c} 
\end{equation}
By upsampling the map to the image size, they can visualize the image regions responsible for a certain class \citep{ZhouKLOT16}. Therefore, every class has its own class activation map. 
The drawback of their approach is, that their method can only be applied to networks that use the \texttt{GAP(Convs)} $\rightarrow$ \texttt{Fully Connected Layer (FC)} $\rightarrow$ \texttt{softmax} architecture at the end \citep{xie2020explainable}. 

\citet{SelvarajuCDVPB17} solve this impediment by generalizing CAMs to gradient-weighted class activation maps (Grad-CAMs). Since their approach only requires the final activation function to be differentiable, they are generally applicable to a broader range of CNN architectures \citep{SelvarajuCDVPB17, xie2020explainable}. For that, they compute an importance score $\Tilde{\mathbf{g}}_{\mathbf{z},c}^k$ as:
\begin{equation}
\label{eq:grad_cam_importance}
     \Tilde{\mathbf{g}}_{\mathbf{z},c}^k = \frac{1}{H_\mathbf{z}W_\mathbf{z}} \sum_{i=1}^{H_\mathbf{z}} \sum_{j=1}^{W_\mathbf{z}} \frac{\partial y_c}{\partial \mathbf{z}^k_{i,j}}.
\end{equation}
Here, $y_c$ is the score before \texttt{softmax} and we calculate the gradient with respect to the feature map $\mathbf{z}^k$ in the final convolutional layer for every neuron positioned at $(i,j)$ in the $H_\mathbf{z} \times W_\mathbf{z}$ feature map \citep{SelvarajuCDVPB17, xie2020explainable}. Afterward, these importance scores get linearly combined for every feature map as shown in \Cref{eq:grad_cam_map} where they get additionally passed through a \texttt{ReLU} function:
\begin{equation}
\label{eq:grad_cam_map}
    \mathbf{M}_c = \mathtt{max}(0,\sum_{k=1}^K\Tilde{\mathbf{g}}_{\mathbf{z},c}^k \mathbf{z}^k).
\end{equation}
This computation inherently yields a $H_\mathbf{z} \times W_\mathbf{z}$ importance map ($14 \times 14$ for VGG \citep{SimonyanZ14a} and AlexNet \citep{KrizhevskySH12}, $7 \times 7$ for ResNet \citep{HeZRS16}) where we upsample it using bilinear interpolation onto the image size to yield the Grad-CAM.

Apart from CAMs, there also exist other methods like layer-wise relevance propagation \citep{MontavonLBSM17, DingLLS17, LapuschkinBMMS16, Bach2015}, deep learning important features (DeepLIFT) \citep{ShrikumarGK17}, or integrated gradients \citep{SundararajanTY17} which are not described in detail here. Please refer to \citet{xie2020explainable} or the original works for more information.

\subsubsection{Perturbation}
Perturbation methods alternate the input features to compute their respective relevance for the model's output by comparing the differences between the original and permutated version.

\citet{ZeilerF14} sweep a gray patch over the image to determine how the model will react to occluded areas. Once, an area with a high correlation to the output is covered, the prediction performance drops \citep{xie2020explainable, ZeilerF14}. \citet{li2016understanding} deploy a similar idea for NLP tasks where they erase words and measure the influence on the model's performance. \citet{FongV17} define three perturbations i) replacing patches with a constant value, ii) adding noise to a region, and  iii) blurring the area \citep{FongV17, xie2020explainable}. \citet{ZintgrafCAW17} propose a method based on \citet{Robnik-SikonjaK08} where they calculate the relevance of a feature for class $c$ through the prediction difference between including the respective feature or occluding it \citep{ZintgrafCAW17}. For that, they simulate the absence of each feature. A positive value for their computed difference means the feature influences the model's decision for class $c$ and a negative value means the feature influences the prediction against class $c$ \citep{ZintgrafCAW17}. \citet{ZintgrafCAW17} extend the initial method by \citet{Robnik-SikonjaK08} via removing patches instead of pixels and adapting the method for intermediate layers \citep{xie2020explainable}.

\subsection{Model distillation}
Model distillation methods allow for post-training explanations where we learn a distilled model which imitates the original model's decisions on the same data \citep{xie2020explainable}. It has access to information from the initial model and can therefore give insights about the features and output correlations \citep{xie2020explainable}. Generally, we can divide these methods into \emph{local approximation} and \emph{model translation} approaches. These either replicate the model behavior on a small subset of the input data based on the idea that the mechanisms a network uses to discriminate in a local area of the data manifold is simpler (local approximation) or stick to using the entire dataset with a smaller model (model translation) \citep{xie2020explainable}. 

\subsubsection{Local approximations}
Even though it may seem unintuitive to pursue approaches that don't explain every decision made by the DNN, practitioners often want to interpret decisions made for a specific data subset \eg employee performance indicators for those fired with poor performance \citep{xie2020explainable}. 
One of the most popular local approximations is the method proposed by \citet{Ribeiro0G16} called local interpretable model-agnostic explanations (\lime). They propose a notation where from an unexplainable global model $\modelf$ and an original representation of an instance $\xxi$ we want an interpretable model $\modelg$ from the class of potentially interpretable models $\modelg \in \gs$. Since not all models $\modelg$ have the same degree of interpretability, they define a complexity measure $\compl{\modelg}$ which could be the depth of the tree for decision trees or the number of non-zero weights in linear models \citep{Ribeiro0G16}. They incorporate this complexity measure, together with the prediction of $\modelg$ for $\modelf$ in a certain locality, in their loss term. There also exist many other works which build upon \lime{} to solve certain drawbacks  \citep{ElenbergDFK17, Ribeiro0G18}. We don't go into details here as it is only partially related to this thesis.

\subsubsection{Model Translation}
The idea of model translation is to mimic the behavior of the original deep neural network on the whole dataset, contrary to local approximations which only use a smaller subset. Some works have tried to distill neural networks into decision trees \citep{FrosstH17, tan2018learning, ZhangYMW19}, finite state automata \citep{HouZ20}, Graphs \citep{ZhangCWZ17, ZhangCSWZ18, ZhangYMW19}, or causal- and rule-based models \citep{harradon2018causal, MurdochS17}. Generally, the distilled models could be easier to deploy, faster to converge, or simply be more explainable \citep{xie2020explainable}.

\subsection{Intrinsic methods}
Finally, intrinsic methods jointly output an explanation in combination with their prediction. In an ideal world, such methods would be on par with state-of-the-art models without explainability. This approach introduces an additional task that gets jointly trained with the original task of the model \citep{xie2020explainable}. The additional task usually tries to provide either \emph{text explanations} \citep{HindWCCDMRV19, CamburuRLB18, HendricksARDSD16, ZellersBFC19}, an \emph{explanation association} \citep{DongSZZ17, LeiBJ16, IyerLL0SS18, Alvarez-MelisJ18}, or \emph{prototypes} \citep{LiLCR18, ChenLTBRS19} which differ in the provided explainability type as well as the degree of insight. 

\subsubsection{Attention mechanism}
The attention mechanism \citep{VaswaniSPUJGKP17} takes motivation from the human visual focus and peripheral perception \citep{schmidt2019recurrent}. With that, humans can focus on certain regions to achieve high resolution while adjacent objects are perceived with a rather low resolution \citep{schmidt2019recurrent}. In the attention mechanism, we learn a conditional distribution over given inputs using weighted contextual alignment scores (attention weights) \citep{xie2020explainable}. These allow for insights on how strongly different input features are considered during model inference \citep{xie2020explainable}. The alignment scores can be computed differently, for example, either content-based \citep{graves2014neural}, additive \citep{BahdanauCB14}, based on the matrix dot-product \citep{LuongPM15}, or as a scaled version of the matrix dot-product \citep{VaswaniSPUJGKP17}. Especially due to the transformer architecture \citep{VaswaniSPUJGKP17}, attention has shown to improve the neural network performance originally in natural language processing \citep{DevlinCLT19, brown2020language, lan2019albert}, but also more recently in image classification and other computer vision tasks \citep{AnwarB19, ZamirAKHK0020}. It has also been shown that attention is the update rule of a modern Hopfield network with continuous states \citep{ramsauer2020hopfield}, an architecture that hasn't been used very much in modern neural network models. There has also been a discussion on whether attention counts as explanation and to which degree the process offers insights into the inner workings of a neural network \citep{JainW19, WiegreffeP19, SenHYKR20}.

\subsubsection{Text explanations}
Text explanations are natural language outputs that explain the model decision using a form like ``This image is of class $A$ because of $B$''. As such, they are quite easy to understand regardless of the user's background. Works that take this approach are, for example, \citet{HendricksARDSD16} or \citet{ParkHARSDR18}. Drawbacks of textual explanations are that they i) require supervision for explanations during training and ii) explanations have been shown to be inconsistent which questions the validity of these types of explanations \citep{CamburuSMLB20}. 

\subsubsection{Explanation association}
Latent features or input elements that are combined with human-understandable concepts are classified under explanation associations. Such explanations either combine input or latent features with semantic concepts, associate the model prediction with a set of input elements, or utilize object saliency maps to visualize relevant image parts \citep{xie2020explainable}.

\subsubsection{Prototypes}
\label{sec:prototypes}

\begin{figure}[t]
    \centering
    \includegraphics[width=0.49\textwidth, height=2cm]{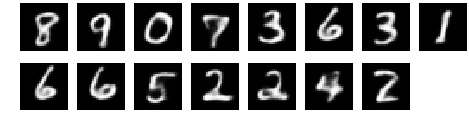}
    \includegraphics[width=0.4\textwidth, height=2cm]{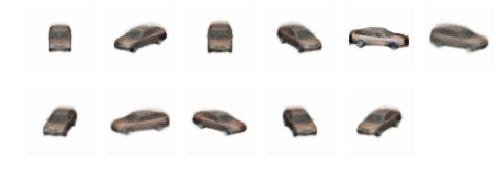}
    \caption[Prototypes for the MNIST and Car dataset]{Prototypes for the MNIST (left) and Car (right) dataset: \citep{LiLCR18}}
    \label{fig:prototypes}
\end{figure}

Finally, model prototype approaches are specifically designed for classification tasks \citep{Bien2011, KimRS14, PriebeMDS03, wu2017prototypal}. The term \emph{prototype} in few- and zero-shot learning settings are points in the feature space representing a single class \citep{LiLCR18}. In such methods, the distance to the prototype determines how an observation is classified. The prototypes are not limited to a single observation but can also be obtained using a combination of observations or latent representations \citep{xie2020explainable}. A \emph{criticism}, on the other hand, is a data instance that is not well represented by the set of prototypes \citep{molnar2019}. To obtain explainability using prototypes, one can trace the reasoning path for the prediction back to the learned prototypes \citep{xie2020explainable}. \citet{LiLCR18} use a prototype layer to deploy an explainable image classifier. They propose an architecture with an autoencoder and a prototype classifier. This prototype classifier calculates the $\ell^2$ distance between the encoded input and each of the prototypes, passes this through a fully connected layer to compute the sums of these distances, and finally normalizes them through a softmax layer \citep{LiLCR18}. Since these prototypes live in the same space as the encoded inputs, they can be visualized with a jointly trained decoder \citep{LiLCR18}. This property, coupled with the fully connected weights, allows for explainability through visualization of the prototypes and their respective influence on the prediction. \Cref{fig:prototypes} shows the visualizations for the generic number prototypes obtained by \citet{LiLCR18} on the MNIST \citep{lecun-mnisthandwrittendigit-2010} and the car angle prototypes on the Car \citep{FidlerDU12} dataset.

\begin{figure}[h]
    \centering
    \includegraphics[width=0.95\textwidth]{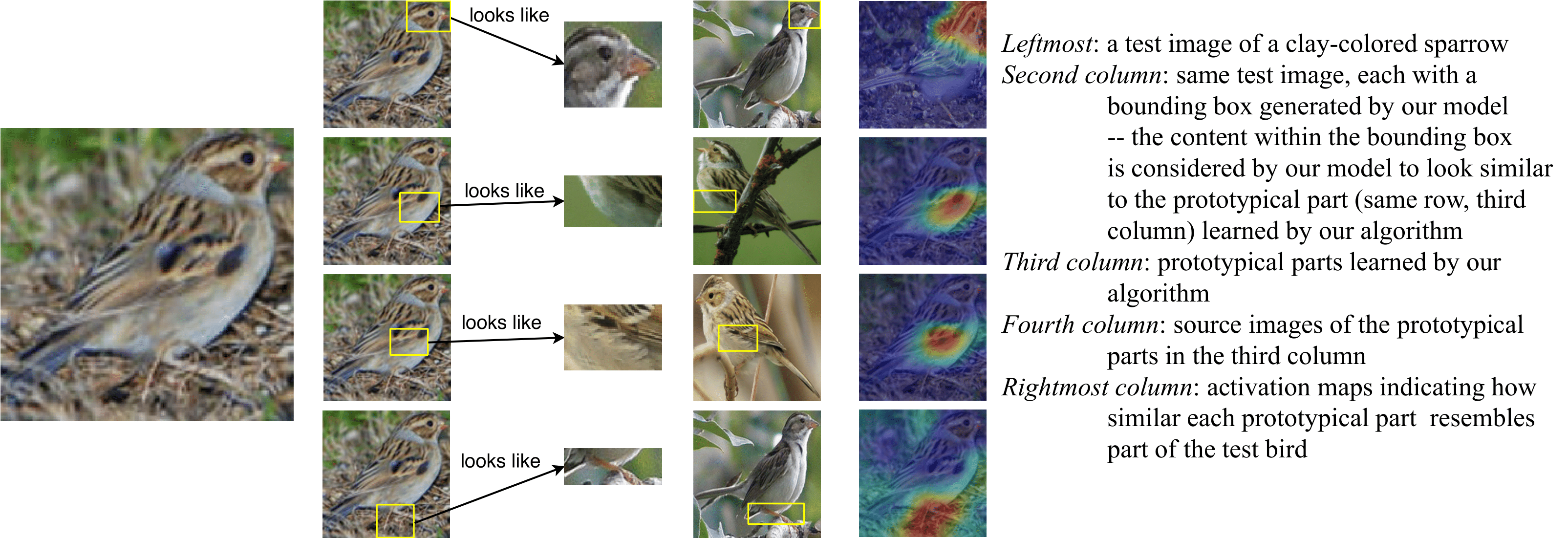}
    \caption[Image of a clay colored sparrow and its decomposition into prototypes]{Image of a clay colored sparrow and its decomposition into prototypes: \citep{ChenLTBRS19}}
    \label{fig:looke_like}
\end{figure}

\citet{ChenLTBRS19} introduce a prototypical part network (ProtoPNet) which has similar components to \citet{LiLCR18}, namely a convolutional neural network projecting onto a latent space and a prototype classifier. The approach chosen by \citet{ChenLTBRS19} is different as the prototypes are more fine-grained and represent parts of the input image \citep{xie2020explainable}. Hence, their model associates image patches with prototypes for explanations \citep{xie2020explainable}. \Cref{fig:looke_like} illustrates this approach for bird species classification.

As a general framework, we would like to learn $m$ prototypes $\prots=\left\{\prot\right\}_{j=1}^{m}$ with $\prot \in \mathbb{R}^{H_{\proti} \times W_{\proti} \times K}$ which each resemble a prototypical activation pattern in a patch of the convolutional output \citep{ChenLTBRS19}. Each prototype unit $\unit$ of the prototype layer $\player$ computes \emph{some} distance metric (\eg $\ell^2$ norm) between the $j$-th prototype $\prot$ and all patches of $\zz$ with the same shape as $\prot$ and inverts that into a similarity score using \emph{some} mapping function \citep{ChenLTBRS19}. This computation yields a similarity map $\simmap \in \mathbb{R}^{H_\zz \times W_\zz}$ which shows how representative the $j$-th prototype is for each latent patch and this can be upsampled to the initial image size for an overlay heatmap \citep{ChenLTBRS19}. When max-pooling this similarity map, we obtain a similarity score that measures how strong the $j$-th prototype is represented by \emph{any} latent patch. 

\citet{ChenLTBRS19} compute the maximum similarity score for each prototype unit $\unit$ by:
\begin{equation}
\label{eq:prot_layer_function}
    \unit(\zz) = \max_{\zpatch \in \text{patches}(\zz)} \log \left( \frac{\left\|\zpatch - \prot\right\|^2_2 + 1}{\left\|\zpatch - \prot\right\|^2_2 + \stability} \right),
\end{equation}
where the squared $\ell^2$ distance is used and $\stability$ is a small numerical stability factor. Their modified logarithm function satisfies the property of a similarity mapping function since with an increasing $\ell^2$-norm the function returns a smaller value \ie larger distance values correspond to smaller similarities. Keep in mind, that this needs to be appropriately adjusted when using any other distance metric. For example, both the cosine and dot product measures have \emph{increasing} similarity values for \emph{increasing} distance values. In theory, any Bregman divergence \citep{BanerjeeMDG04} is applicable as a distance metric. However, \citet{SnellSZ17} have observed that this choice can be very impactful and the $\ell^2$-norm has a better performance than cosine distance for few-shot tasks. 

To enforce that every class $c$ will be represented by at least one prototype, there are a pre-determined number of prototypes for each class which is denoted as $\prots_c$ with $\prots_c \subseteq \prots$. During training, \citet{ChenLTBRS19} minimize the objective:
\begin{equation}
\label{eq:min_prototypes}
    \min_{\prots, \p_\featureex} \frac{1}{n} \sum_{i=1}^n \mathcal{L}_{\mathrm{ce}}(\underbrace{\classifier \circ \player \circ \featureex}_{\text{Prediction}\ \ypred_{i}}, \yyi) + \lambda_1 \mathcal{L}_{\mathrm{clst}} + \lambda_2 \mathcal{L}_{\mathrm{sep}},
\end{equation}
where the cluster loss $\mathcal{L}_{\mathrm{clst}}$ and separation loss $\mathcal{L}_{\mathrm{sep}}$ are defined as:
\begin{alignat}{3}
\label{eq:prototype_losses}
    \mathcal{L}_{\mathrm{clst}} &= &&\frac{1}{n} \sum_{i=1}^n \min_{j: \prot \in \prots_{\yyi}} \min_{\zpatch \in \text{patches}(\zz)} \left\|\zpatch - \prot\right\|^2_2\\
    \mathcal{L}_{\mathrm{sep}} &= -&&\frac{1}{n} \sum_{i=1}^n \min_{j: \prot \notin \prots_{\yyi}} \min_{\zpatch \in \text{patches}(\zz)} \left\|\zpatch - \prot\right\|^2_2.
\end{alignat}
Note, that this is only the first part of a multi-step training procedure where \Cref{eq:min_prototypes} solely optimizes the parameters of the featurizer $\p_\featureex$ and the prototypes $\prots$, but not the classifier as its weights $\p_w$ are frozen with an initialization for each connection $w_{c,j}$ between the $j$-th prototype unit $\unit$ and the logit for class $c$ and $\forall j: \prot \in \prots_c$ as $w_{c,j} = 1$ while $\forall j: \prot \notin \prots_c$ it is set to $w_{c,j} = -0.5$. The positive connection for the similarity to a prototype of that specific class increases the prediction value for class $c$ while the negative connection for the similarity to a prototype of a different class decreases it. This initialization, together with the separation loss, guide the prototypes to represent semantic concepts for a class but also ensure that the same semantic concept is not learned by the other classes. Later on, \citet{ChenLTBRS19} optimize the classifier parameters $\p_\classifier$  for sparsity while fixing all other parameters to reduce the effect of \emph{negative} network reasoning for classification. 

While \citet{LiLCR18} need a decoder for visualizing the prototypes, \citet{ChenLTBRS19} don't require this component since they visualize the closest latent image patch across the full training dataset instead of directly visualizing the prototype \citep{ChenLTBRS19}. They also show that when combining several of their networks into a larger network, their method is on par with best-performing deep models \citep{ChenLTBRS19}.

\section{Explainability for Domain Generalization}

To the best of our knowledge, the only work using explanations for domain generalization is by \citet{zunino2020explainable}. They introduce a saliency-based approach utilizing a 2D binary map of pixel locations for the ground-truth object segmentation as input. This map contains a $1$ in a pixel location where the class label object is present and $0$ otherwise. Even though they were able to show that their method better focuses on relevant regions, we identify the additional annotations of class label objects as a major drawback which we solve in this work. Indeed, we not only avoid the use of annotations by directly restoring to Grad-CAMs (see \Cref{sec:divcam}), but our idea is also different in spirit. In fact, we do not force the network to focus on relevant regions as in \cite{zunino2020explainable} but we want the network to use i) different explanations for the same objects (to achieve better generalization) and ii) the explanations to be consistent across domains (to avoid overfitting a single input distribution).
\chapter{Proposed Methods}

In order to apply some of the previously mentioned topics from the explainability literature to the domain generalization task,  we specifically investigate the usage of gradient class activation maps from \Cref{sec:CAMs}, as well as prototypes from \Cref{sec:prototypes}. Our methods which are based upon these approaches are respectively described in \Cref{sec:divcam} (\divcam), \Cref{sec:prototype_networks} (\prodrop), and \Cref{sec:dtransformers} (\dtransformers).

\section{Diversified Class Activation Maps (\divcam)}
\label{sec:divcam}

\begin{figure}[t]
    \centering
    \includegraphics[width=\textwidth]{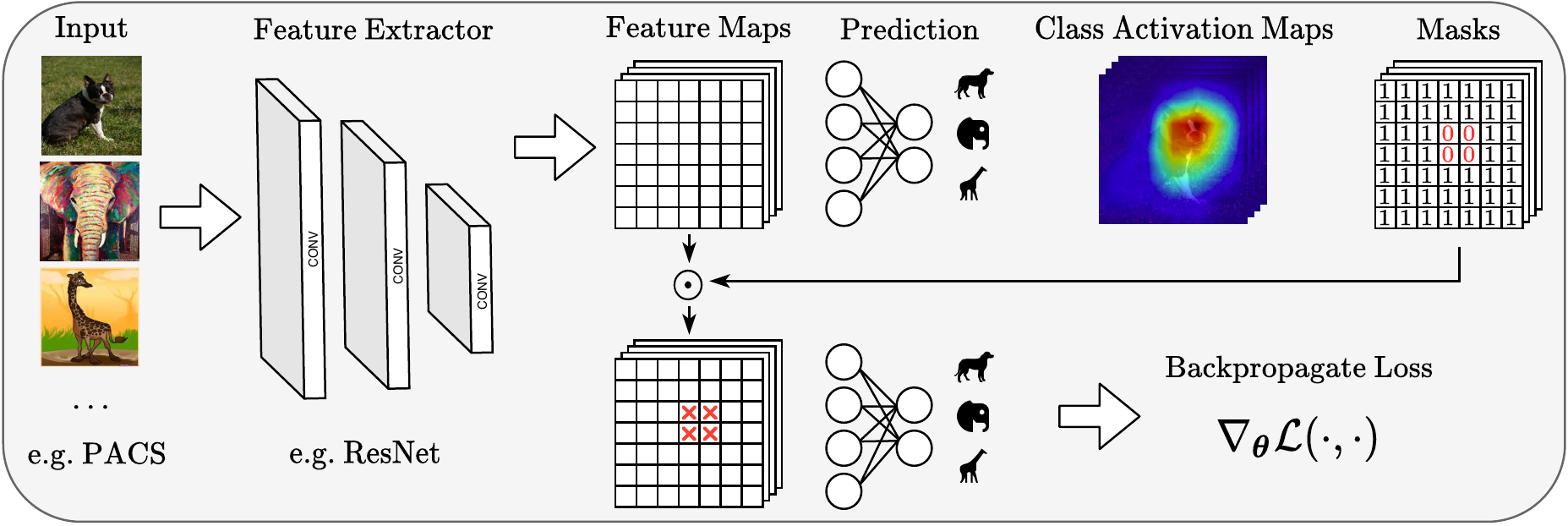}
    \caption{Visualization of the \divcam training process}
    \label{fig:divcam-overview}
\end{figure}

In \Cref{sec:RSC} we introduce the concept of Representation Self-Challenging for domain generalization while in \Cref{sec:CAMs} class activation maps, and specifically Grad-CAM gets introduced. It is quite easy to see that the importance scores $ \Tilde{\mathbf{g}}_{\mathbf{z},c}^k$ in Grad-CAM from \Cref{eq:grad_cam_importance} are a generalization of the spatial mean $\featurega$ used in Channel-Wise RSC from \Cref{eq:ChannelRSCavg}. The spatial mean $\featurega$ only computes the gradient with respect to the features for the most probable class while the importance scores $ \Tilde{\mathbf{g}}_{\mathbf{z},c}^k$ are formulated theoretically for all possible classes but similarly compute the gradient with respect to the feature representation. Both perform spatial average pooling.  

Despite the effectiveness of the model, we believe the approach of \citet{huang2020selfchallenging} does not fully exploit the relation between a feature vector and the actual content of the image. We argue (and experimentally demonstrate) that we can directly use CAMs to construct the self-challenging task. In particular, while the raw target gradient represents the significance of each channel in each spatial location for the prediction, CAMs allow us to better capture the actual importance of each image region. Thus, performing the targeted masking on highest CAM values means explicitly excluding the most relevant \emph{region} of the image that where used for the prediction, forcing the model to focus on other (and interpretable) visual cues for recognizing the object of interests. 

Therefore, as an intuitive baseline, we propose Diversified Class Activation Maps (\divcam), combining the two approaches as shown in \Cref{alg:ActivationMasking} or visualized on a high level in \Cref{fig:divcam-overview}. For that, during each step of the training procedure, we extract the features, compute the gradients with respect to the features as in \Cref{eq:gz}, and perform spatial average pooling to yield $\featurega$ according to \Cref{eq:ChannelRSCavg}. Our method deviates from Channel-Wise RSC by next computing class activation maps $\mathbf{M}_c \in \mathbb{R}^{H_\mathbf{z} \times W_\mathbf{z} \times 1}$ according to \Cref{eq:Map} for the ground truth class label.
\begin{equation}
\label{eq:Map}
    \mathbf{M}_c = \mathtt{max}(0,\sum_{k=1}^K\Tilde{\mathbf{g}}_{\mathbf{z},c}^k \mathbf{z}^k)
\end{equation}
Based on these maps and similar to \Cref{eq:Masking}, we compute a mask $\mathbf{m} \in \mathbb{R}^{H_\mathbf{z} \times W_\mathbf{z} \times 1}$ for the Top-$p$ percentile of map activations as: 
\begin{equation}
\mathbf{m}_{c,i,j}=\left\{\begin{array}{ll}
0, & \text { if } \quad \mathbf{M}_{c, i,j} \geq q_{p} \\
1, & \text { otherwise }
\end{array}\right.
\label{eq:MaskMap}
\end{equation}
As class activation maps and the corresponding masks are averaged along the channel dimension to be specific for each spatial location, we duplicate the mask along all channels to yield a mask with the same size of the features $\mathbf{m} \in \mathbb{R}^{H_\mathbf{z} \times W_\mathbf{z} \times K}$ which can directly be multiplied with the features to mask them and to regularize the training procedure:
\begin{equation}
\label{eq:mutate}
\Tilde{\mathbf{z}} = \mathbf{m} \odot \mathbf{z},
\end{equation}
where $\odot$ is the Hadamard product. The new feature vector $\Tilde{\mathbf{z}}$ is used as input to the classifier $w$ in place of the original $\mathbf{z}$ to regularize the training procedure. For the masked features, we compute the Cross-Entropy Loss from \Cref{eq:cross_entropy} and backpropagate the gradient of the loss to the whole network to update the parameters.

\begin{figure}[t]
    \centering
    \begin{tabularx}{\textwidth}{lYYYY}
       \textbf{Label}  & \textbf{Original} & \textbf{Step 300} & \textbf{Step 2700} & \textbf{Step 4500} \\[0.2cm]
        Giraffe & \includegraphics[width=\imagequadsizecams]{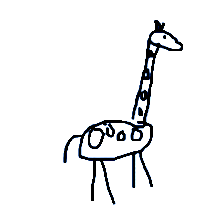} &  \includegraphics[width=\imagequadsizecams]{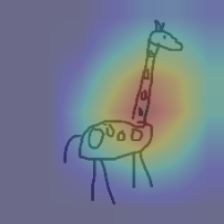} & \includegraphics[width=\imagequadsizecams]{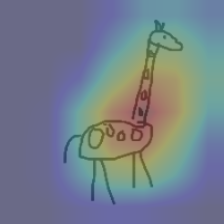} &
        \includegraphics[width=\imagequadsizecams]{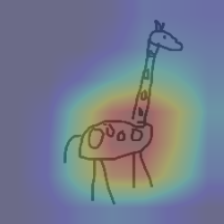} \\
        Elephant & \includegraphics[width=\imagequadsizecams]{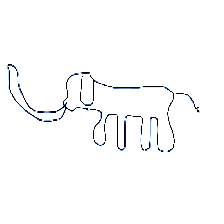} &
        \includegraphics[width=\imagequadsizecams]{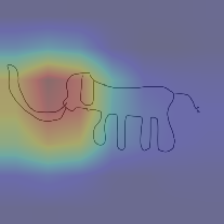} &
        \includegraphics[width=\imagequadsizecams]{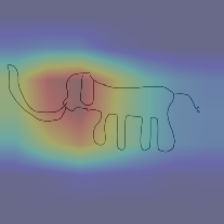} &
        \includegraphics[width=\imagequadsizecams]{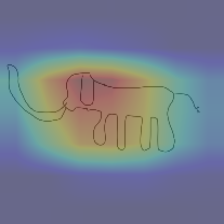} \\
    \end{tabularx}
    \caption[Used class activation maps in \divcams throughout training]{Used class activation maps for \divcams in update step $300/5000$, $2700/5000$, and $4500/5000$ using a ResNet-50 backbone. For the giraffe, we initially focus on the neck while our masks force the network to also take into consideration the overall shape, finally settling on the torso. For the elephant, we initially focus mainly on the elephant trunk and later guide the network towards taking also the shape into consideration.}
    \label{fig:cams_and_masks_divcam}
\end{figure}

Intuitively, constantly applying this masking for all samples within each batch disregards important features and results in relatively poor performance as the network isn't able to learn discriminative features in the first place. Therefore, applying the mask only for certain samples within each batch as mentioned by \citet[Secton~3.3]{huang2020selfchallenging} should yield a better performance. For convenience, we call this process \emph{mask batching}. On top of that, one could schedule the mask batching with an increasing factor (\eg linear schedule) such that masking gets applied more in the later training epochs where discriminative features have been learned. We apply the mask only if the sample $n$ is within the $(100-b)\mathrm{th}$ percentile of confidences for the correct class (stored in the change vector $\mathbf{c}$ for each sample $n$) and reset the mask otherwise by setting each spatial location $(i,j)$ back to $1$:
\begin{equation}
    \mathbf{m}^n_{c,i,j}=\left\{\begin{array}{ll}
1, & \text { if } \quad \mathbf{c}^n \leq q_{b} \\
-, & \text { otherwise }
\end{array}\right.
\label{eq:MaskingReversion}
\end{equation}
 where $ \mathbf{c}^n$ is is the confidence on the ground truth for sample $n$. This procedure enforces that masks only get applied to samples which are already classified well enough such that the network can now focus on other discriminative properties. For our full ablation study on applying the masks within each batch, please see \Cref{sec:ablation_study_batching}. \Cref{fig:cams_and_masks_divcam} also shows some of the class activation maps produced by \divcam throughout the training procedure.

To try and improve the effectiveness of our CAM-based regularization approach, we can borrow some practices from the weakly-supervised object localization literature. In particular, we explore the use of Homogeneous Negative CAMs (HNC) \citep{sun2020fixing} and Threshold Average Pooling (TAP) \citep{Bae2020RethinkingCAM}. Both methods improve the performance of ordinary CAMs and focus them better on the relevant aspects of an image. See \Cref{sec:abl-masks} for an evaluation of these variants.

\subsection{Global Average Pooling bias for small activation areas}
According to \citet{Bae2020RethinkingCAM}, one problem of traditional class activation maps is that the activated areas for each feature map differ by the respective channels because these capture different class information which isn't properly reflected in the global average pooling operation. Since every channel is globally averaged, smaller feature activation areas result in smaller globally averaged values despite a similar maximum activation value. This doesn't necessarily mean that one of the features is more relevant for the prediction, but can simply be caused by a large area with small activations. To combat this problem, the weight $w_{k,c}$ corresponding to the smaller value, is often trained to be higher when comparing two channels \citep{Bae2020RethinkingCAM}. Instead of the global average pooling operation, they propose \emph{Threshold Average Pooling} (TAP). When adapting their approach for our notation, we receive \Cref{eq:tap} where $\tau_{t a p} = \lambda_{tap} \cdot \mathtt{max}(\Tilde{\mathbf{z}}^k)$ with $\lambda_{tap} \in [0,1)$ as a hyperparameter and $p^k_{tap}$ denotes the scalar from the $k$-th channel of $\mathbf{p}_{tap}$ as it is a $k$-dimensional vector. 
\begin{equation}
\label{eq:tap}
p^k_{tap} =\frac{\sum_{i=1}^{H_\mathbf{z}} \sum_{j=1}^{W_\mathbf{z}} \mathds{1}\left(\Tilde{\mathbf{z}}^k_{i,j} > \tau_{t a p}\right) \Tilde{\mathbf{z}}^k_{i,j}}{\sum_{i=1}^{H_\mathbf{z}} \sum_{j=1}^{W_\mathbf{z}} \mathds{1}\left(\Tilde{\mathbf{z}}^k_{i,j} > \tau_{t a p}\right)}
\end{equation}
When incorporating this into \divcam, this results in changing the global average pooling after self-challenging has been applied to a threshold average pooling. Generally, this plug-in replacement can be seen as a trade-off between \emph{global max pooling} which is better at identifying the important activations of each channel and \emph{global average pooling} which has the advantage that it expands the activation to broader regions, allowing the loss to backpropagate.

\begin{algorithm}[t]
    \SetAlgoLined
    \SetKwInOut{Input}{Input}
    \Input{Data $\mathbf{X}, \mathbf{Y}$ with $\mathbf{x}_i \in \mathbb{R}^{H \times W \times 3}$, drop factor $p,b$, epochs $T$}
    \BlankLine
    \While{$epoch \leq T$}{
        \For{every batch $\mathbf{x}, \mathbf{y}$}{
            Extract features $\mathbf{z} = \phi(\mathbf{x})$ \tcp*[r]{$\mathbf{z}$ has shape  $\mathbb{R}^{H_\mathbf{z} \times W_\mathbf{z} \times K} $}
            Compute $\mathbf{g}_{\mathbf{z},c}$ with \Cref{eq:gz}\;
            Compute $\Tilde{\mathbf{g}}_{\mathbf{z},c}^k$ with \Cref{eq:ChannelRSCavg} \tcp*[r]{$\featurega$ has shape $\mathbb{R}^{1 \times 1 \times K}$}
            Compute $\mathbf{M}_c$ with \Cref{eq:Map} \tcp*[r]{$\mathbf{M}$ has shape $\mathbb{R}^{H_\mathbf{z} \times W_\mathbf{z} \times 1}$}
            Compute $\mathbf{m}_{c,i,j}$ with \Cref{eq:MaskMap} \;
            Repeat mask along channels \tcp*[r]{Afterwards $\mathbf{m}$ has shape $\mathbb{R}^{H_\mathbf{z} \times W_\mathbf{z} \times K}$}
            Adapt $\mathbf{m}_{c,i,j}$ with \Cref{eq:MaskingReversion} \;
            Compute $\Tilde{\mathbf{z}}$ with \Cref{eq:mutate} \;
            Backpropagate loss $\mathcal{L}_{ce}(w(\Tilde{\mathbf{z}}), \mathbf{y})$ \;
            }
    }
\caption{Diversified Class Activation Maps (\divcam)}
\label{alg:ActivationMasking}
\end{algorithm}

\subsection{Smoothing negative Class Activation Maps}

Based on the analysis of \citet{sun2020fixing}, negative class activation maps, \ie the class activation maps for classes other than the ground truth, often have false activations even when they are not present in an image. To solve this localization error, they propose a loss function which adds a weighted \emph{homogeneous negative CAM} (HNC) loss term to the existing Cross-Entropy loss. This is shown in \Cref{eq:hnc} where $\lambda_1$ controls the weight of the additional loss term. 
\begin{equation}
\label{eq:hnc}
\mathcal{L}_{neg} =\mathcal{L}_{ce}(\mathbf{y}, w(\Tilde{\mathbf{z}}))+\lambda_1 \mathcal{L}_{hnc}(\mathbf{y}, \boldsymbol{M})
\end{equation}
\citet{sun2020fixing} propose two approaches for implementing $\mathcal{L}_{hnc}$ in their work, both operating on the Top-$m$ most confident negative classes. The first one is based on the mean squared error which suppresses peak responses in the CAMs, while the second one utilizes the Kullback–Leibler (KL) divergence trying to minimize the difference between negative CAMs and an uniform probability map. Since they report similar performance for these variants and the KL loss applies a comparably smoother penalty, we use the KL divergence for our method:
\begin{equation}
\label{eq:hnc-kl}
\mathcal{L}_{hnc}(\mathbf{y}, \boldsymbol{M})=\sum_{c \in J^{m}_>} D_{K L}\left(\boldsymbol{U} \| \boldsymbol{M}_{c}^{\prime}\right).
\end{equation}
Here, $J^{m}_>$ is the set of Top-$m$ negative classes with the highest confidence score, $\boldsymbol{U} \in \mathbb{R}^{H_\mathbf{z} \times W_\mathbf{z}}$ is a uniform probability matrix with all elements having the value $(H_\mathbf{z}W_\mathbf{z})^{-1}$, and $\boldsymbol{M}_{c}^{\prime} = \sigma(\boldsymbol{M}_{c})$ is a probability map produced by applying the softmax function $\sigma$ to each negative class activation map $\boldsymbol{M}_{c}$. Plugging in the definition of the KL divergence and removing the constant as in \Cref{eq:kl-simplification} finally results in a simplified version as
\Cref{eq:hnc-kl-simple}.
\begin{equation}
\label{eq:kl-simplification}
D_{K L}\left(\boldsymbol{U} \| \boldsymbol{M}_{c}^{\prime}\right)=\sum_{i=1}^{H_\mathbf{z}} \sum_{j=1}^{W_\mathbf{z}} \boldsymbol{U}_{i,j} \cdot \log \left( \frac{\boldsymbol{U}_{i,j}}{\boldsymbol{M}_{c, i, j}^{\prime}} \right) =\text { const }-\frac{1}{H_\mathbf{z} W_\mathbf{z}} \sum_{i=1}^{H_\mathbf{z}} \sum_{j=1}^{W_\mathbf{z}} \log \left(\boldsymbol{M}_{c, i, j}^{\prime}\right)
\end{equation}
Generally, with this approach, we add two hyperparametes in the form of the weighting parameter $\lambda$ and the cut-off number $k$ for the Top-$k$ negative classes.
\begin{equation}
\label{eq:hnc-kl-simple}
    \mathcal{L}_{hnc}(\mathbf{y}, \boldsymbol{M})= -\frac{1}{H_\mathbf{z} W_\mathbf{z}} \sum_{c \in J^{m}_>} \sum_{i=1}^{H_\mathbf{z}} \sum_{j=1}^{W_\mathbf{z}} \log \left(\boldsymbol{M}_{c, i, j}^{\prime}\right)
\end{equation}
Since we use Grad-CAMs instead of ordinary CAMs in \divcam, na\"{i}vely applying this would require computing the gradient for every negative class $c$ in the set $J^m_>$ which would result in computing \Cref{eq:hnc-kl-grad-cam} where $y_c$ is the confidence of the negative class. 
\begin{equation}
\label{eq:hnc-kl-grad-cam}
    \mathcal{L}_{hnc}(\mathbf{y}, \boldsymbol{M})= -\frac{1}{H_\mathbf{z} W_\mathbf{z}} \sum_{c \in J^{m}_>} \sum_{i=1}^{H_\mathbf{z}} \sum_{j=1}^{W_\mathbf{z}} \log \left( \sigma \left(\mathtt{max}\left(0,\sum_{k=1}^K\left(\frac{1}{H_\mathbf{z}W_\mathbf{z}} \sum_{i=1}^{H_\mathbf{z}} \sum_{j=1}^{W_\mathbf{z}} \frac{\partial y_c}{\partial \mathbf{z}_{i,j}^k}\right)^k \mathbf{z}^k\right)\right)\right)
\end{equation}
To speed up the training for tasks with a large number of classes, we approximate the loss by summing the negative class confidences before backpropagating as shown in \Cref{eq:hnc-kl-grad-cam-approx}. This amounts to considering all negative classes within $J^\prime_>$ as one negative class. 
\begin{equation}
\label{eq:hnc-kl-grad-cam-approx}
    \widehat{\mathcal{L}}_{hnc}(\mathbf{y}, \boldsymbol{M})= -\frac{1}{H_\mathbf{z} W_\mathbf{z}} \sum_{i=1}^{H_\mathbf{z}} \sum_{j=1}^{W_\mathbf{z}} \log \left( \sigma \left(\mathtt{max}\left(0,\sum_{k=1}^K\left(\frac{1}{H_\mathbf{z}W_\mathbf{z}} \sum_{i=1}^{H_\mathbf{z}} \sum_{j=1}^{W_\mathbf{z}} \frac{\partial \sum_{c \in J^{m}_>} y_c}{\partial \mathbf{z}_{i,j}^k}\right)^k \mathbf{z}^k\right)\right)\right)
\end{equation}
To finally implement this into \divcam, we simply substitute the current loss $\mathcal{L}_{ce}(\mathbf{y}, w(\Tilde{\mathbf{z}}))$ in line $12$ from \Cref{alg:ActivationMasking} with \Cref{eq:hnc} where $\mathcal{L}_{hnc}(\mathbf{y}, \boldsymbol{M})$ is implemented through our approximation $\widehat{\mathcal{L}}_{hnc}(\mathbf{y}, \boldsymbol{M})$ given in \Cref{eq:hnc-kl-grad-cam-approx}.

Next, we can try to utilize domain information in \divcam by aligning distributions of class activation maps produced by the same class across domains. We want their distributions to align as close as possible such that we cannot identify which domain produced which class activation map. For that, we can utilize some methods previously introduced in \Cref{sec:invariant_features}, in particular we explore minimizing the sample maximum mean discrepancy introduced in \Cref{eq:mmd} and using a conditional domain adversarial neural network (CDANN). See \Cref{sec:abl-masks} for an evaluation of these variants.

\subsection{Conditional Domain Adversarial Neural Networks}
We combine the domain adversarial neural network (CDANN) approach, originally introduced by \citet{LiTGLLZT18}, with \divcam to align the distributions of CAMs across domains. For that, we try to predict the domain to which a class activation map belongs by passing it to a multi-layer perceptron $\omega$. We compute the cross entropy loss between the predictions and the domain ground truth $\mathbf{d}$ and weight it for each sample by the occurrence probability of the respective class.
After weighting, we can sum up all the losses and add it to our overall loss, weighted by $\lambda_2$:
\begin{equation}
\label{eq:adv}
\mathcal{L}_{adv} =\mathcal{L}_{ce}(\mathbf{y}, w(\Tilde{\mathbf{z}}))+\lambda_2 ( \mathcal{L}_{ce}(\mathbf{d}, \omega(\boldsymbol{M})) + \eta \left\|\nabla_{\boldsymbol{M}} \mathcal{L}_{ce}(\mathbf{d}, \omega(\boldsymbol{M}))\right\|_2).
\end{equation}
During each training step, we either update the discriminator, \ie the predictor for the domain, or the generator, \ie the main network including featurizer and classifier. The discriminator loss inherently includes a $\ell^2$ penalty on the gradients, weighted by $\eta$.

\subsection{Maximum Mean Discrepancy}

Given two samples $\mathbf{x}^{\xi_1}$ and $\mathbf{x}^{\xi_2}$ drawn from two individual, unknown  domain distributions $\mathcal{D}_{\xi_1}$ and $\mathcal{D}_{\xi_2}$, the maximum mean discrepancy (MMD) is given by \Cref{eq:mmd_maps} where $\varphi: \mathbb{R}^{d} \rightarrow \mathcal{H}$ is a feature map and $k(\cdot, \cdot)$ is the kernel function induced by $\varphi(\cdot)$. We consider every distinct pair of source domains $(\xi_u, \xi_v)$, representing training domains $\xi_u$ and $\xi_v$, with $\xi_u\neq \xi_v$ to be in the set $\mathfrak{P}$.
\begin{equation}
\label{eq:mmd_maps}
    \mathcal{L}_{dist} =\sum_{\xi_u,\xi_v \in \mathfrak{P}}\left\|\mathbb{E}_{\mathbf{x}^{\xi_u} \sim \mathcal{D}_{\xi_u}}[\varphi(\featureex(\mathbf{x}^{\xi_u}))]-\mathbb{E}_{\mathbf{x}^{\xi_v} \sim \mathcal{D}_{\xi_v}}[\varphi(\featureex(\mathbf{x}^{\xi_v}))]\right\|_{\mathcal{H}}
\end{equation}
In simpler terms, we map features into a reproducing kernel Hilbert space $\mathcal{H}$, and compute their mean differences within the RKHS. This loss pushes samples from different domains, which represent the same class, to lie nearby in the embedding space. According to \citet{SriperumbudurFGLS09}, this mean embedding is injective, \ie arbitrary distributions are uniquely represented in the RKHS, if we use a characteristic kernel. For this work, we choose the gaussian kernel shown in \Cref{eq:gaussian_kernel} which is a well-known characteristic kernel.
\begin{equation}
\label{eq:gaussian_kernel}
    k(x,x') = \exp \left(-\frac{\|x-x'\|^{2}}{2 \sigma^{2}}\right)
\end{equation}
Since the choice of kernel function can have a significant impact on the distance metric, we adopt the approach of \citet{LiPWK18} and use a mixture kernel by averaging over multiple choices of $\sigma$ as already implemented in \domainbed. This gets incorporated into our loss function weighted by $\lambda_3$ with:
\begin{equation}
    \mathcal{L}_{mmd} = \mathcal{L}_{ce}(\mathbf{y}, w(\Tilde{\mathbf{z}}))+\lambda_3 \mathcal{L}_{dist}.
\end{equation}
With this approach, we inherently align the computed masks by aligning the individual samples from different domains, aiming at producing domain invariant masks. This procedure can be applied at different levels \eg on the features, class activation maps, or masked class activation maps. In \Cref{sec:abl-masks}, we only provide results for the feature level due to the effectiveness that similar approaches showed in \domainbed. However, we observe a similar trend for the other application levels as well.

\section{Prototype Networks for Domain Generalization}
\label{sec:prototype_networks}

Another approach to combine explainability methods with the task of domain generalization is to use the prototype method outlined in \Cref{sec:prototypes}.  In particular, we can directly adapt the approach of \citet{ChenLTBRS19} as a baseline where we associate each class with a pre-defined number of prototypes. The cluster and separation losses from \Cref{eq:prototype_losses} ensure that each prototype resembles a prototypical attribute for the associated class and we minimize them according to \Cref{eq:min_prototypes}.

\begin{figure*}[t]
    \centering
    \includegraphics[width=\textwidth]{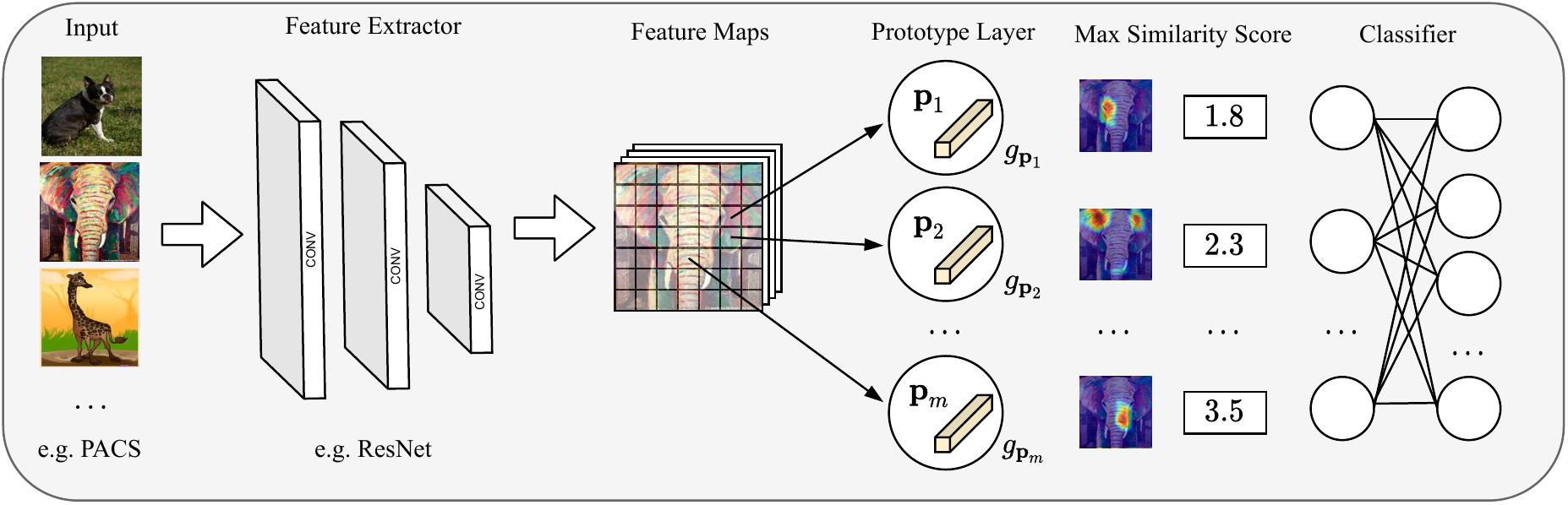}
    \caption{Domain-agnostic Prototype Network}
    \label{fig:domain_prototype_network}
\end{figure*}

For our application scenario, this prototype layer is used after the domain-agnostic featurizer $\phi$ and operates on the features like illustrated in \Cref{fig:domain_prototype_network}. As each prototype is trained with data from all training domains $\Xi$, they become inherently domain agnostic. This baseline uses a joint classifier $w$ to output the final prediction which operates on the maximum similarity for all the prototypes and \emph{some} latent patch. Similar to \citet{ChenLTBRS19}, we preface the prototype layer with two convolutional layers with kernel size $1$, a ReLU function between them, and finally a sigmoid activation function. We observe that having roughly $100$ initial update steps where only these in-between layers are trained is crucial for competitive performance. We anticipate that these steps are used to adapt the randomly-initialized convolutional weights to the image statics imposed by the pre-trained backbone.\footnote{Further implementation details can be found here: \url{https://github.com/SirRob1997/DomainBed/}} While this baseline is meaningful, we also consider a second variant where we build an ensemble of prototype layers, each learning domain-specific prototypes.

\subsection{Ensemble Prototype Network}
Following the intuition provided by works that utilize model ensembling, which have been described in \Cref{sec:model_ensembling}, we can use domain information by up-scaling the network to use a prototype layer for each domain separately. For the PACS dataset, this would correspond to having three prototype layers, one for each training domain \eg a photo, art, and cartoon prototype layer when predicting sketch images. Each prototype layer is only trained with images from their corresponding domain.

\begin{figure*}[t]
    \centering
    \includegraphics[width=\textwidth]{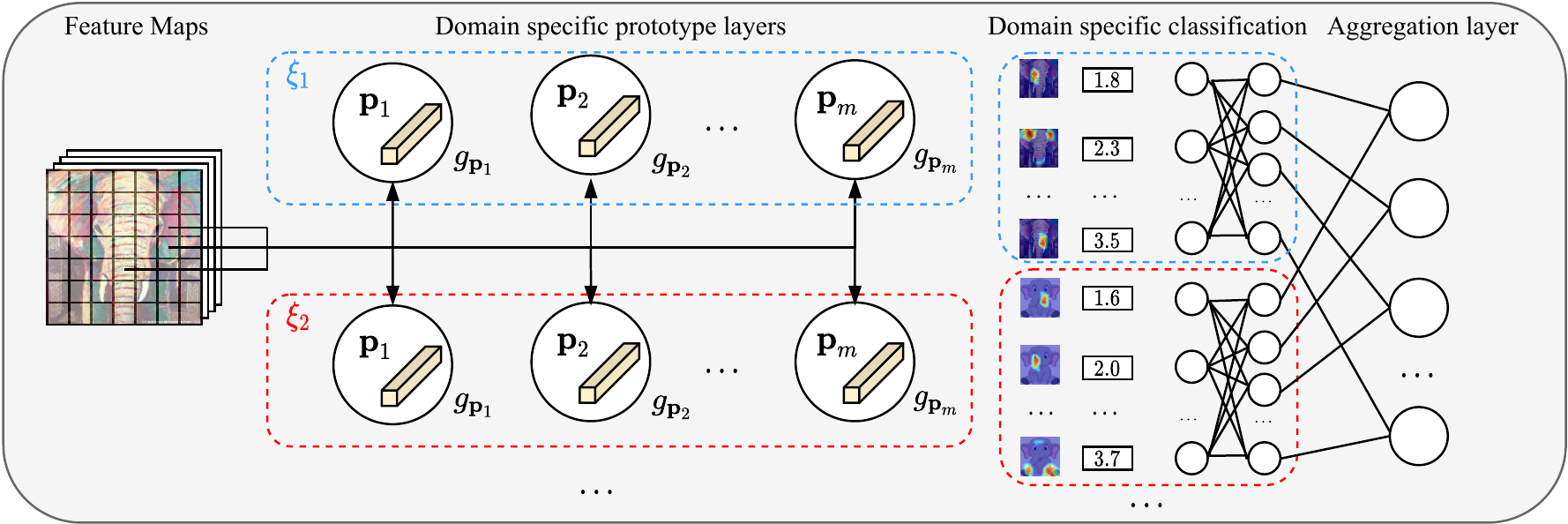}
    \caption{Ensemble Prototype Network}
    \label{fig:ensemble_prototype_network}
\end{figure*}

As shown in \Cref{fig:ensemble_prototype_network}, we associate each domain with both a prototype layer and a classifier which takes similarity scores of that domain's prototypes as input. During training, we only feed images of the associated domain to the respective prototype layer and classifier to enforce this domain correspondence. The aggregation weights of the final linear layer are set to a one-hot encoding representing the correct domain. During testing, we can then feed the new unseen domain to each domain prototype layer, allowing each domain's prototypes to influence the final prediction. 

There exist multiple strategies for setting the aggregation weights during this stage. The most simple version is to set the influence of each domain uniform, \ie if we have three training domains the connections from each domain would have the weight $\frac{1}{3}$ such that each domain has the same influence on the final prediction. Our second approach is to jointly train a domain predictor to output the weights for the aggregation layer which can either be used both, during training and testing, or only during testing, similar to what is done by \citet{ManciniBC018}. This method allows for a more flexible aggregation of the separated predictions coming from the different prototype layers, enabling the network to put more emphasis on the relevant domain prototypes.  

Lastly, we also experiment with an ensemble variant that is specific to prototype layers. Instead of only pre-defining a number of prototypes for each class in the domain agnostic prototype layer outlined in \Cref{sec:prototype_networks}, we can also pre-define domain correspondence for each prototype. Training is then done by passing each domain separately to the prototype layer and masking the overall prototype outputs if they do not correspond to the current environment. The cluster and separation losses are adapted to an average over the individual environments as:
\begin{alignat}{3}
\label{eq:prototype_losses_domain}
    \mathcal{L}_{\mathrm{clst}} &= \frac{1}{s} \sum_{\env \in \envs} &&\frac{1}{n_\env} \sum_{i=1}^{n_\env} \min_{j: \prot\in \prots_{\yyi}^\env} \min_{\zpatch \in \text{patches}(\zz^\env)} \left\|\zpatch - \prot\right\|^2_2\\
    \mathcal{L}_{\mathrm{sep}} &= \frac{1}{s} \sum_{\env \in \envs}-&&\frac{1}{n_\env} \sum_{i=1}^{n_\env} \min_{j: \prot \notin \prots_{\yyi}^\env} \min_{\zpatch \in \text{patches}(\zz^\env)} \left\|\zpatch - \prot\right\|^2_2,
\end{alignat}
where $\prots_{\yyi}^\env$ denotes the prototypes associated with the specific class \emph{and} environment while $\zz^\env$ denotes the latent representation of an image corresponding to the current domain. This ensemble variant inherently removes the need for setting appropriate aggregations weights as prototype activations are simply masked during training while during testing all prototypes are kept, allowing each prototype from each source domain to influence the prediction.  With a domain specific cross entropy loss:
\begin{equation}
    \lterm_{\mathrm{ce}} = \frac{1}{s} \sum_{\env \in \envs} \frac{1}{n_\env} \sum_{i=1}^{n_\env} -\sum_{c=1}^{C} y_{i, c}^{\env} \cdot \log \left(\hat{y}_{i, c}^{\env}\right),
\end{equation}
we optimize the final loss of our ensemble model as:
\begin{equation}
    \lterm = \lterm_{\mathrm{ce}} + \lambda_{\mathrm{clst}} \mathcal{L}_{\mathrm{clst}} + \lambda_{\mathrm{sep}} \mathcal{L}_{\mathrm{sep}}.
\end{equation}
While all of these ensemble variants \emph{should} work given the intuition from previous works that have been using model ensembles for learned domain-specific latent spaces, we observe that these assumptions do \emph{not} hold for prototype networks based on our additional experiments of the proposed variants. For us, any prototype ensemble was consistently outperformed by one domain-agnostic prototype layer. 

\subsection{Diversified Prototypes (\prodrop)}
\label{sec:divpro}

Initial experiments with both domain-agnostic and domain-specific prototype layers lead to unsatisfactory results. To investigate this behavior, we analyze the pairwise prototype $\ell_2$-distance as well as the cosine-distance $\cdistance$ which for any two prototypes $\proti_i$ and $\prot$ are given by:
\begin{alignat}{3}
\label{eq:pairwise_prot_distances}
    &\ell_2 &&= &&\left\|\proti_i - \prot  \right\|_2\\ 
    &\cdistance &&= 1- &&\frac{\proti_i\prot}{\left\|\proti_i \right\|_2 \left\|\prot \right\|_2} .
\end{alignat}
Through the $\ell_2$-distance we can grasp the euclidean distance between any two prototypes while the cosine-distance $\cdistance \in [0,2]$ is the inverted version of the cosine similarity which is a metric to judge the cosine of the angle between them. Here, we visualize the cosine-distance instead of the cosine similarity to match the color-scheme of the $\ell_2$-distance \ie low values resemble closeness. 

\begin{figure}[t]
    \centering
    \includegraphics[width=\textwidth]{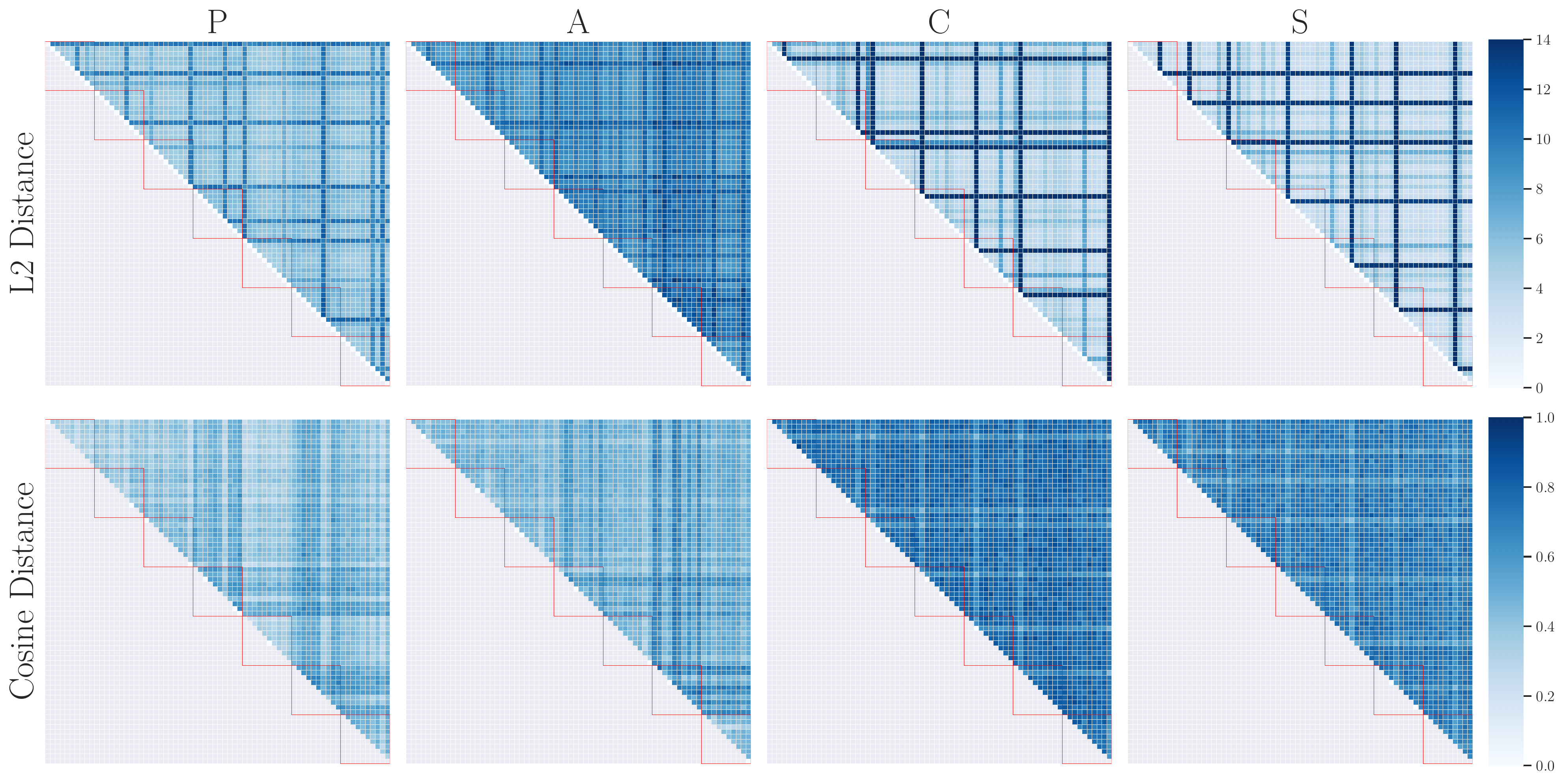}
    \caption[Second data split pairwise prototype distances with $w_{c,j} = -1.0$] {Pairwise learned prototype $\ell_2$-distance (top) and cosine-distance $\cdistance$ (bottom) of the best-performing model with negative weight $w_{c,j} = -1.0\; \forall j: \prot \notin \prots_c$ for each testing domain. Red squares denote prototype class correspondence for the $7$ different classes in the PACS dataset. No self-challenging is applied and colormap bounds are adjusted per metric for visualization purposes. Second data split.}
    \label{fig:pw_distance_trial1}
\end{figure}

\begin{figure}[t]
    \centering
    \includegraphics[width=\textwidth]{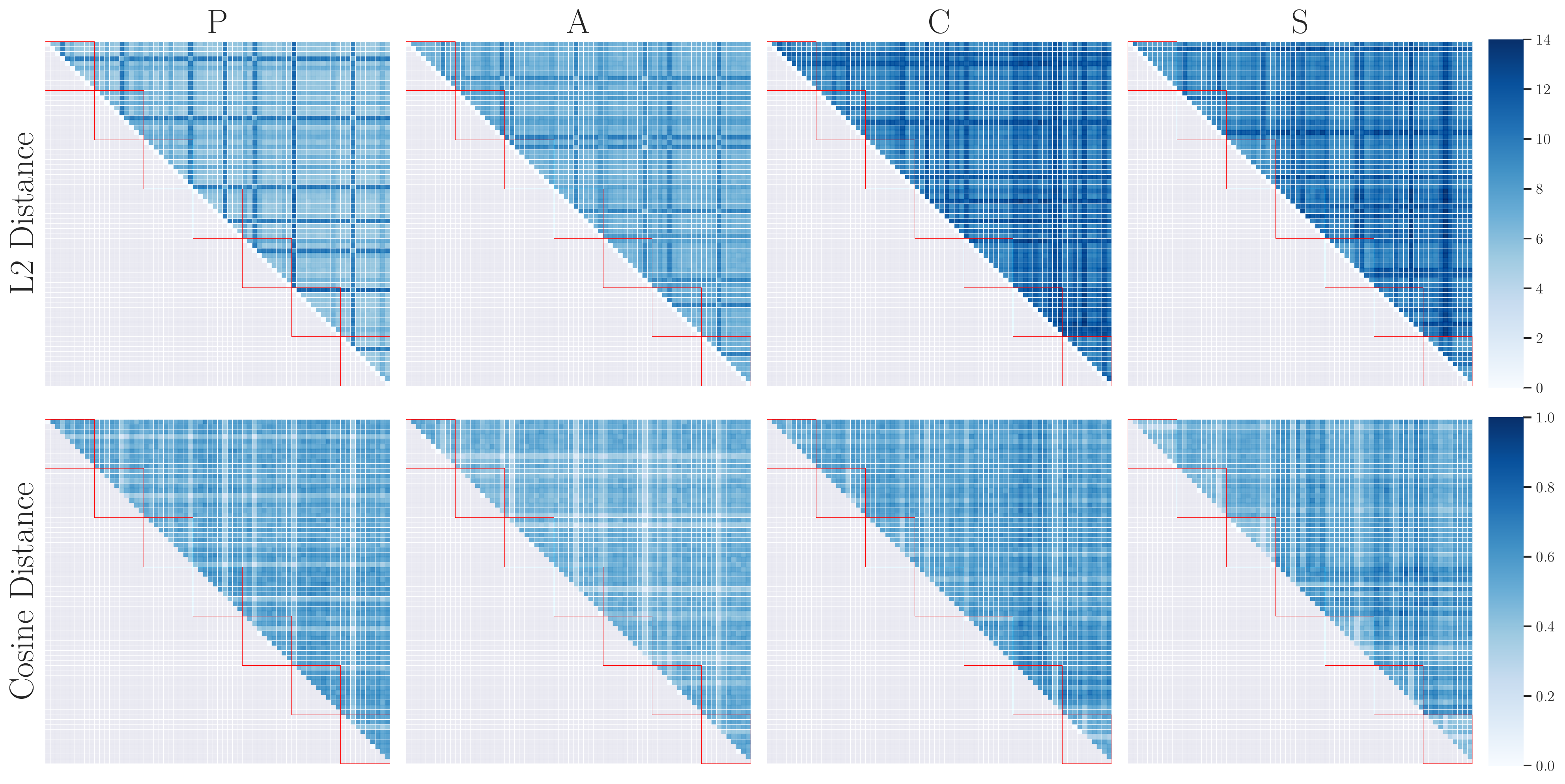}
    \caption[Second data split pairwise self-challenging prototype distances with $w_{c,j} = -1.0$] {Pairwise learned prototype $\ell_2$-distance (top) and cosine-distance $\cdistance$ (bottom) of the best-performing model with negative weight $w_{c,j} = -1.0\; \forall j: \prot \notin \prots_c$ for each testing domain. Red squares denote prototype class correspondence for the $7$ different classes in the PACS dataset. Self-challenging is applied and colormap bounds are adjusted per metric for visualization purposes. Second data split.}
    \label{fig:pw_distance_trial1-sc}
\end{figure}

The results of this analysis can be seen in \Cref{fig:pw_distance_trial1} and \Cref{fig:pw_distance_trial1-sc} for the first data split and a negative weight of $w_{c,j} = -1.0\; \forall j: \prot \notin \prots_c$ but we also show the same plots for $w_{c,j} = 0.0$ and the two additional data splits in \Cref{sec:additional_distances}. As both of the used metrics are symmetric, only the upper triangle is visualized. We observe that depending on the data split and negative weight, the model sometimes converges to having one or two prototypes per class that have a large $\ell_2$ distance while many of the other prototypes are close. Striking examples for this behavior can be seen in \Cref{fig:pairwise_distance} for the sketch domain or in \Cref{fig:pw_distance_trial1} for the sketch and cartoon domain. This observation suggests, that in these scenarios the model fails to properly use all of the available prototypes and only relies on a significantly reduced subset per class, not training the other prototypes. In relation to the cosine-distance, however, we can often observe that exactly these prototypes with a high pairwise $\ell_2$-distance to all the other prototypes have a slightly lower cosine-distance. Such behavior can be seen for example in \Cref{fig:pairwise_distance} in the art and cartoon environment or in \Cref{fig:pw_distance_trial1} for the sketch and cartoon domain. For the most part, many cosine-distances tend to be low and more or less uniformly spaced. Nevertheless, we can occasionally identify ``streaking'' patterns in the cosine-distances where prototypes for certain classes are well-spaced but they have a larger (or equal) cosine-distance to prototypes of other classes. See for example \Cref{fig:pairwise_distance} for the sketch domain, \Cref{fig:pw_distance_0.0_trial1} for the sketch and photo domain, and \Cref{fig:pw_distance_0.0_trial2}  for the sketch and cartoon domain.

From a design standpoint, we would like the prototypes within each class to be reasonably well spaced out in the latent space such that they can resemble different discriminative attributes about each class. That is, we would like the network to utilize all the prototypes and not only rely on a small subset of prototypes or discriminative features for their prediction. The distances of these prototypes to the prototypes of the other classes, however, should \emph{not} be restrained in any way and should be learned automatically. For example, when predicting different bird species, this allows the network to place similar head prototypes of different classes closer together. The existing cluster and separation losses enforce that each prototype associated with that class is close to at least one latent patch of that class while maximizing the distance to the prototypes of other classes. However, this does not enforce that each prototype associated with that class acts on a different discriminative feature.

One approach to possibly enforce this behavior is to incorporate the self-challenging method previously applied to \divcam to the presented prototype network resulting in a novel algorithm we call \emph{prototype dropping} (\prodrop) which is described in \Cref{alg:ProDrop}. In essence, we extract features by passing our input images to the featurizer $\mathbf{z} = \phi(\mathbf{x})$ and compute the similarity scores for each prototype by passing it to the prototype layer $\unit(\zz)$ with \Cref{eq:prot_layer_function}. Based on these similarity scores, we compute a mask $\mathbf{m}_{c,j}$ for the prototypes of the respective class $c$ with the Top-$p$ highest activation:
\begin{equation}
\mathbf{m}_{c,j}=\left\{\begin{array}{ll}
0, & \text { if } \quad \unit(\zz) \geq q_{c, p}\quad \forall j: \prot \in \prots_c \\
1, & \text { otherwise },
\end{array}\right.
\label{eq:ProDropFeatureMask}
\end{equation}
where $q_{c, p}$ is the corresponding threshold value. We also apply the mask batching from \divcam without scheduling which only applies this type of masking for the highest confidence samples on the ground truth. Finally, we can mask the samples using the Hadamard product $\odot$ with:
\begin{equation}
\label{eq:mutate_prodrop}
\mplayer(\zz) = \mathbf{m} \odot \player(\zz).
\end{equation}
In practice, all of these operations can be efficiently implemented using \texttt{torch.quantile}, \texttt{torch.lt}, and \texttt{torch.logical\_or} on small tensors, all of which pose no significant computational overhead.

\begin{algorithm}[t]
    \SetAlgoLined
    \SetKwInOut{Input}{Input}
    \Input{Data $\mathbf{X}, \mathbf{Y}$ with $\mathbf{x}_i \in \mathbb{R}^{H \times W \times 3}$, drop factor $p,b$, epochs $T$}
    \BlankLine
    \While{$epoch \leq T$}{
        \For{every batch $\mathbf{x}, \mathbf{y}$}{
            Extract features $\mathbf{z} = \phi(\mathbf{x})$ \tcp*[r]{$\mathbf{z}$ has shape  $\mathbb{R}^{H_\mathbf{z} \times W_\mathbf{z} \times K} $}
            Compute $\unit(\zz)$ with \Cref{eq:prot_layer_function}\;
            Compute $\mathbf{m}_{c,j}$ with \Cref{eq:ProDropFeatureMask} \;
            Adapt $\mathbf{m}_{c,j}$ with \Cref{eq:MaskingReversion} \;
            Compute $\mplayer(\zz)$ with \Cref{eq:mutate_prodrop} \;
            Backpropagate loss $\mathcal{L}_{ce}(w(\mplayer(\zz)), \mathbf{y}) + \lambda_4 \mathcal{L}_{\mathrm{clst}} + \lambda_5 \mathcal{L}_{\mathrm{sep}}$ \;
            }
    }
\caption{Prototype Dropping (\prodrop)}
\label{alg:ProDrop}
\end{algorithm}

The effect of this approach on the pairwise prototype distances can be seen in \Cref{fig:pairwise_distance_sc} as well as \Cref{sec:additional_distances}. We observe, that even though self-challenging helps to boost the overall performance (see \Cref{sec:abl_self_challenging}), it does not particularly well achieve the previously described desired distance properties and only improves them marginally. Positive effects can be seen in \Cref{fig:pw_distance_trial1-sc} for the sketch and cartoon domain, \Cref{fig:pw_distance_0.0_trial2-sc} for the sketch domain, or \Cref{fig:pw_distance_0.0_trial1-sc} for the cartoon domain.

Our second approach to enforce the desired distance structures is to add an additional intra-class prototype loss term $\mathcal{L}_\mathrm{intra}$ which maximizes the intra-class prototoype $\ell_2$- and/or cosine-distance weighted by $\lambda_6$. Again, this loss term can \emph{in theory} have a few different definitions depending on the chosen distance metrics, we experiment with: 
\begin{equation}
\label{eq:intra_loss}
    \mathcal{L}_\mathrm{intra} = \sum_{\proti_i, \prot \in \prots_c} \lambda_{\ell_2} \underbrace{\left\|\proti_i - \prot  \right\|_2 \vphantom{\frac{\proti_i\prot}{\left\|\proti_i \right\|_2 \left\|\prot \right\|_2}}}_{\ell_2-\text{distance}} + \lambda_{\cdistance} \underbrace{ (1-\frac{\proti_i\prot}{\left\|\proti_i \right\|_2 \left\|\prot \right\|_2})}_{\text{cosine-distance}},
\end{equation}
where the $\ell_2$-distance and the cosine-distance are weighted by $\lambda_{\ell_2}$ and $\lambda_{\cdistance}$ respectively. Performance results for the loss presented in \Cref{eq:intra_loss} with $\lambda_{\ell_2} = 1$ and $\lambda_{\cdistance} = 1$ are shown in \Cref{sec:intra_loss}. Since the cosine-distance is bounded, this commonly amounts to the $\ell_2$-distance having a higher influence. We also experimented with other values for $\lambda_{\ell_2}$ and $\lambda_{\cdistance}$, such as setting either one of them to zero and only applying either the $\ell_2$- or the cosine-distance ($\lambda_{\ell_2} = 0$, $\lambda_{\cdistance} = 1$ and $\lambda_{\ell_2} = 1$, $\lambda_{\cdistance} = 0$), but couldn't find any further benefits by canceling or re-weighting them differently. 

\paragraph{Influence of the negative weight on the distance metrics}
We also analyze the discrepancies between the distance metrics when comparing $w_{c,j} = -1.0\; \forall j: \prot \notin \prots_c$ and $w_{c,j} = 0.0\; \forall j: \prot \notin \prots_c$ which can be seen in the figures presented in \Cref{sec:additional_distances}. However, from these plots there were no \emph{consistent} trends observable which can be made on how the negative weight influences the training behavior of the prototypes. 

\subsection{Using Support Sets (\dtransformers)}
\label{sec:dtransformers}

Instead of directly learning the set of prototypes $\prots$, we can rely on a support set similar to what is done by \citet{DoerschGZ20} or \citet{SnellSZ17}. Here, the prototypes are based on a support set $\support$ consisting of $n_c$ sample images $\xxi$. This support set exists for each of the classes $c$ as $\supportc = \{\xxic\}_{i=1}^{n_c}$ where $\xxic$ is an image from class $c$. In the classical setting, the set of prototypes for each class $\prots_c$ is then obtained by computing one prototype for each class $\protc$ averaging the average-pooled latent representations of the support set as:
\begin{equation}
    \protc = \frac{1}{|\prots_c|} \sum_{\xxic \in \supportc} \featureex(\xxic)
\end{equation}
Contrary to the previous approach, by averaging all the average-pooled latent representations, there exists only one prototype for each class \ie $|\prots_c| = 1$ and $j=1$ which loses spatial information. Predictions are again made by computing \emph{some} distance function between the prototypes and the latent representation of the image to be classified. \citet{DoerschGZ20} preserve this spatial information of the feature extractor and use the attention weights to guide the averaging across support set images and latent patches in a method that they call \textsc{crossTransformers}. We extend their approach for the domain generalization case here, where we compute attention across multiple training environments. In the following sections this adaptation is referenced as \dtransformers. 

In particular, similar to Transformers \citep{VaswaniSPUJGKP17} and the original idea by \citet{DoerschGZ20}, we use three linear transformations to compute keys, values, and queries. In practice, the key head $\keyh : \mathbb{R}^{K} \mapsto \mathbb{R}^{\dimkey}$, the value head $\valueh : \mathbb{R}^{K} \mapsto \mathbb{R}^{\dimvalue}$, and the query head $\queryh : \mathbb{R}^{K} \mapsto \mathbb{R}^{\dimkey}$  can each be implemented through convolutions with \texttt{kernel\_size = 1}. Prototypes for each domain are computed by passing the support set $\supportc^\env = \{\xxic^\env\}_{i=1}^{n_{\env, c}}$ associated with domain $\env$ and class $c$ through the feature extractor. For each spatial location $m$ in the resulting features (indexed over $H_\mathbf{z} \times W_\mathbf{z}$) of the support set image $i$, we compute dot-product attention scores between keys $\keys = \keyh (\featureex(\xxic^\env))_m$ and the query vectors $\queries = \queryh(\featureex(\xxq^\env))_p$ for a query image $\xxq$ at spatial location $p$ (indexed over $H_\mathbf{z} \times W_\mathbf{z}$). Explicitly, the dot similarity $\attentionw$ between them is computed as:
\begin{equation}
    \attentionw = \keys \cdot \queries.
\end{equation}
Afterwards, we re-scale this dot similarity by using $\scaler=\sqrt{\dimkey}$ and obtain the final attention weights $\attentionwf$ using a \texttt{softmax} operation summing over all spatial locations and images in the support set:
\begin{equation}
    \attentionwf = \frac{\exp(\attentionw / \scaler)}{\sum_{i,m} \exp(\attentionw / \scaler)}.
\end{equation}
Finally, we can use the support-set values $\values = \valueh(\featureex(\xxic^\env))_m$  with the attention weights to compute a prototype vector per spatial location $p$ for each domain and class:
\begin{equation}
    \protipc = \sum_{i,m} \attentionwf \values.
\end{equation}
As a distance, we compute the dot similarity between the prototype and the query image values $\valuesq = \valueh(\featureex(\xxic^\env))_p$ where we sum over the training environments $\env \in \envs$ and the spatial locations $p$ of the query image:
\begin{equation}
    \operatorname{dist}(\xxic, \supportc) = \sum_{\env \in \envs} \frac{1}{H_\mathbf{z} W_\mathbf{z}} \sum_p \protipc \cdot  \valuesq.
\end{equation}
Most notably, we deploy the dot similarity as a distance metric here instead of the squared $\ell^2$-norm used by \citet{DoerschGZ20} since we observe numerical instability for that in our experiments which might be due to the distribution shift in domain generalization. In their work, they reason about using the same value-head $\valueh$ for the queries as used on the support-set images to ensure that the architecture works as a distance \ie if the support set contains the same images as the query, they want the euclidean distance to be $0$ \citep{DoerschGZ20}. Because we observe no performance gains for keeping them separate, we also use the same value head $\valueh$ for both the queries and support set images to reduce additional parameters to a minimum.

\chapter{Experiments}
In an effort to improve comparability and reproducability, we use \domainbed \citep{gulrajani2020search} for all ablation studies and experimental results. We compare to all methods currently in \domainbed which includes our provided implementation of \rsc. Further, all experiments show results for both \emph{training-domain validation} which assumes that training and testing domains have similar distributions  and \emph{oracle validation} which has limited access to the testing domain. We omit \emph{leave-one-domain-out cross-validation} as it requires the most computational resources and performs the worst out of the three validation techniques currently available in \domainbed \citep{gulrajani2020search}.

\section{Datasets and splits}
Since the size of the validation dataset can have a heavy impact on performance, we follow the design choices of \domainbed and choose $20\%$ of each domain as the validation size for all experiments and ablation studies. Here, we present results for VLCS \citep{FangXR13}, PACS \citep{LiYSH17}, Office-Home \citep{VenkateswaraECP17}, Terra Incognita \citep{BeeryHP18}, and DomainNet \citep{PengBXHSW19}. Although sometimes disregarded in the body of literature for domain generalization, we provide results for three different dataset splits to assess the stability regarding the model selection and to avoid overfitting on one split. 

\section{Hyperparameter Distributions \& Schedules}
For the main results, we use the official \domainbed hyperparameter distributions as well as the \adam optimizer with no learning rate schedule. This corresponds to the setup in which all baselines from \Cref{tab:perfom} have been evaluated in by \citet{gulrajani2020search} to provide a fair comparison. For the hyperparameters that get introduced in our methods, we choose similar distributions as used in the ablation studies from \Cref{sec:ablation}. Further, \Cref{sec:ablation} also shows the official distributions of \domainbed for all other shared hyperparameters such as learning rate $\alpha$ or batch size $\mathcal{B}$.

\section{Results}

\begin{table*}[t]
\small
\centering
\begin{tabular}{lccccc}
\toprule
\textbf{Algorithm} & \textbf{P} & \textbf{A} & \textbf{C} & \textbf{S} &  \textbf{Avg.} \\
\midrule
\divcams & 94.4 $\pm$ 0.7 & 80.5 $\pm$ 0.4 & 74.6 $\pm$ 2.2 & 79.0 $\pm$ 0.9 & 82.1 $\pm$ 0.3  \\
\prodrop & 93.6 $\pm$ 0.6 & 82.1 $\pm$ 0.9 & 76.4 $\pm$ 0.9 & 76.3 $\pm$ 0.6 & 82.1 $\pm$ 0.6 \\
\dtransformers & 93.7 $\pm$ 0.6 & 80.1 $\pm$ 1.2 & 73.3 $\pm$ 1.3 & 72.6 $\pm$ 1.5  & 79.9 $\pm$ 0.1 \\
\bottomrule
\end{tabular}
\caption[Performance comparison of the proposed methods on the PACS dataset]{Performance comparison of the proposed methods for the PACS dataset with a ResNet-18 backbone.}
\label{tab:pacs_ours}
\end{table*}

The high-level results for \divcams inside the \domainbed framework and across datasets are shown in \Cref{tab:perfom}. For completeness, we also show results outside of \domainbed on the official PACS split in \Cref{tab:official} using a ResNet-18 backbone. The full results, including the performance for choosing any domain inside each dataset as a testing domain, are shown in \Cref{sec:DomainResults}. While we are able to achieve state-of-the-art performance outside of the \domainbed framework, utilizing learning rate schedules and hyperparameter fine-tuning, we are also able to achieve Top-$4$ performance across five datasets within \domainbed. Notably, we outperform \rsc in almost every scenario, while exhibiting less standard deviation. This leads to more stable results and a more suitable method which can easily be used as a plug-and-play approach. In comparison with all other methods, we achieve good performance for the two most challenging datasets, namely Terra Incognita (Top-2) and DomainNet (Top-4), outperforming \rsc for both datasets by up to 2\%. This suggests, that directly reconstructing to Grad-CAMs in \divcams specifically provides value for the more challenging datasets in \domainbed. Keep in mind, that this is possible \emph{without} adding any additional parameters and while providing a framework where intermediate class activation maps can be visualized that guide the networks training process. While this might not be the \emph{best} explainability method, it certainly can offer more insights than  treating the optimization procedure as a block-box without this guidance.

\begin{table*}[t]
\footnotesize
\centering
\begin{tabular}{llcccccc}
\toprule
\textbf{Algorithm}  & \textbf{Ref.}       & \textbf{VLCS}             & \textbf{PACS}             & \textbf{Office-Home}       & \textbf{Terra Inc.}   & \textbf{DomainNet}        & \textbf{Avg.}              \\
\midrule
ERM                       & \citep{vapnik1998statistical}            & 77.5 $\pm$ 0.4            & 85.5 $\pm$ 0.2            & 66.5 $\pm$ 0.3            & 46.1 $\pm$ 1.8            & 40.9 $\pm$ 0.1            & 63.3                     \\
IRM                       & \citep{arjovsky2019invariant}             & 78.5 $\pm$ 0.5            & 83.5 $\pm$ 0.8            & 64.3 $\pm$ 2.2            & 47.6 $\pm$ 0.8            & 33.9 $\pm$ 2.8            & 61.5                      \\
GroupDRO                  & \citep{sagawa2019distributionally}        & 76.7 $\pm$ 0.6            & 84.4 $\pm$ 0.8            & 66.0 $\pm$ 0.7            & 43.2 $\pm$ 1.1            & 33.3 $\pm$ 0.2            & 60.7                      \\
Mixup                     & \citep{yan2020improve}            & 77.4 $\pm$ 0.6            & 84.6 $\pm$ 0.6            & 68.1 $\pm$ 0.3            & 47.9 $\pm$ 0.8            & 39.2 $\pm$ 0.1            & 63.4                      \\
MLDG                      & \citep{LiYSH18}            & 77.2 $\pm$ 0.4            & 84.9 $\pm$ 1.0            & 66.8 $\pm$ 0.6            & 47.7 $\pm$ 0.9            & 41.2 $\pm$ 0.1            & 63.5                      \\
CORAL                     &  \citep{SunS16}             & 78.8 $\pm$ 0.6            & 86.2 $\pm$ 0.3            & 68.7 $\pm$ 0.3            & 47.6 $\pm$ 1.0            & 41.5 $\pm$ 0.1            & 64.5                      \\
MMD                       & \citep{LiPWK18}           & 77.5 $\pm$ 0.9            & 84.6 $\pm$ 0.5            & 66.3 $\pm$ 0.1            & 42.2 $\pm$ 1.6            & 23.4 $\pm$ 9.5            & 58.8                      \\
DANN                      & \citep{GaninUAGLLML16}         & 78.6 $\pm$ 0.4            & 83.6 $\pm$ 0.4            & 65.9 $\pm$ 0.6            & 46.7 $\pm$ 0.5            & 38.3 $\pm$ 0.1            & 62.6                      \\
CDANN                     & \citep{LiTGLLZT18}          & 77.5 $\pm$ 0.1            & 82.6 $\pm$ 0.9            & 65.8 $\pm$ 1.3            & 45.8 $\pm$ 1.6            & 38.3 $\pm$ 0.3            & 62.0                      \\
MTL                       & \citep{blanchard2017domain}           & 77.2 $\pm$ 0.4            & 84.6 $\pm$ 0.5            & 66.4 $\pm$ 0.5            & 45.6 $\pm$ 1.2            & 40.6 $\pm$ 0.1            & 62.8                      \\
SagNet                    & \citep{nam2019reducing}             & 77.8 $\pm$ 0.5            & 86.3 $\pm$ 0.2            & 68.1 $\pm$ 0.1            & 48.6 $\pm$ 1.0            & 40.3 $\pm$ 0.1            & 64.2                      \\
ARM                       & \citep{zhang2020adaptive}           & 77.6 $\pm$ 0.3            & 85.1 $\pm$ 0.4            & 64.8 $\pm$ 0.3            & 45.5 $\pm$ 0.3            & 35.5 $\pm$ 0.2            & 61.7                      \\
VREx                      & \citep{krueger2020outofdistribution}            & 78.3 $\pm$ 0.2            & 84.9 $\pm$ 0.6            & 66.4 $\pm$ 0.6            & 46.4 $\pm$ 0.6            & 33.6 $\pm$ 2.9            & 61.9                      \\
RSC  		& \citep{huang2020selfchallenging}	      & 77.1 $\pm$ 0.5            & 85.2 $\pm$ 0.9            & 65.5 $\pm$ 0.9             & 46.6 $\pm$ 1.0           & 38.9 $\pm$ 0.5             & 62.7                      \\
\divcams               & (ours)            &77.8 $\pm$ 0.3            & 85.4 $\pm$ 0.2             & 65.2 $\pm$ 0.3           & 48.0 $\pm$ 1.2             & 40.7  $\pm$ 0.0    & 63.4                    \\
\dtransformers & (ours) & 78.7 $\pm$ 0.5  & 84.2 $\pm$ 0.1 & -- & 42.9 $\pm$ 1.1 & -- & -- \\
\midrule
ERM*           &   \citep{vapnik1998statistical}               	  & 77.6 $\pm$ 0.3            & 86.7 $\pm$ 0.3            & 66.4 $\pm$ 0.5            & 53.0 $\pm$ 0.3            & 41.3 $\pm$ 0.1            & 65.0                      \\
IRM*             &    \citep{arjovsky2019invariant}             	  & 76.9 $\pm$ 0.6            & 84.5 $\pm$ 1.1            & 63.0 $\pm$ 2.7            & 50.5 $\pm$ 0.7            & 28.0 $\pm$ 5.1            & 60.5                      \\
GroupDRO*      &   \citep{sagawa2019distributionally}                    & 77.4 $\pm$ 0.5            & 87.1 $\pm$ 0.1            & 66.2 $\pm$ 0.6            & 52.4 $\pm$ 0.1            & 33.4 $\pm$ 0.3            & 63.3                      \\
Mixup*               &   \citep{yan2020improve}            		 & 78.1 $\pm$ 0.3            & 86.8 $\pm$ 0.3            & 68.0 $\pm$ 0.2            & 54.4 $\pm$ 0.3            & 39.6 $\pm$ 0.1            & 65.3                      \\
MLDG*                 &  \citep{LiYSH18}           				& 77.5 $\pm$ 0.1            & 86.8 $\pm$ 0.4            & 66.6 $\pm$ 0.3            & 52.0 $\pm$ 0.1            & 41.6 $\pm$ 0.1            & 64.9                      \\
CORAL*                &  \citep{SunS16}           			  & 77.7 $\pm$ 0.2            & 87.1 $\pm$ 0.5            & 68.4 $\pm$ 0.2            & 52.8 $\pm$ 0.2            & 41.8 $\pm$ 0.1            & 65.5                      \\
MMD*               &   \citep{LiPWK18}               			& 77.9 $\pm$ 0.1            & 87.2 $\pm$ 0.1            & 66.2 $\pm$ 0.3            & 52.0 $\pm$ 0.4            & 23.5 $\pm$ 9.4            & 61.3                      \\
DANN*              &  \citep{GaninUAGLLML16}              		& 79.7 $\pm$ 0.5            & 85.2 $\pm$ 0.2            & 65.3 $\pm$ 0.8            & 50.6 $\pm$ 0.4            & 38.3 $\pm$ 0.1            & 63.8                      \\
CDANN*            &  \citep{LiTGLLZT18}              			& 79.9 $\pm$ 0.2            & 85.8 $\pm$ 0.8            & 65.3 $\pm$ 0.5            & 50.8 $\pm$ 0.6            & 38.5 $\pm$ 0.2            & 64.0                      \\
MTL*                  &  \citep{blanchard2017domain}           	  & 77.7 $\pm$ 0.5            & 86.7 $\pm$ 0.2            & 66.5 $\pm$ 0.4            & 52.2 $\pm$ 0.4            & 40.8 $\pm$ 0.1            & 64.7                      \\
SagNet*             &   \citep{nam2019reducing}           	& 77.6 $\pm$ 0.1            & 86.4 $\pm$ 0.4            & 67.5 $\pm$ 0.2            & 52.5 $\pm$ 0.4            & 40.8 $\pm$ 0.2            & 64.9                      \\
ARM*                  &   \citep{zhang2020adaptive}           		& 77.8 $\pm$ 0.3            & 85.8 $\pm$ 0.2            & 64.8 $\pm$ 0.4            & 51.2 $\pm$ 0.5            & 36.0 $\pm$ 0.2            & 63.1                      \\
VREx*                 &    \citep{krueger2020outofdistribution}       	  & 78.1 $\pm$ 0.2            & 87.2 $\pm$ 0.6            & 65.7 $\pm$ 0.3            & 51.4 $\pm$ 0.5            & 30.1 $\pm$ 3.7            & 62.5                      \\
RSC*  		& \citep{huang2020selfchallenging}	       & 77.8 $\pm$ 0.6            & 86.2 $\pm$ 0.5            & 66.5 $\pm$ 0.6            & 52.1 $\pm$ 0.2            & 38.9 $\pm$ 0.6             & 64.3                      \\
\tdivcams              & (ours) & 78.1 $\pm$ 0.6            & 87.2 $\pm$ 0.1            & 65.2 $\pm$ 0.5             & 51.3 $\pm$ 0.5           & 41.0 $\pm$ 0.0             & 64.6                      \\
\tdtransformers & (ours) & 77.7 $\pm$ 0.1 & 86.9 $\pm$ 0.3 & -- & 52.4 $\pm$ 0.8 & -- & -- \\
\bottomrule
\end{tabular}
\caption[Performance comparison across datasets]{Performance comparison across datasets using training-domain validation (top) and  oracle validation denoted with * (bottom). We use a ResNet-50 backbone, optimize with \adam, and follow the distributions specified in \domainbed. Only \rsc and our methods have been added as part of this work, the other baselines are taken from \domainbed.}
\label{tab:perfom}
\end{table*}

The fact that we are able to achieve state-of-the-art results for \divcams \emph{outside} of \domainbed shows a common problem with works in domain generalization, namely consistency in algorithm comparisons and reproducability. Due to computational constraints, novel methods are often only compared to the results provided in previous works. As a result, details such as the hyperparameter tuning procedure, learning rate schedules, or even the optimizer are often omitted or chosen to fit the algorithm at hand. From some of our experiments, we observe that simply fine-tuning a learning rate schedule for any of the methods from \Cref{tab:perfom} offers a bigger performance increase than choosing a better algorithm in the first place. As such, design choices can have a heavy impact on how well the algorithm performs. Having a common benchmarking procedure such as \domainbed, where these are fixed, is necessary to make \emph{substantial} progress in this field. We hope that we can push adoption in the community with our addition of \rsc and the methods proposed in this work. However, not using learning rate schedules and following the pre-defined distributions for learning rates, batch sizes, or weight decays might inherently bias this comparison and shouldn't be neglected as a factor.  

\begin{table*}[t]
\small
\centering
\begin{tabular}{llcccccc}
\toprule
\textbf{Algorithm} & \textbf{Ref.} & \textbf{Backbone} & \textbf{P} & \textbf{A} & \textbf{C} & \textbf{S} &  \textbf{Avg.} \\
\midrule
\textsc{Baseline}		&\cite{CarlucciDBCT19}				&	ResNet-18	&	$95.73$		&	$77.85$		&	$74.86$		&	$67.74$		&	$79.05$		 \\
\textsc{MASF}		&\cite{DouCKG19}					&	ResNet-18	&	$94.99$		&	$80.29$		&	$77.17$		&	$71.69$		&	$81.03$		 \\
\textsc{Epi-FCR}		&\cite{LiZYLSH19}					&	ResNet-18	&	$93.90$		&	$82.10$		&	$77.00$		&	$73.00$		&	$81.50$		 \\
\textsc{JiGen}		&\cite{CarlucciDBCT19}				&	ResNet-18	&	$96.03$		&	$79.42$		&	$75.25$		&	$71.35$		&	$80.51$		\\
\textsc{MetaReg}		& \cite{BalajiSC18}					&	ResNet-18	&	$95.50$		&	$83.70$		&	$77.20$		&	$70.30$		&	$81.70$		\\
\textsc{RSC} (reported)	& \cite{huang2020selfchallenging}		&	ResNet-18	&	$95.99$		&	$83.43$		&	$80.31$		&	$80.85$		&	$85.15$		\\
\textsc{RSC} (reproduced) & \cite{huang2020selfchallenging}		&	ResNet-18	&	$93.73$		&	$80.41$		&	$77.53$		&	$80.79$		&	$83.12$		\\
\divcams			&     (ours)						&	ResNet-18	&	$96.11$		&	$80.27$		&	$77.82$		&	$82.18$		&	$84.10$		\\
\bottomrule
\end{tabular}
\caption[Performance comparison for official PACS splits outside of \domainbed]{Performance comparison for PACS outside of the \domainbed framework with the official data split.}
\label{tab:official}
\end{table*}

On top of that, \Cref{tab:perfom} also shows the results for \dtransformers. Generally, we observe that many variants of the prototype based approaches outlined in \Cref{sec:prototypes} fail to generalize well to ResNet-50, even though they exhibit promising performance on ResNet-18 (see \Cref{tab:pacs_ours} for \prodrop results on PACS and ResNet-18). This might be due to the fact that prototypical approaches commonly require higher resolution feature maps \eg $14\times14$ in \citep{DoerschGZ20} via dilated convolutions. Nevertheless, \dtransformers already perform quite well for the benchmarking procedure outlined by \domainbed even without these additional changes. Keeping in mind that these adaptions can be made to push the performance even more, makes the approach even more suited for domain generalization although it is unclear how much the other algorithms would benefit from such a change. The value of prototypical approaches does not only lie in good performance but any prototypical approach can also offer a significant amount of explainability since these allow for visualizing the prototype similarity maps, the closest image patches to the prototypes, or even directly the prototypes if a sufficient decoder has jointly been trained. In particular, \dtransformers is able to achieve Top-2 performance in \domainbed for VLCS but performance seems to be not so good for the TerraIncognita dataset. We believe that this is because the dataset commonly includes only parts of the different animals at a very high distance, making it hard for the network to extract meaningful prototypes in the first place with the small feature resolution available. As such, it should be the hardest dataset for \dtransformers in \domainbed. In \Cref{tab:perfom} the results for \dtransformers on OfficeHome and DomainNet are omited due to computational constraints but we would expect the performances to be on-par if not better compared to the other methods, as these datasets do not share the same prototype extraction difficulties from TerraIncognita.

\section{Ablation Studies}
\label{sec:ablation}

In this section, we are going to look at different ablations of the presented methods and how the individual components impact the performance. In particular, for \divcams, we analyze different methods of resetting the masks (mask batching) in \Cref{sec:ablation_study_batching} and see how additional methods that are supposed to improve the underlying class activation maps impact performance in \Cref{sec:abl-masks}. On top of that, for \prodrop, we evaluate the effect of self-challenging for different negative weights in \Cref{sec:abl_self_challenging}, as well as the impact of the additional intra-loss factor in \Cref{sec:intra_loss}.

\subsection{Hyperparameter Distributions \& Schedules}
\label{sec:abl-distr}
For the mask batching ablation study we use \adam \cite{Kingma2015} and the distributions from \Cref{tab:abl-distributions-mask-batching}. When the batch drop factor is scheduled, we use an increasing linear schedule while the learning rate is always scheduled with a step-decay which decays the learning rate by factor $0.1$ at epoch $80/100$.
\begin{table}[t]
    \centering
    \begin{tabular}{lll}
        \toprule
         & \textbf{Hyperparameter} & \textbf{Distribution} \\
        \midrule
        $\alpha$ & learning rate & $\loguni{-5}{-1}$ \\
        $\mathcal{B}$ & batch size  & $\floor{\logunitwo{3}{9}}$ \\
        $\gamma$ & weight decay  & $\loguni{-6}{-2}$ \\
        $p$ & feature drop factor  & $1/3$ \\
        $b$ & batch drop factor  & $\uni{0}{1}$ \\
         \bottomrule 
    \end{tabular}
    \caption[Hyperparameters and distributions used for the mask batching ablation study]{Hyperparameters and distributions used in random search for the mask batching ablation study. $\logunix{a}{b}$ denotes a log-uniform distribution between $a$ and $b$ for base $x$, the uniform distribution is denoted as $\uni{a}{b}$ and $\floor{\cdot}$ is the floor operator.}
    \label{tab:abl-distributions-mask-batching}
\end{table}

For the mask ablation study, we use \adam \cite{Kingma2015} and the distributions from \Cref{tab:abl-distributions-mask}. When the batch drop factor is scheduled, we use an increasing linear schedule while the learning rate is \emph{not} scheduled. This corresponds to the tuning distributions provided in \domainbed which are also used for all the main results and all other ablations. If not marked otherwise, each experiment evaluates $20$ hyperparameter samples, similar to what is suggested in \domainbed.

\begin{table}[t]
\small
    \centering
    \begin{tabular}{lll}
        \toprule
        & \textbf{Hyperparameter} & \textbf{Distribution} \\
        \midrule
        $\alpha$ & learning rate & $\loguni{-5}{-3.5}$ \\
        $\mathcal{B}$ & batch size  & $\floor{\logunitwo{3}{5.5}}$ \\
        $\gamma$ & weight decay  & $\loguni{-6}{-2}$ \\
        $p$ & feature drop factor  & $\uni{0.2}{0.5}$ \\
        $b$ & batch drop factor  & $\uni{0}{1}$ \\
        $\lambda_1$ & hnc factor & $\loguni{-3}{-1}$ \\
        $k$ & negative classes & $\mathrm{num\_classes} - 1$ \\
        $\lambda_{tap}$ & tap factor & $\uni{0}{1}$ \\
        $\lambda_2$ & adversarial factor & $\loguni{-2}{2}$ \\
        $s$ &disc per gen step & $\floor{\logunitwo{0}{3}}$ \\
        $\eta$ & gradient penalty & $\loguni{-2}{1}$ \\
        $\omega_s$ & mlp width & $512$ \\
        $\omega_d$ & mlp depth & $3$ \\
        $\omega_{dr}$ & mlp dropout & $0.5$ \\
        $\lambda_3$ & mmd factor & $\loguni{-1}{1}$ \\
        \bottomrule 
    \end{tabular}
    \caption[Hyperparameters and distributions used for the mask ablation study]{Hyperparameters and distributions used in random search for the mask ablation study. $\logunix{a}{b}$ denotes a log-uniform distribution between $a$ and $b$ for base $x$, the uniform distribution is denoted as $\uni{a}{b}$, $\floor{\cdot}$ is the floor operator.}
    \label{tab:abl-distributions-mask}
\end{table}

\subsection{\divcam: Mask Batching}
\label{sec:ablation_study_batching}

There exist several methods how we can compute the vector $\mathbf{c}$ in our method and hence determine how we should apply the masks within each batch. Here, we analyze the effect on performance for a few possible choices. By default, \divcam uses \Cref{eq:conf_scamb} where $y_{gt}$ is the confidence on the ground truth class after softmax. This applies the masks on samples with the highest confidence on the correct class within each batch.   
\begin{equation}
\label{eq:conf_scamb}
	\mathbf{c}^n = y_{gt}
\end{equation}
Another option is \divcamc with \Cref{eq:conf_scamc} which computes the change in confidence on the ground truth class when applying the mask. The masked confidence after softmax is denoted as $\tilde{y}_{gt}$. This variation applies the masks for samples where the mask decreases confidence on the ground truth class the most.
\begin{equation}
\label{eq:conf_scamc}
   \mathbf{c}^n = y_{gt} - \tilde{y}_{gt}
\end{equation}
The last variation is \divcamt where we apply the masks randomly for samples which are correctly classified. All variants can further be extended by adding a linear schedule, denoted with an additional ``S'', or computing $\mathbf{c}$ for each domain separately, denoted with an additional ``D''. By adding a schedule, we apply masks more in the later training epochs where discriminant features have been learned and by enforcing it for each domain we can ensure that the masks aren't applied biased towards a subset of domains and disregarded for others. \Cref{tab:scam_batching} shows experiments for these variants.

\begin{table*}[t]
\small
    \centering
    \begin{tabular}{lccccc}
    \toprule
    \textbf{Name} &  \textbf{P} & \textbf{A} & \textbf{C} & \textbf{S} &  \textbf{Avg.} \\
    \midrule
    \divcam & $94.0\pm0.4$ & $80.6\pm1.2$ & $75.4\pm0.7$ & $76.7\pm0.7$ & $81.7\pm0.6$ \\
    \divcams & $94.4\pm0.7$ & $80.5\pm0.4$ & $74.6\pm2.2$ & $\tabtop{79.0\pm0.9}$ & $\tabtop{82.1\pm0.3}$   \\
    \divcamd & $94.3\pm0.1$ & $80.1\pm0.1$ & $74.5\pm0.9$ & $76.6\pm1.7$ & $81.4\pm0.2$   \\
    \divcamds & $93.9\pm0.2$ & $80.4\pm0.4$ & $73.4\pm2.2$ & $74.8\pm1.2$ & $80.6\pm0.9$   \\
    \divcamc & $92.6\pm0.4$ & $80.1\pm1.1$ & $73.6\pm1.4$ & $75.0\pm1.2$ & $80.3\pm0.9$  \\
    \divcamcs & $95.0\pm0.6$ & $79.9\pm1.0$ & $74.5\pm0.7$ & $78.1\pm0.8$ & $81.9\pm0.4$   \\
    \divcamdc & $\tabtop{95.1\pm0.4}$ & $79.5\pm1.0$ & $73.7\pm0.9$ & $75.2\pm1.2$ & $80.9\pm0.4$ \\
    \divcamdcs & $93.5\pm0.1$ & $80.1\pm0.2$ & $75.1\pm0.1$ & $77.2\pm1.6$ & $81.5\pm0.5$  \\
    \divcamt & $95.0\pm0.3$ & $80.3\pm0.3$ & $74.8\pm0.8$ & $75.3\pm1.1$ & $81.4\pm0.4$  \\
    \divcamts & $95.0\pm0.1$ & $79.9\pm0.8$ & $72.6\pm1.3$ & $77.1\pm1.4$ & $81.2 \pm 0.4$  \\
    \divcamdt & $94.8\pm0.6$ & $79.6\pm0.6$ & $74.0\pm1.1$ & $78.5\pm0.4$ & $81.7\pm0.1$  \\
    \divcamdts & $95.1\pm0.2$ & $\tabtop{81.5\pm1.3}$ & $\tabtop{75.5\pm0.4}$ & $74.9\pm2.0$ &  $81.7\pm0.5$  \\
    \midrule
    \tdivcam & $\tabtop{94.9\pm0.7}$ & $81.5\pm0.7$ & $76.6\pm0.4$ & $\tabtop{80.5\pm0.7}$ & $83.4\pm0.3$  \\
    \tdivcams & $94.9\pm0.3$ & $\tabtop{82.7\pm0.7}$ & $76.3\pm0.7$ & $80.1\pm0.4$ & $83.5\pm0.3$   \\
    \tdivcamd & $94.8\pm0.2$ & $81.0\pm0.7$ & $\tabtop{77.6\pm0.6}$ & $79.9\pm0.6$ &  $83.3\pm0.3$  \\
    \tdivcamds & $94.6\pm0.5$ & $80.7\pm0.3$ & $77.0\pm0.4$ & $79.3\pm0.3$ & $82.9\pm0.1$   \\
    \tdivcamc & $94.7\pm0.5$ & $82.6\pm0.6$ & $77.0\pm0.5$ & $80.1\pm1.0$ & $\tabtop{83.6\pm0.3}$  \\
    \tdivcamcs & $94.2\pm0.2$ & $82.5\pm0.8$ & $76.9\pm0.3$ & $79.9\pm0.7$ & $83.4\pm0.3$  \\
    \tdivcamdc & $94.8\pm0.4$ & $82.0\pm0.4$ & $76.6\pm0.9$ & $80.1\pm0.4$ & $83.4\pm0.1$  \\
    \tdivcamdcs & $94.7\pm0.4$ & $81.0\pm0.3$ & $77.6\pm0.2$ & $80.3\pm1.3$ & $83.4\pm0.3$ \\
    \tdivcamt & $94.5\pm0.4$ & $81.6\pm0.8$ & $76.7\pm0.2$ & $79.6\pm0.4$ & $83.1\pm0.4$ \\
    \tdivcamts & $94.8\pm0.3$ & $81.3\pm0.2$ & $76.7\pm0.5$ & $79.7\pm0.5$  & $83.2 \pm 0.2$  \\
    \tdivcamdt & $94.7\pm0.5$ & $80.9\pm1.1$ & $77.3\pm0.5$ & $79.9\pm0.6$ & $83.2\pm0.2$ \\
    \tdivcamdts & $94.7\pm0.5$ & $82.1\pm1.0$ & $76.4\pm0.6$ & $79.5\pm1.2$ & $83.2\pm0.1$  \\
    \bottomrule
    \end{tabular}
    \caption[Ablation study for the \divcam mask batching on the PACS dataset]{Ablation study for the \divcam mask batching on the PACS dataset using training-domain validation (top) and oracle validation denoted with * (bottom). We use a ResNet-18 backbone, schedules and distributions from \Cref{sec:abl-distr}, $25$ hyperparameter samples, and $3$ split seeds for standard deviations.}
    \label{tab:scam_batching}
\end{table*}

We observe that adding a schedule helps in most cases, achieving the highest training domain validation performance for \divcams.  Enforcing the application of masks within each domain, however, doesn't consistently improve performance and therefore we don't consider it for the final method.

\subsection{\divcam: Class Activation Maps}
\label{sec:abl-masks}

We combine our class activation maps with other methods from domain generalization, as well as methods to boost the explainability for class activation maps from the weakly-supervised object localization literature. The results are shown in \Cref{tab:scam_masks} where MAP + CDANN drops the self-challenging part from \divcam and just computes ordinary cross entropy while aligning the class activation maps.

We observe that, surprisingly, none of the methods have a positive effect for training domain validation even though some of them exhibit better performance for oracle validation. Notably, especially the CDANN approach tends to exhibit a high 
standard deviation for some of the domains (\eg art) which suggests that this approach can be fine-tuned when only reporting performance on a single seed. However, since we are looking for a method which reliably provides competitive results, regardless of the used seed and without needing extensive fine-tuning, we disregard this option.
\begin{table}[t]
\small
    \centering
    \begin{tabular}{lccccc}
    \toprule
    \textbf{Name} &  \textbf{P} & \textbf{A} & \textbf{C} & \textbf{S} & \textbf{Avg.} \\
    \midrule 
    \divcam & $\tabtop{97.6\pm0.4}$ & $85.2\pm0.8$ & $80.5\pm0.7$ & $78.3\pm0.8$ & $\tabtop{85.4\pm0.5}$ \\
    \divcams & $97.3\pm0.4$ & $86.2\pm1.4$ & $79.1\pm2.2$ & $\tabtop{79.2\pm0.1}$ & $85.4\pm0.2$ \\
    \divcams + TAP & $96.9\pm0.1$ & $85.1\pm1.5$ & $78.7\pm0.4$ & $75.3\pm0.6$ & $84.0\pm0.4$ \\
    \divcams + HNC & $97.2\pm0.3$ & $\tabtop{87.2\pm0.9}$ & $79.2\pm0.6$ & $71.7\pm3.1$ & $83.8\pm0.4$ \\
    \divcams + CDANN & $97.5\pm0.4$ & $85.2\pm2.8$ & $78.3\pm2.0$ & $74.8\pm0.9$ & $84.0\pm1.5$ \\
    \divcams + MMD & $97.0\pm0.2$ & $85.4\pm1.0$ & $\tabtop{81.5\pm0.4}$ & $75.8\pm3.5$ & $84.9\pm1.1$ \\
    CAM + CDANN & $97.2\pm0.3$ & $86.7\pm0.5$ & $77.3\pm1.7$ & $71.5\pm1.3$ & $83.2\pm0.8$ \\
    \midrule
    \tdivcam & $96.2\pm1.2$ & $87.0\pm0.5$ & $82.0\pm0.9$ & $80.8\pm0.6$ & $86.5\pm0.1$ \\
    \tdivcams & $97.2\pm0.3$ & $86.5\pm0.4$ & $83.0\pm0.5$ & $82.2\pm0.1$ & $87.2\pm0.1$ \\
    \divcams + TAP* & $97.3\pm0.3$ & $87.2\pm0.8$ & $\tabtop{83.2\pm0.8}$ & $\tabtop{82.8\pm0.2}$ & $\tabtop{87.6\pm0.0}$ \\
    \divcams + HNC*  & $97.3\pm0.2$ & $\tabtop{87.4\pm0.5}$ & $81.4\pm0.6$ & $79.7\pm1.1$ & $86.5\pm0.4$ \\
    \divcams + CDANN* & $\tabtop{97.3\pm0.5}$ & $85.9\pm1.2$ & $80.6\pm0.4$ & $80.9\pm0.4$ & $86.2\pm0.2$ \\
    \divcams + MMD* & $\tabtop{97.3\pm0.5}$ & $86.8\pm0.7$ & $83.2\pm0.4$ & $80.9\pm0.7$ & $87.1\pm0.4$ \\
    CAM + CDANN* & $97.2\pm0.4$ & $86.7\pm0.5$ & $81.9\pm0.2$ & $80.6\pm0.7$ & $86.6\pm0.2$ \\
    \bottomrule
    \end{tabular}
    \caption[Ablation study for the \divcam masks on the PACS dataset]{Ablation study for the \divcam masks on the PACS dataset using training-domain validation (top) and oracle validation denoted with * (bottom). We use a ResNet-50 backbone, schedules and distributions from \Cref{sec:abl-distr}, $20$ hyperparameter samples, and $3$ split seeds for standard deviations. Results are directly integratable in \Cref{tab:perfom} as we use the same tuning protocol provided in \domainbed.}
    \label{tab:scam_masks}
\end{table}

\subsection{\prodrop: Self-Challenging}
\label{sec:abl_self_challenging}

\Cref{tab:scabl} shows the ablation results for the self-challenging addition. We observe that for most negative weights, adding self-challenging results in an performance increase. Most notably, this occurs for cases where the performance without self-challenging is very poor such as $w_{c,j} = -0.2\; \forall j: \prot \notin \prots_c$,  $w_{c,j} = -1.0\; \forall j: \prot \notin \prots_c$, or $w_{c,j} = 0.0\; \forall j: \prot \notin \prots_c$. Generally, in cases where it does not lead to a performance increase, the downside seems to be very small where we at most drop by 0.5\% performance for $w_{c,j} = -0.5\; \forall j: \prot \notin \prots_c$. 

If we look at the performance changes solely based on the different negative weights, we can't observe any consistent trends. In fact, it is very surprising that small changes such as from  $w_{c,j} = -0.1$ to $w_{c,j} = -0.2$ without self-challenging can lead to a 2\% performance change. 

\begin{table}[t]
    \centering
    \begin{tabular}{lcccccc}
    \toprule
    \textbf{Weight}  & \textbf{SC} & \textbf{P} & \textbf{A} & \textbf{C} & \textbf{S} & \textbf{Avg.} \\
     \midrule
     \phantom{-}0.0 & \ding{55} & 93.2 $\pm$ 0.0 & 80.4 $\pm$ 1.0 & 73.7 $\pm$ 0.4 & 72.6 $\pm$ 2.6 & 80.0 $\pm$ 0.7 \\
     -0.1 & \ding{55} & 94.3 $\pm$ 0.3 & 78.8 $\pm$ 0.3 & 74.4 $\pm$ 0.7 & 75.3 $\pm$ 1.6 & 80.7 $\pm$ 0.4 \\
     -0.2 & \ding{55} & 93.2 $\pm$ 0.5 & 76.8 $\pm$ 1.5 & 72.9 $\pm$ 0.1 & 71.8 $\pm$ 1.0 & 78.7 $\pm$ 0.5 \\
     -0.3 & \ding{55} & 94.0 $\pm$ 0.4 & 79.8 $\pm$ 1.0 & 75.6 $\pm$ 1.5 & 73.9 $\pm$ 1.0 & 80.8 $\pm$ 0.1 \\
     -0.4 & \ding{55} & 93.6 $\pm$ 0.1 & 79.8 $\pm$ 0.5 & 74.3 $\pm$ 1.3 & 75.7 $\pm$ 2.3 & 80.8 $\pm$ 0.5 \\
     -0.5 & \ding{55} & 93.0 $\pm$ 0.9 & 79.4 $\pm$ 1.6 & 73.2 $\pm$ 0.9 & 75.5 $\pm$ 1.0 & 80.3 $\pm$ 0.4 \\
     -1.0 & \ding{55} & 94.2 $\pm$ 0.4 & 80.2 $\pm$ 1.1 & 72.7 $\pm$ 1.4 & 68.6 $\pm$ 0.6 & 78.9 $\pm$ 0.2 \\
     -2.0 & \ding{55} & $\tabtop{94.7 \pm 0.4}$ & 78.7 $\pm$ 0.5 & 75.5 $\pm$ 1.0 & 71.1 $\pm$ 2.2 & 80.0 $\pm$ 0.7 \\
     \phantom{-}0.0 $\to$ -1.0 & \ding{55} & 94.3 $\pm$ 0.5 & 79.5 $\pm$ 0.6 & 74.2 $\pm$ 0.3 & 72.2 $\pm$ 2.1 & 80.0 $\pm$ 0.4 \\
      \midrule
      \phantom{-}0.0 & \ding{51} & 93.4 $\pm$ 0.6 & 80.5 $\pm$ 0.8 & 75.6 $\pm$ 0.1 & 74.3 $\pm$ 2.0 & 81.0 $\pm$ 0.4 \\
     -0.1 & \ding{51} & 93.7 $\pm$ 0.3 & $\tabtop{83.2 \pm 1.4}$ & 75.9 $\pm$ 1.2 & 71.1 $\pm$ 1.9 & 81.0 $\pm$ 0.8 \\
     -0.2 & \ding{51} & 93.2 $\pm$ 0.2 & 81.0 $\pm$ 0.7 & 73.9 $\pm$ 0.7 & 75.0 $\pm$ 0.6 & 80.8 $\pm$ 0.1 \\
     -0.3 & \ding{51} & 93.4 $\pm$ 0.8 & 81.4 $\pm$ 0.4 & 71.3 $\pm$ 1.2 & 76.9 $\pm$ 0.9 & 80.7 $\pm$ 0.2 \\
     -0.4 & \ding{51} & 94.0 $\pm$ 0.3 & 81.7 $\pm$ 0.9 & 72.9 $\pm$ 0.4 & 73.5 $\pm$ 1.0 & 80.5 $\pm$ 0.2 \\
     -0.5 & \ding{51} & 93.5 $\pm$ 0.7 & 80.7 $\pm$ 1.6 & 71.6 $\pm$ 1.4 & 73.6 $\pm$ 1.5 & 79.8 $\pm$ 0.9 \\
     -1.0 & \ding{51} & 94.6 $\pm$ 0.2 & 81.6 $\pm$ 1.2 & 72.9 $\pm$ 0.4 & $\tabtop{77.0 \pm 1.6}$ & $\tabtop{81.5 \pm 0.2}$ \\
     -2.0 & \ding{51} & 94.0 $\pm$ 0.5 & 79.5 $\pm$ 1.2  & $\tabtop{76.4 \pm 0.4 }$ & 73.9 $\pm$ 1.7  & 80.9 $\pm$ 0.6 \\
     \phantom{-}0.0 $\to$ -1.0 & \ding{51} & 94.1 $\pm$ 0.4 & 79.2 $\pm$ 0.6 & 74.1 $\pm$ 1.1 & 71.9 $\pm$ 0.2 & 79.8 $\pm$ 0.3  \\
    \bottomrule
    \end{tabular}
    \caption[Self-challenging performance comparison for different negative class weights]{Performance comparison for different negative class weights on the PACS dataset without (top) and with self-challenging (bottom)  using training-domain validation and a ResNet-18 backbone. The feature and batch drop factors are kept constant with $p=0.5$ and $b=\frac{1}{3}$, a linear schedule from $a$ to $b$ throughout training is denoted with $a \to b$.}
    \label{tab:scabl}
\end{table}

\subsection{\prodrop: Intra-Loss}
\label{sec:intra_loss}

\Cref{tab:intrabl} shows the ablation results for different intra factor strengths $\lambda_6$ with and without self-challenging. As expected, if the intra factor grows too large, performance degrades. For smaller values of $\lambda_6$ without self-challenging, we can observe a consistent performance increase of varying degrees similar to what self-challenging is able to gain.

Even though we experimented with different weighting of the individual distance metrics as well as the overall loss, we observe no \emph{consistent} performance improvements on top of self-challenging across testing environments and data splits. The observations from \Cref{sec:intra_loss}, \Cref{fig:pw_distance_trial1-sc}, and \Cref{sec:additional_distances} suggest that self-challenging inherently already enforces the desired properties up to the near-optimal extent such that the additional loss term does not provide any consistent further benefits.

\begin{table}[ht]
    \centering
    \begin{tabular}{lcccccc}
    \toprule
    Intra factor \textbf{$\lambda_6$}  & \textbf{SC} & \textbf{P} & \textbf{A} & \textbf{C} & \textbf{S} & \textbf{Avg.} \\
     \midrule
     \phantom{-}0.0 & \ding{55} & 94.2 $\pm$ 0.4 & 80.2 $\pm$ 1.1 & 72.7 $\pm$ 1.4 & 68.6 $\pm$ 0.6 & 78.9 $\pm$ 0.2 \\
     -0.1 & \ding{55} & 94.1 $\pm$ 0.4 & 81.2 $\pm$ 1.2 & 73.6 $\pm$ 0.8 & 75.2 $\pm$ 2.4 & 81.0 $\pm$ 0.6 \\
     -0.2 & \ding{55} & 94.5 $\pm$ 0.4 & 81.5 $\pm$ 1.1 & 74.4 $\pm$ 1.6 & 74.4 $\pm$ 1.1 & 81.2 $\pm$ 0.5 \\
     -0.5 & \ding{55} & 92.9 $\pm$ 0.8 & 82.4 $\pm$ 0.4 & 73.5 $\pm$ 1.6 & 73.4 $\pm$ 2.0 & 80.6 $\pm$ 0.7 \\
     -1.0 & \ding{55} & 94.7 $\pm$ 0.4 & 80.4 $\pm$ 0.6 & 73.9 $\pm$ 0.9 & 75.2 $\pm$ 1.8 & 81.1 $\pm$ 0.7 \\
      \midrule
      \phantom{-}0.0 & \ding{51} & 94.6 $\pm$ 0.2 & 81.6 $\pm$ 1.2 & 72.9 $\pm$ 0.4 & $\tabtop{77.0 \pm 1.6}$ & $81.5 \pm 0.2$ \\
      -0.1 & \ding{51} & 94.6 $\pm$ 0.3 & 82.6 $\pm$ 0.9 & 72.2 $\pm$ 0.5 & 75.4 $\pm$ 0.1 & 81.2 $\pm$ 0.3 \\
      -0.2 & \ding{51} & 94.1 $\pm$ 0.1 & 81.9 $\pm$ 0.2 & 73.3 $\pm$ 0.7 & 75.8 $\pm$ 0.9 & 81.2 $\pm$ 0.1 \\
      -0.3 & \ding{51} & 94.8 $\pm$ 0.2 & 81.5 $\pm$ 0.1 & 75.1 $\pm$ 0.0 & 74.5 $\pm$ 0.4 & 81.5 $\pm$ 0.2 \\
      -0.4 & \ding{51} & 93.9 $\pm$ 0.5 & 82.4 $\pm$ 0.2 & 74.3 $\pm$ 1.3 & 76.1 $\pm$ 1.1 & 81.7 $\pm$ 0.2 \\
      -0.5 & \ding{51} & 93.6 $\pm$ 0.6 & 82.1 $\pm$ 0.9 & $\tabtop{76.4 \pm 0.9}$ & 76.3 $\pm$ 0.6 & $\tabtop{82.1 \pm 0.6}$ \\
      -0.6 & \ding{51} & 93.8 $\pm$ 0.6 & 82.1 $\pm$ 0.3 & 75.7 $\pm$ 0.2 & 73.1 $\pm$ 3.1 & 81.2 $\pm$ 1.0 \\
      -0.7 & \ding{51} & 94.0 $\pm$ 0.4 & $\tabtop{82.9 \pm 1.2}$ & 74.1 $\pm$ 0.5 & 76.3 $\pm$ 1.0 & $81.8 \pm 0.4$ \\
      -0.8 & \ding{51} & 94.1 $\pm$ 0.5 & 81.8 $\pm$ 0.3 & 75.5 $\pm$ 0.2 & 72.7 $\pm$ 0.3 & $81.0 \pm 0.1$ \\
      -1.0 & \ding{51} & $\tabtop{95.1 \pm 0.6}$ & 80.7 $\pm$ 1.5 & 74.5 $\pm$ 1.1 & 73.7 $\pm$ 1.2 & 81.0 $\pm$ 0.4 \\
      -2.0 & \ding{51} & 94.3 $\pm$ 0.3 & $\tabtop{82.9 \pm 0.5}$ & 75.1 $\pm$ 1.0 & 75.0 $\pm$ 1.1 & 81.8 $\pm$ 0.5 \\
      -3.0 & \ding{51} & 94.5 $\pm$ 0.3 & 79.8 $\pm$ 1.0 & 74.9 $\pm$ 1.0 & 71.2 $\pm$ 0.4 & 80.1 $\pm$ 0.3 \\
      -10.0 & \ding{51} & 90.8 $\pm$ 3.7 & 80.2 $\pm$ 0.3 & 72.9 $\pm$ 0.1 & 75.0 $\pm$ 1.8 & 79.7 $\pm$ 0.6 \\
      -100.0 & \ding{51} & 72.0 $\pm$ 11. & 76.3 $\pm$ 2.2 & 73.0 $\pm$ 1.1 & 69.1 $\pm$ 2.6 & 72.6 $\pm$ 4.0 \\
    \bottomrule
    \end{tabular}
    \caption[Self-challenging performance comparison for different intra factors]{Performance comparison for different intra factors on the PACS dataset without (top) and with self-challenging (bottom)  using training-domain validation and a ResNet-18 backbone. The feature and batch drop factors are kept constant with $p=0.5$ and $b=\frac{1}{3}$, negative weight is $-1.0$. Distance metrics are weighted with $\lambda_{\ell_2} = 1$ and $\lambda_\cdistance = 1$.}
    \label{tab:intrabl}
\end{table}
\chapter{Conclusion and Outlook}

In this work, we investigated if we can deploy explainability methods during the training procedure and gain, both, better performance on the domain generalization task as well as a framework that enables more explainability for the users. In particular, we develop a regularization technique based on class activation maps that visualize parts of an image responsible for certain predictions (\divcam) as well as prototypical representations that serve as a number of class or attribute centroids which the network uses to make its predictions (\prodrop and \dtransformers).

From the results and ablations presented in this work, we have shown that especially \divcam is a reliable method for achieving domain generalization for small to medium sized ResNet backbones while offering an architecture that allows for additional insights into \emph{how} and \emph{why} the network arrives at it's predictions. Depending on the backbone and dataset, \prodrop and \dtransformers can also be powerful options for this cause. The possibilities for explainability in these methods is a property that is highly desirable in practice, especially for safety-critical scenarios such as self-driving cars, any application in the medical field \eg cancer or tumor prediction, or any other automation robot that needs to operate in a diverse set of environments. We hope that the presented methods can find application in such scenarios and establish additional trust and confidence into the machine learning systems to work reliable.

Building upon the methods presented in this work, there are also quite a number of ablations or extension points that might be interesting to investigate. First, even though the results for \prodrop looked very promising on ResNet-18, it failed to generalize well enough to ResNet-50 to properly compete with the other methods in \domainbed. Looking into some of the details coming from this transition, might yield a better understanding of what causes this problem and a solution can probably be found. Secondly, even though the explainability methods we build upon have very deep roots in the explainability literature, it would be interesting to either jointly train a suitable decoder for the prototypes or visualize the closest latent patch across the whole dataset. Since we're training with images coming from different domains, there could be interesting visualizations possible, potentially also in a video format that shows the change throughout training. In this work, we especially focused on achieving good performance with these methods rather than the explainability itself. Thirdly, many prototypical networks upscale the feature map in size, for example to $14 \times 14$ from $7 \times 7$, and report great performance gains coming from the increased spatial resolution of the latent representation \citep{DoerschGZ20}. We deliberately refrain from changing the backbone in such a way to improve comparability in the \domainbed framework without needing to re-compute the extensive set of baselines, although it might be possible that both \prodrop and \dtransformers benefit more heavily from such a change compared to other methods. In a similar fashion, many prototypical networks use the euclidean distance as a distance measure while some works report better performance for the dot or cosine similarity \citep{XuXWSA20}. We experimented with a few options across the different methods but a clear ablation study of this detail for the domain generalization task would be very helpful. In particular, one can also think about deploying any other Bregman divergence measure and taking a metric learning approach with, for example, the Mahalanobis distance \citep{BanerjeeMDG04} that might be able to achieve additional performance gains. 

Finally, especially for \dtransformers, our method of aggregating across multiple environments is very simple in nature and finding a more elaborate way to utilize the multiple aligned prototypes for each environment might be a great opportunity for a follow-up work. Nevertheless, we hope that our analysis might serve as a first stepping stone and groundwork for developing more domain generalization algorithms based on explainability methods.


\printbibliography[heading=bibintoc]


\appendix 



\chapter{Domain-specific results} 
\label{sec:DomainResults} 

Since \Cref{tab:perfom} only shows the average performance for the respective dataset, we provide the full experimental results here for each of the domains across datasets. The results for the other algorithms besides \textsc{RSC}, \divcam, and \dtransformers are taken from \domainbed.

\begin{table*}
\begin{center}
\begin{tabular}{lccccc}
\toprule
\textbf{Algorithm}   & \textbf{C}           & \textbf{L}           & \textbf{S}           & \textbf{V}           & \textbf{Avg.}         \\
\midrule
ERM                  & 97.7 $\pm$ 0.4       & 64.3 $\pm$ 0.9       & 73.4 $\pm$ 0.5       & 74.6 $\pm$ 1.3       & 77.5                 \\
IRM                  & 98.6 $\pm$ 0.1       & 64.9 $\pm$ 0.9       & 73.4 $\pm$ 0.6       & 77.3 $\pm$ 0.9       & 78.5                 \\
GroupDRO             & 97.3 $\pm$ 0.3       & 63.4 $\pm$ 0.9       & 69.5 $\pm$ 0.8       & 76.7 $\pm$ 0.7       & 76.7                 \\
Mixup                & 98.3 $\pm$ 0.6       & 64.8 $\pm$ 1.0       & 72.1 $\pm$ 0.5       & 74.3 $\pm$ 0.8       & 77.4                 \\
MLDG                 & 97.4 $\pm$ 0.2       & 65.2 $\pm$ 0.7       & 71.0 $\pm$ 1.4       & 75.3 $\pm$ 1.0       & 77.2                 \\
CORAL                & 98.3 $\pm$ 0.1       & 66.1 $\pm$ 1.2       & 73.4 $\pm$ 0.3       & 77.5 $\pm$ 1.2       & 78.8                 \\
MMD                  & 97.7 $\pm$ 0.1       & 64.0 $\pm$ 1.1       & 72.8 $\pm$ 0.2       & 75.3 $\pm$ 3.3       & 77.5                 \\
DANN                 & 99.0 $\pm$ 0.3       & 65.1 $\pm$ 1.4       & 73.1 $\pm$ 0.3       & 77.2 $\pm$ 0.6       & 78.6                 \\
CDANN                & 97.1 $\pm$ 0.3       & 65.1 $\pm$ 1.2       & 70.7 $\pm$ 0.8       & 77.1 $\pm$ 1.5       & 77.5                 \\
MTL                  & 97.8 $\pm$ 0.4       & 64.3 $\pm$ 0.3       & 71.5 $\pm$ 0.7       & 75.3 $\pm$ 1.7       & 77.2                 \\
SagNet               & 97.9 $\pm$ 0.4       & 64.5 $\pm$ 0.5       & 71.4 $\pm$ 1.3       & 77.5 $\pm$ 0.5       & 77.8                 \\
ARM                  & 98.7 $\pm$ 0.2       & 63.6 $\pm$ 0.7       & 71.3 $\pm$ 1.2       & 76.7 $\pm$ 0.6       & 77.6                 \\
VREx                 & 98.4 $\pm$ 0.3       & 64.4 $\pm$ 1.4       & 74.1 $\pm$ 0.4       & 76.2 $\pm$ 1.3       & 78.3                 \\
RSC                  & 97.9 $\pm$ 0.1       & 62.5 $\pm$ 0.7       & 72.3 $\pm$ 1.2       & 75.6 $\pm$ 0.8       & 77.1                 \\
\divcams 	   & 98.7 $\pm$ 0.1       & 64.5 $\pm$ 1.1        & 72.5 $\pm$ 0.7       & 75.5 $\pm$ 0.4       & 77.8                \\
\dtransformers & 98.1 $\pm$ 0.2 & 65.8 $\pm$ 0.6 & 71.7 $\pm$ 0.4 & 79.2 $\pm$ 1.3 & 78.7 \\
\midrule
ERM*                  & 97.6 $\pm$ 0.3       & 67.9 $\pm$ 0.7       & 70.9 $\pm$ 0.2       & 74.0 $\pm$ 0.6       & 77.6                 \\
IRM*                  & 97.3 $\pm$ 0.2       & 66.7 $\pm$ 0.1       & 71.0 $\pm$ 2.3       & 72.8 $\pm$ 0.4       & 76.9                 \\
GroupDRO*             & 97.7 $\pm$ 0.2       & 65.9 $\pm$ 0.2       & 72.8 $\pm$ 0.8       & 73.4 $\pm$ 1.3       & 77.4                 \\
Mixup*                & 97.8 $\pm$ 0.4       & 67.2 $\pm$ 0.4       & 71.5 $\pm$ 0.2       & 75.7 $\pm$ 0.6       & 78.1                 \\
MLDG*                 & 97.1 $\pm$ 0.5       & 66.6 $\pm$ 0.5       & 71.5 $\pm$ 0.1       & 75.0 $\pm$ 0.9       & 77.5                 \\
CORAL*                & 97.3 $\pm$ 0.2       & 67.5 $\pm$ 0.6       & 71.6 $\pm$ 0.6       & 74.5 $\pm$ 0.0       & 77.7                 \\
MMD*                  & 98.8 $\pm$ 0.0       & 66.4 $\pm$ 0.4       & 70.8 $\pm$ 0.5       & 75.6 $\pm$ 0.4       & 77.9                 \\
DANN*                 & 99.0 $\pm$ 0.2       & 66.3 $\pm$ 1.2       & 73.4 $\pm$ 1.4       & 80.1 $\pm$ 0.5       & 79.7                 \\
CDANN*                & 98.2 $\pm$ 0.1       & 68.8 $\pm$ 0.5       & 74.3 $\pm$ 0.6       & 78.1 $\pm$ 0.5       & 79.9                 \\
MTL*                  & 97.9 $\pm$ 0.7       & 66.1 $\pm$ 0.7       & 72.0 $\pm$ 0.4       & 74.9 $\pm$ 1.1       & 77.7                 \\
SagNet*               & 97.4 $\pm$ 0.3       & 66.4 $\pm$ 0.4       & 71.6 $\pm$ 0.1       & 75.0 $\pm$ 0.8       & 77.6                 \\
ARM*                  & 97.6 $\pm$ 0.6       & 66.5 $\pm$ 0.3       & 72.7 $\pm$ 0.6       & 74.4 $\pm$ 0.7       & 77.8                 \\
VREx*                 & 98.4 $\pm$ 0.2       & 66.4 $\pm$ 0.7       & 72.8 $\pm$ 0.1       & 75.0 $\pm$ 1.4       & 78.1                 \\
RSC*                  & 98.0 $\pm$ 0.4       & 67.2 $\pm$ 0.3       & 70.3 $\pm$ 1.3       & 75.6 $\pm$ 0.4       & 77.8                 \\
\tdivcams 	   & 98.0 $\pm$ 0.5       & 66.1 $\pm$ 0.3        & 72.0 $\pm$ 1.0       & 76.4 $\pm$ 0.7        & 78.1                 \\
\tdtransformers & 98.1 $\pm$ 0.3 & 66.5 $\pm$ 0.8 & 72.4 $\pm$ 0.9 & 74.0 $\pm$ 0.4 & 77.7 \\
\bottomrule
\end{tabular}
\caption[Domain specific performance for the VLCS dataset]{Domain specific performance for the VLCS dataset using training-domain validation (top) and  oracle validation denoted with * (bottom). We use a ResNet-50 backbone, optimize with \adam, and follow the distributions specified in \domainbed. Only \rsc and our methods have been added as part of this work, the other baselines are taken from \domainbed.}
\end{center}
\end{table*}

\begin{table*}
\begin{center}
\begin{tabular}{lccccc}
\toprule
\textbf{Algorithm}   & \textbf{A}           & \textbf{C}           & \textbf{P}           & \textbf{S}           & \textbf{Avg.}         \\
\midrule
ERM                  & 84.7 $\pm$ 0.4       & 80.8 $\pm$ 0.6       & 97.2 $\pm$ 0.3       & 79.3 $\pm$ 1.0       & 85.5                 \\
IRM                  & 84.8 $\pm$ 1.3       & 76.4 $\pm$ 1.1       & 96.7 $\pm$ 0.6       & 76.1 $\pm$ 1.0       & 83.5                 \\
GroupDRO             & 83.5 $\pm$ 0.9       & 79.1 $\pm$ 0.6       & 96.7 $\pm$ 0.3       & 78.3 $\pm$ 2.0       & 84.4                 \\
Mixup                & 86.1 $\pm$ 0.5       & 78.9 $\pm$ 0.8       & 97.6 $\pm$ 0.1       & 75.8 $\pm$ 1.8       & 84.6                 \\
MLDG                 & 85.5 $\pm$ 1.4       & 80.1 $\pm$ 1.7       & 97.4 $\pm$ 0.3       & 76.6 $\pm$ 1.1       & 84.9                 \\
CORAL                & 88.3 $\pm$ 0.2       & 80.0 $\pm$ 0.5       & 97.5 $\pm$ 0.3       & 78.8 $\pm$ 1.3       & 86.2                 \\
MMD                  & 86.1 $\pm$ 1.4       & 79.4 $\pm$ 0.9       & 96.6 $\pm$ 0.2       & 76.5 $\pm$ 0.5       & 84.6                 \\
DANN                 & 86.4 $\pm$ 0.8       & 77.4 $\pm$ 0.8       & 97.3 $\pm$ 0.4       & 73.5 $\pm$ 2.3       & 83.6                 \\
CDANN                & 84.6 $\pm$ 1.8       & 75.5 $\pm$ 0.9       & 96.8 $\pm$ 0.3       & 73.5 $\pm$ 0.6       & 82.6                 \\
MTL                  & 87.5 $\pm$ 0.8       & 77.1 $\pm$ 0.5       & 96.4 $\pm$ 0.8       & 77.3 $\pm$ 1.8       & 84.6                 \\
SagNet               & 87.4 $\pm$ 1.0       & 80.7 $\pm$ 0.6       & 97.1 $\pm$ 0.1       & 80.0 $\pm$ 0.4       & 86.3                 \\
ARM                  & 86.8 $\pm$ 0.6       & 76.8 $\pm$ 0.5       & 97.4 $\pm$ 0.3       & 79.3 $\pm$ 1.2       & 85.1                 \\
VREx                 & 86.0 $\pm$ 1.6       & 79.1 $\pm$ 0.6       & 96.9 $\pm$ 0.5       & 77.7 $\pm$ 1.7       & 84.9                 \\
RSC                  & 85.4 $\pm$ 0.8       & 79.7 $\pm$ 1.8       & 97.6 $\pm$ 0.3       & 78.2 $\pm$ 1.2       & 85.2                 \\
\divcams 	        & 86.2 $\pm$ 1.4       & 79.1 $\pm$ 2.2        & 97.3 $\pm$ 0.4       & 79.2 $\pm$ 0.1       & 85.4                \\
\dtransformers & 86.9 $\pm$ 0.8 & 78.2 $\pm$ 1.7 & 96.6 $\pm$ 0.7 & 75.1 $\pm$ 0.5 & 84.2 \\
\midrule
ERM*                  & 86.5 $\pm$ 1.0       & 81.3 $\pm$ 0.6       & 96.2 $\pm$ 0.3       & 82.7 $\pm$ 1.1       & 86.7                 \\
IRM*                  & 84.2 $\pm$ 0.9       & 79.7 $\pm$ 1.5       & 95.9 $\pm$ 0.4       & 78.3 $\pm$ 2.1       & 84.5                 \\
GroupDRO*             & 87.5 $\pm$ 0.5       & 82.9 $\pm$ 0.6       & 97.1 $\pm$ 0.3       & 81.1 $\pm$ 1.2       & 87.1                 \\
Mixup*                & 87.5 $\pm$ 0.4       & 81.6 $\pm$ 0.7       & 97.4 $\pm$ 0.2       & 80.8 $\pm$ 0.9       & 86.8                 \\
MLDG*                 & 87.0 $\pm$ 1.2       & 82.5 $\pm$ 0.9       & 96.7 $\pm$ 0.3       & 81.2 $\pm$ 0.6       & 86.8                 \\
CORAL*                & 86.6 $\pm$ 0.8       & 81.8 $\pm$ 0.9       & 97.1 $\pm$ 0.5       & 82.7 $\pm$ 0.6       & 87.1                 \\
MMD*                  & 88.1 $\pm$ 0.8       & 82.6 $\pm$ 0.7       & 97.1 $\pm$ 0.5       & 81.2 $\pm$ 1.2       & 87.2                 \\
DANN*                 & 87.0 $\pm$ 0.4       & 80.3 $\pm$ 0.6       & 96.8 $\pm$ 0.3       & 76.9 $\pm$ 1.1       & 85.2                 \\
CDANN*                & 87.7 $\pm$ 0.6       & 80.7 $\pm$ 1.2       & 97.3 $\pm$ 0.4       & 77.6 $\pm$ 1.5       & 85.8                 \\
MTL*                  & 87.0 $\pm$ 0.2       & 82.7 $\pm$ 0.8       & 96.5 $\pm$ 0.7       & 80.5 $\pm$ 0.8       & 86.7                 \\
SagNet*               & 87.4 $\pm$ 0.5       & 81.2 $\pm$ 1.2       & 96.3 $\pm$ 0.8       & 80.7 $\pm$ 1.1       & 86.4                 \\
ARM*                  & 85.0 $\pm$ 1.2       & 81.4 $\pm$ 0.2       & 95.9 $\pm$ 0.3       & 80.9 $\pm$ 0.5       & 85.8                 \\
VREx*                 & 87.8 $\pm$ 1.2       & 81.8 $\pm$ 0.7       & 97.4 $\pm$ 0.2       & 82.1 $\pm$ 0.7       & 87.2                 \\
RSC*                  & 86.0 $\pm$ 0.7       & 81.8 $\pm$ 0.9       & 96.8 $\pm$ 0.7       & 80.4 $\pm$ 0.5       & 86.2                 \\
\tdivcams 	          & 86.5 $\pm$ 0.4       & 83.0 $\pm$ 0.5        & 97.2 $\pm$ 0.3       & 82.2 $\pm$ 0.1        & 87.2                 \\
\tdtransformers &  87.8 $\pm$ 0.6 & 81.6 $\pm$ 0.3 & 97.2 $\pm$ 0.5 & 80.9 $\pm$ 0.5 & 86.9 \\
\bottomrule
\end{tabular}
\caption[Domain specific performance for the PACS dataset]{Domain specific performance for the PACS dataset using training-domain validation (top) and  oracle validation denoted with * (bottom). We use a ResNet-50 backbone, optimize with \adam, and follow the distributions specified in \domainbed. Only \rsc and our methods have been added as part of this work, the other baselines are taken from \domainbed.}
\end{center}
\end{table*}

\begin{table*}
\begin{center}
\begin{tabular}{lccccc}
\toprule
\textbf{Algorithm}   & \textbf{A}           & \textbf{C}           & \textbf{P}           & \textbf{R}           & \textbf{Avg.}         \\
\midrule
ERM                  & 61.3 $\pm$ 0.7       & 52.4 $\pm$ 0.3       & 75.8 $\pm$ 0.1       & 76.6 $\pm$ 0.3       & 66.5                 \\
IRM                  & 58.9 $\pm$ 2.3       & 52.2 $\pm$ 1.6       & 72.1 $\pm$ 2.9       & 74.0 $\pm$ 2.5       & 64.3                 \\
GroupDRO             & 60.4 $\pm$ 0.7       & 52.7 $\pm$ 1.0       & 75.0 $\pm$ 0.7       & 76.0 $\pm$ 0.7       & 66.0                 \\
Mixup                & 62.4 $\pm$ 0.8       & 54.8 $\pm$ 0.6       & 76.9 $\pm$ 0.3       & 78.3 $\pm$ 0.2       & 68.1                 \\
MLDG                 & 61.5 $\pm$ 0.9       & 53.2 $\pm$ 0.6       & 75.0 $\pm$ 1.2       & 77.5 $\pm$ 0.4       & 66.8                 \\
CORAL                & 65.3 $\pm$ 0.4       & 54.4 $\pm$ 0.5       & 76.5 $\pm$ 0.1       & 78.4 $\pm$ 0.5       & 68.7                 \\
MMD                  & 60.4 $\pm$ 0.2       & 53.3 $\pm$ 0.3       & 74.3 $\pm$ 0.1       & 77.4 $\pm$ 0.6       & 66.3                 \\
DANN                 & 59.9 $\pm$ 1.3       & 53.0 $\pm$ 0.3       & 73.6 $\pm$ 0.7       & 76.9 $\pm$ 0.5       & 65.9                 \\
CDANN                & 61.5 $\pm$ 1.4       & 50.4 $\pm$ 2.4       & 74.4 $\pm$ 0.9       & 76.6 $\pm$ 0.8       & 65.8                 \\
MTL                  & 61.5 $\pm$ 0.7       & 52.4 $\pm$ 0.6       & 74.9 $\pm$ 0.4       & 76.8 $\pm$ 0.4       & 66.4                 \\
SagNet               & 63.4 $\pm$ 0.2       & 54.8 $\pm$ 0.4       & 75.8 $\pm$ 0.4       & 78.3 $\pm$ 0.3       & 68.1                 \\
ARM                  & 58.9 $\pm$ 0.8       & 51.0 $\pm$ 0.5       & 74.1 $\pm$ 0.1       & 75.2 $\pm$ 0.3       & 64.8                 \\
VREx                 & 60.7 $\pm$ 0.9       & 53.0 $\pm$ 0.9       & 75.3 $\pm$ 0.1       & 76.6 $\pm$ 0.5       & 66.4                 \\
RSC                  & 60.7 $\pm$ 1.4       & 51.4 $\pm$ 0.3       & 74.8 $\pm$ 1.1       & 75.1 $\pm$ 1.3       & 65.5                \\
\divcams 	         & 59.5 $\pm$ 0.3       & 49.7 $\pm$ 0.1        & 75.4 $\pm$ 0.7       & 76.2 $\pm$ 0.2        & 65.2                 \\
\midrule
ERM*                  & 61.7 $\pm$ 0.7       & 53.4 $\pm$ 0.3       & 74.1 $\pm$ 0.4       & 76.2 $\pm$ 0.6       & 66.4                 \\
IRM*                  & 56.4 $\pm$ 3.2       & 51.2 $\pm$ 2.3       & 71.7 $\pm$ 2.7       & 72.7 $\pm$ 2.7       & 63.0                 \\
GroupDRO*             & 60.5 $\pm$ 1.6       & 53.1 $\pm$ 0.3       & 75.5 $\pm$ 0.3       & 75.9 $\pm$ 0.7       & 66.2                 \\
Mixup*                & 63.5 $\pm$ 0.2       & 54.6 $\pm$ 0.4       & 76.0 $\pm$ 0.3       & 78.0 $\pm$ 0.7       & 68.0                 \\
MLDG*                 & 60.5 $\pm$ 0.7       & 54.2 $\pm$ 0.5       & 75.0 $\pm$ 0.2       & 76.7 $\pm$ 0.5       & 66.6                 \\
CORAL*                & 64.8 $\pm$ 0.8       & 54.1 $\pm$ 0.9       & 76.5 $\pm$ 0.4       & 78.2 $\pm$ 0.4       & 68.4                 \\
MMD*                  & 60.4 $\pm$ 1.0       & 53.4 $\pm$ 0.5       & 74.9 $\pm$ 0.1       & 76.1 $\pm$ 0.7       & 66.2                 \\
DANN*                 & 60.6 $\pm$ 1.4       & 51.8 $\pm$ 0.7       & 73.4 $\pm$ 0.5       & 75.5 $\pm$ 0.9       & 65.3                 \\
CDANN*                & 57.9 $\pm$ 0.2       & 52.1 $\pm$ 1.2       & 74.9 $\pm$ 0.7       & 76.2 $\pm$ 0.2       & 65.3                 \\
MTL*                  & 60.7 $\pm$ 0.8       & 53.5 $\pm$ 1.3       & 75.2 $\pm$ 0.6       & 76.6 $\pm$ 0.6       & 66.5                 \\
SagNet*               & 62.7 $\pm$ 0.5       & 53.6 $\pm$ 0.5       & 76.0 $\pm$ 0.3       & 77.8 $\pm$ 0.1       & 67.5                 \\
ARM*                  & 58.8 $\pm$ 0.5       & 51.8 $\pm$ 0.7       & 74.0 $\pm$ 0.1       & 74.4 $\pm$ 0.2       & 64.8                 \\
VREx*                 & 59.6 $\pm$ 1.0       & 53.3 $\pm$ 0.3       & 73.2 $\pm$ 0.5       & 76.6 $\pm$ 0.4       & 65.7                 \\
RSC*                  & 61.7 $\pm$ 0.8       & 53.0 $\pm$ 0.9       & 74.8 $\pm$ 0.8       & 76.3 $\pm$ 0.5       & 66.5                \\
\tdivcams 	          & 58.4 $\pm$ 1.1       & 52.7 $\pm$ 0.7        & 74.3 $\pm$ 0.5       & 75.2 $\pm$ 0.2        & 65.2                 \\
\bottomrule
\end{tabular}
\caption[Domain specific performance for the Office-Home dataset]{Domain specific performance for the Office-Home dataset using training-domain validation (top) and  oracle validation denoted with * (bottom). We use a ResNet-50 backbone, optimize with \adam, and follow the distributions specified in \domainbed. Only \rsc and our methods have been added as part of this work, the other baselines are taken from \domainbed.}
\end{center}
\end{table*}

\begin{table*}
\begin{center}
\begin{tabular}{lccccc}
\toprule
\textbf{Algorithm}   & \textbf{L100}        & \textbf{L38}         & \textbf{L43}         & \textbf{L46}         & \textbf{Avg.}         \\
\midrule
ERM                  & 49.8 $\pm$ 4.4       & 42.1 $\pm$ 1.4       & 56.9 $\pm$ 1.8       & 35.7 $\pm$ 3.9       & 46.1                 \\
IRM                  & 54.6 $\pm$ 1.3       & 39.8 $\pm$ 1.9       & 56.2 $\pm$ 1.8       & 39.6 $\pm$ 0.8       & 47.6                 \\
GroupDRO             & 41.2 $\pm$ 0.7       & 38.6 $\pm$ 2.1       & 56.7 $\pm$ 0.9       & 36.4 $\pm$ 2.1       & 43.2                 \\
Mixup                & 59.6 $\pm$ 2.0       & 42.2 $\pm$ 1.4       & 55.9 $\pm$ 0.8       & 33.9 $\pm$ 1.4       & 47.9                 \\
MLDG                 & 54.2 $\pm$ 3.0       & 44.3 $\pm$ 1.1       & 55.6 $\pm$ 0.3       & 36.9 $\pm$ 2.2       & 47.7                 \\
CORAL                & 51.6 $\pm$ 2.4       & 42.2 $\pm$ 1.0       & 57.0 $\pm$ 1.0       & 39.8 $\pm$ 2.9       & 47.6                 \\
MMD                  & 41.9 $\pm$ 3.0       & 34.8 $\pm$ 1.0       & 57.0 $\pm$ 1.9       & 35.2 $\pm$ 1.8       & 42.2                 \\
DANN                 & 51.1 $\pm$ 3.5       & 40.6 $\pm$ 0.6       & 57.4 $\pm$ 0.5       & 37.7 $\pm$ 1.8       & 46.7                 \\
CDANN                & 47.0 $\pm$ 1.9       & 41.3 $\pm$ 4.8       & 54.9 $\pm$ 1.7       & 39.8 $\pm$ 2.3       & 45.8                 \\
MTL                  & 49.3 $\pm$ 1.2       & 39.6 $\pm$ 6.3       & 55.6 $\pm$ 1.1       & 37.8 $\pm$ 0.8       & 45.6                 \\
SagNet               & 53.0 $\pm$ 2.9       & 43.0 $\pm$ 2.5       & 57.9 $\pm$ 0.6       & 40.4 $\pm$ 1.3       & 48.6                 \\
ARM                  & 49.3 $\pm$ 0.7       & 38.3 $\pm$ 2.4       & 55.8 $\pm$ 0.8       & 38.7 $\pm$ 1.3       & 45.5                 \\
VREx                 & 48.2 $\pm$ 4.3       & 41.7 $\pm$ 1.3       & 56.8 $\pm$ 0.8       & 38.7 $\pm$ 3.1       & 46.4                 \\
RSC                  & 50.2 $\pm$ 2.2       & 39.2 $\pm$ 1.4       & 56.3 $\pm$ 1.4       & 40.8 $\pm$ 0.6       & 46.6                  \\
\divcams 	        & 51.6 $\pm$ 2.2       & 44.4 $\pm$ 2.1        & 55.2 $\pm$ 1.7       & 40.7 $\pm$ 2.6        & 48.0                 \\
\dtransformers & $53.7 \pm 1.0$ & $29.4 \pm 2.5$ & $53.9 \pm 1.0$ & $34.5 \pm 3.1$ & 42.9 \\
\midrule
ERM*                  & 59.4 $\pm$ 0.9       & 49.3 $\pm$ 0.6       & 60.1 $\pm$ 1.1       & 43.2 $\pm$ 0.5       & 53.0                 \\
IRM*                  & 56.5 $\pm$ 2.5       & 49.8 $\pm$ 1.5       & 57.1 $\pm$ 2.2       & 38.6 $\pm$ 1.0       & 50.5                 \\
GroupDRO*             & 60.4 $\pm$ 1.5       & 48.3 $\pm$ 0.4       & 58.6 $\pm$ 0.8       & 42.2 $\pm$ 0.8       & 52.4                 \\
Mixup*                & 67.6 $\pm$ 1.8       & 51.0 $\pm$ 1.3       & 59.0 $\pm$ 0.0       & 40.0 $\pm$ 1.1       & 54.4                 \\
MLDG*                 & 59.2 $\pm$ 0.1       & 49.0 $\pm$ 0.9       & 58.4 $\pm$ 0.9       & 41.4 $\pm$ 1.0       & 52.0                 \\
CORAL*                & 60.4 $\pm$ 0.9       & 47.2 $\pm$ 0.5       & 59.3 $\pm$ 0.4       & 44.4 $\pm$ 0.4       & 52.8                 \\
MMD*                  & 60.6 $\pm$ 1.1       & 45.9 $\pm$ 0.3       & 57.8 $\pm$ 0.5       & 43.8 $\pm$ 1.2       & 52.0                 \\
DANN*                 & 55.2 $\pm$ 1.9       & 47.0 $\pm$ 0.7       & 57.2 $\pm$ 0.9       & 42.9 $\pm$ 0.9       & 50.6                 \\
CDANN*                & 56.3 $\pm$ 2.0       & 47.1 $\pm$ 0.9       & 57.2 $\pm$ 1.1       & 42.4 $\pm$ 0.8       & 50.8                 \\
MTL*                  & 58.4 $\pm$ 2.1       & 48.4 $\pm$ 0.8       & 58.9 $\pm$ 0.6       & 43.0 $\pm$ 1.3       & 52.2                 \\
SagNet*               & 56.4 $\pm$ 1.9       & 50.5 $\pm$ 2.3       & 59.1 $\pm$ 0.5       & 44.1 $\pm$ 0.6       & 52.5                 \\
ARM*                  & 60.1 $\pm$ 1.5       & 48.3 $\pm$ 1.6       & 55.3 $\pm$ 0.6       & 40.9 $\pm$ 1.1       & 51.2                 \\
VREx*                 & 56.8 $\pm$ 1.7       & 46.5 $\pm$ 0.5       & 58.4 $\pm$ 0.3       & 43.8 $\pm$ 0.3       & 51.4                 \\
RSC*                  & 59.9 $\pm$ 1.4       & 46.7 $\pm$ 0.4       & 57.8 $\pm$ 0.5       & 44.3 $\pm$ 0.6       & 52.1                  \\
\tdivcams 	          & 57.7 $\pm$ 1.1       & 46.0 $\pm$ 1.3        & 58.9 $\pm$ 0.4       & 42.5 $\pm$ 0.7        & 51.3                 \\
\tdtransformers & 59.7 $\pm$ 0.6 & 51.1 $\pm$ 1.4 & 56.5 $\pm$ 0.4 & 42.2 $\pm$ 1.0 & 52.4 \\
\bottomrule
\end{tabular}
\caption[Domain specific performance for the Terra Incognita dataset]{Domain specific performance for the Terra Incognita dataset using training-domain validation (top) and  oracle validation denoted with * (bottom). We use a ResNet-50 backbone, optimize with \adam, and follow the distributions specified in \domainbed. Only \rsc and our methods have been added as part of this work, the other baselines are taken from \domainbed.}
\end{center}
\end{table*}

\begin{table*}
\small
\begin{center}
\begin{tabular}{lccccccc}
\toprule
\textbf{Algorithm}   & \textbf{clip}        & \textbf{info}        & \textbf{paint}       & \textbf{quick}       & \textbf{real}        & \textbf{sketch}      & \textbf{Avg.}         \\
\midrule
ERM                  & 58.1 $\pm$ 0.3       & 18.8 $\pm$ 0.3       & 46.7 $\pm$ 0.3       & 12.2 $\pm$ 0.4       & 59.6 $\pm$ 0.1       & 49.8 $\pm$ 0.4       & 40.9                 \\
IRM                  & 48.5 $\pm$ 2.8       & 15.0 $\pm$ 1.5       & 38.3 $\pm$ 4.3       & 10.9 $\pm$ 0.5       & 48.2 $\pm$ 5.2       & 42.3 $\pm$ 3.1       & 33.9                 \\
GroupDRO             & 47.2 $\pm$ 0.5       & 17.5 $\pm$ 0.4       & 33.8 $\pm$ 0.5       & 9.3 $\pm$ 0.3        & 51.6 $\pm$ 0.4       & 40.1 $\pm$ 0.6       & 33.3                 \\
Mixup                & 55.7 $\pm$ 0.3       & 18.5 $\pm$ 0.5       & 44.3 $\pm$ 0.5       & 12.5 $\pm$ 0.4       & 55.8 $\pm$ 0.3       & 48.2 $\pm$ 0.5       & 39.2                 \\
MLDG                 & 59.1 $\pm$ 0.2       & 19.1 $\pm$ 0.3       & 45.8 $\pm$ 0.7       & 13.4 $\pm$ 0.3       & 59.6 $\pm$ 0.2       & 50.2 $\pm$ 0.4       & 41.2                 \\
CORAL                & 59.2 $\pm$ 0.1       & 19.7 $\pm$ 0.2       & 46.6 $\pm$ 0.3       & 13.4 $\pm$ 0.4       & 59.8 $\pm$ 0.2       & 50.1 $\pm$ 0.6       & 41.5                 \\
MMD                  & 32.1 $\pm$ 13.3      & 11.0 $\pm$ 4.6       & 26.8 $\pm$ 11.3      & 8.7 $\pm$ 2.1        & 32.7 $\pm$ 13.8      & 28.9 $\pm$ 11.9      & 23.4                 \\
DANN                 & 53.1 $\pm$ 0.2       & 18.3 $\pm$ 0.1       & 44.2 $\pm$ 0.7       & 11.8 $\pm$ 0.1       & 55.5 $\pm$ 0.4       & 46.8 $\pm$ 0.6       & 38.3                 \\
CDANN                & 54.6 $\pm$ 0.4       & 17.3 $\pm$ 0.1       & 43.7 $\pm$ 0.9       & 12.1 $\pm$ 0.7       & 56.2 $\pm$ 0.4       & 45.9 $\pm$ 0.5       & 38.3                 \\
MTL                  & 57.9 $\pm$ 0.5       & 18.5 $\pm$ 0.4       & 46.0 $\pm$ 0.1       & 12.5 $\pm$ 0.1       & 59.5 $\pm$ 0.3       & 49.2 $\pm$ 0.1       & 40.6                 \\
SagNet               & 57.7 $\pm$ 0.3       & 19.0 $\pm$ 0.2       & 45.3 $\pm$ 0.3       & 12.7 $\pm$ 0.5       & 58.1 $\pm$ 0.5       & 48.8 $\pm$ 0.2       & 40.3                 \\
ARM                  & 49.7 $\pm$ 0.3       & 16.3 $\pm$ 0.5       & 40.9 $\pm$ 1.1       & 9.4 $\pm$ 0.1        & 53.4 $\pm$ 0.4       & 43.5 $\pm$ 0.4       & 35.5                 \\
VREx                 & 47.3 $\pm$ 3.5       & 16.0 $\pm$ 1.5       & 35.8 $\pm$ 4.6       & 10.9 $\pm$ 0.3       & 49.6 $\pm$ 4.9       & 42.0 $\pm$ 3.0       & 33.6                 \\
RSC                   & 55.0 $\pm$ 1.2       & 18.3 $\pm$ 0.5       & 44.4 $\pm$ 0.6       & 12.2 $\pm$ 0.2       & 55.7 $\pm$ 0.7       & 47.8 $\pm$ 0.9       & 38.9                 \\
\divcams 	        & 57.7 $\pm$ 0.3       & 19.3 $\pm$ 0.3        & 46.8 $\pm$ 0.2       & 12.7 $\pm$ 0.4        & 58.9 $\pm$ 0.2	    & 48.5 $\pm$ 0.4     & 40.7                 \\
\midrule
ERM*                  & 58.6 $\pm$ 0.3       & 19.2 $\pm$ 0.2       & 47.0 $\pm$ 0.3       & 13.2 $\pm$ 0.2       & 59.9 $\pm$ 0.3       & 49.8 $\pm$ 0.4       & 41.3                 \\
IRM*                  & 40.4 $\pm$ 6.6       & 12.1 $\pm$ 2.7       & 31.4 $\pm$ 5.7       & 9.8 $\pm$ 1.2        & 37.7 $\pm$ 9.0       & 36.7 $\pm$ 5.3       & 28.0                 \\
GroupDRO*             & 47.2 $\pm$ 0.5       & 17.5 $\pm$ 0.4       & 34.2 $\pm$ 0.3       & 9.2 $\pm$ 0.4        & 51.9 $\pm$ 0.5       & 40.1 $\pm$ 0.6       & 33.4                 \\
Mixup*                & 55.6 $\pm$ 0.1       & 18.7 $\pm$ 0.4       & 45.1 $\pm$ 0.5       & 12.8 $\pm$ 0.3       & 57.6 $\pm$ 0.5       & 48.2 $\pm$ 0.4       & 39.6                 \\
MLDG*                 & 59.3 $\pm$ 0.1       & 19.6 $\pm$ 0.2       & 46.8 $\pm$ 0.2       & 13.4 $\pm$ 0.2       & 60.1 $\pm$ 0.4       & 50.4 $\pm$ 0.3       & 41.6                 \\
CORAL*                & 59.2 $\pm$ 0.1       & 19.9 $\pm$ 0.2       & 47.4 $\pm$ 0.2       & 14.0 $\pm$ 0.4       & 59.8 $\pm$ 0.2       & 50.4 $\pm$ 0.4       & 41.8                 \\
MMD*                  & 32.2 $\pm$ 13.3      & 11.2 $\pm$ 4.5       & 26.8 $\pm$ 11.3      & 8.8 $\pm$ 2.2        & 32.7 $\pm$ 13.8      & 29.0 $\pm$ 11.8      & 23.5                 \\
DANN*                 & 53.1 $\pm$ 0.2       & 18.3 $\pm$ 0.1       & 44.2 $\pm$ 0.7       & 11.9 $\pm$ 0.1       & 55.5 $\pm$ 0.4       & 46.8 $\pm$ 0.6       & 38.3                 \\
CDANN*                & 54.6 $\pm$ 0.4       & 17.3 $\pm$ 0.1       & 44.2 $\pm$ 0.7       & 12.8 $\pm$ 0.2       & 56.2 $\pm$ 0.4       & 45.9 $\pm$ 0.5       & 38.5                 \\
MTL*                  & 58.0 $\pm$ 0.4       & 19.2 $\pm$ 0.2       & 46.2 $\pm$ 0.1       & 12.7 $\pm$ 0.2       & 59.9 $\pm$ 0.1       & 49.0 $\pm$ 0.0       & 40.8                 \\
SagNet*               & 57.7 $\pm$ 0.3       & 19.1 $\pm$ 0.1       & 46.3 $\pm$ 0.5       & 13.5 $\pm$ 0.4       & 58.9 $\pm$ 0.4       & 49.5 $\pm$ 0.2       & 40.8                 \\
ARM*                  & 49.6 $\pm$ 0.4       & 16.5 $\pm$ 0.3       & 41.5 $\pm$ 0.8       & 10.8 $\pm$ 0.1       & 53.5 $\pm$ 0.3       & 43.9 $\pm$ 0.4       & 36.0                 \\
VREx*                 & 43.3 $\pm$ 4.5       & 14.1 $\pm$ 1.8       & 32.5 $\pm$ 5.0       & 9.8 $\pm$ 1.1        & 43.5 $\pm$ 5.6       & 37.7 $\pm$ 4.5       & 30.1                 \\
RSC*                  & 55.0 $\pm$ 1.2       & 18.3 $\pm$ 0.5       & 44.4 $\pm$ 0.6       & 12.5 $\pm$ 0.1       & 55.7 $\pm$ 0.7       & 47.8 $\pm$ 0.9       & 38.9                \\
\tdivcams 	           & 57.8 $\pm$ 0.2       & 19.3 $\pm$ 0.3        & 47.0 $\pm$ 0.1       & 13.1 $\pm$ 0.3      & 59.6 $\pm$ 0.1	     & 48.9 $\pm$ 0.2       & 41.0                 \\
\bottomrule
\end{tabular}
\caption[Domain specific performance for the DomainNet dataset]{Domain specific performance for the DomainNet dataset using training-domain validation (top) and  oracle validation denoted with * (bottom). We use a ResNet-50 backbone, optimize with \adam, and follow the distributions specified in \domainbed. Only \rsc and our methods have been added as part of this work, the other baselines are taken from \domainbed.}
\end{center}
\end{table*}
\chapter{Additional distance plots}
\label{sec:additional_distances}

\begin{figure}[ht]
    \centering
    \includegraphics[width=\textwidth]{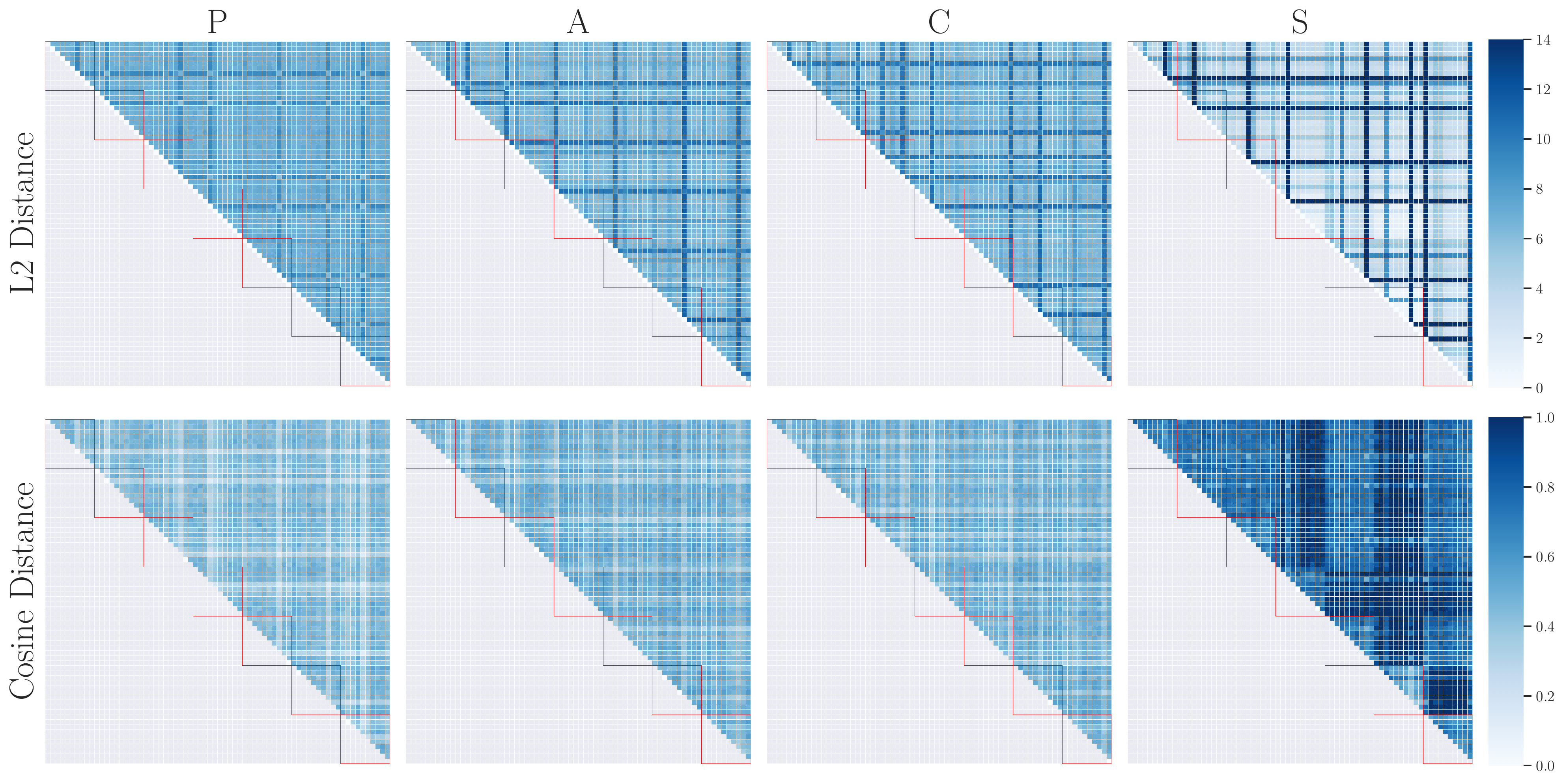}
    \caption[First data split pairwise prototype distances with $w_{c,j} = -1.0$] {Pairwise learned prototype $\ell_2$-distance (top) and cosine-distance $\cdistance$ (bottom) of the best-performing model with negative weight $w_{c,j} = -1.0\; \forall j: \prot \notin \prots_c$ for each testing domain. Red squares denote prototype class correspondence for the $7$ different classes in the PACS dataset. No self-challenging is applied and colormap bounds are adjusted per metric for visualization purposes. First data split.}
    \label{fig:pairwise_distance}
\end{figure}

\begin{figure}[ht]
    \centering
    \includegraphics[width=\textwidth]{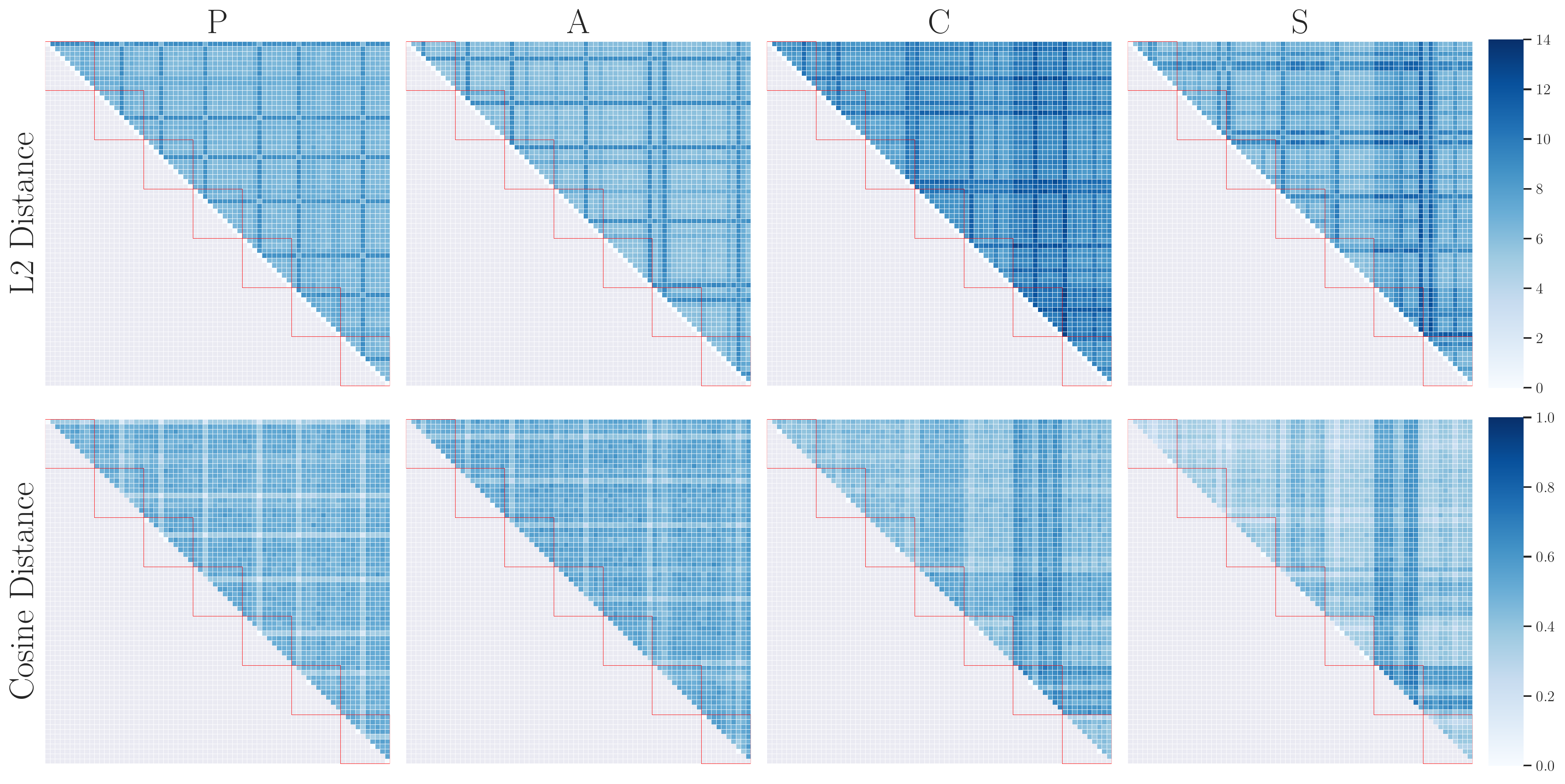}
    \caption[Third data split pairwise prototype distances with $w_{c,j} = -1.0$] {Pairwise learned prototype $\ell_2$-distance (top) and cosine-distance $\cdistance$ (bottom) of the best-performing model with negative weight $w_{c,j} = -1.0\; \forall j: \prot \notin \prots_c$ for each testing domain. Red squares denote prototype class correspondence for the $7$ different classes in the PACS dataset. No self-challenging is applied and colormap bounds are adjusted per metric for visualization purposes. Third data split.}
    \label{fig:pw_distance_trial2}
\end{figure}

\begin{figure}[ht]
    \centering
    \includegraphics[width=\textwidth]{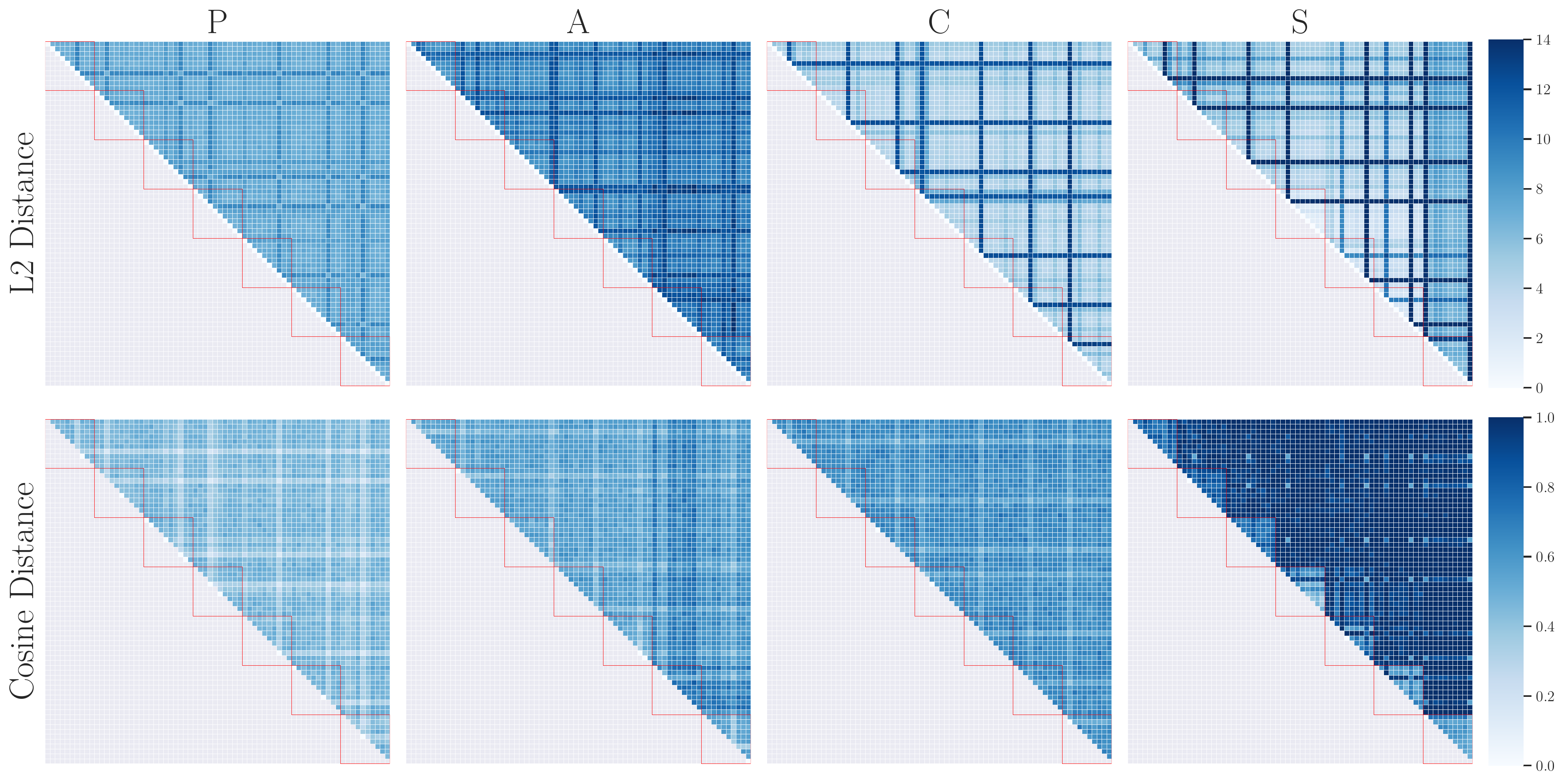}
    \caption[First data split pairwise self-challenging prototype distances with $w_{c,j} = -1.0$] {Pairwise learned prototype $\ell_2$-distance (top) and cosine-distance $\cdistance$ (bottom) of the best-performing model with negative weight $w_{c,j} = -1.0\; \forall j: \prot \notin \prots_c$ for each testing domain. Red squares denote prototype class correspondence for the $7$ different classes in the PACS dataset. Self-challenging is applied and colormap bounds are adjusted per metric for visualization purposes. First data split.}
    \label{fig:pairwise_distance_sc}
\end{figure}

\begin{figure}[ht]
    \centering
    \includegraphics[width=\textwidth]{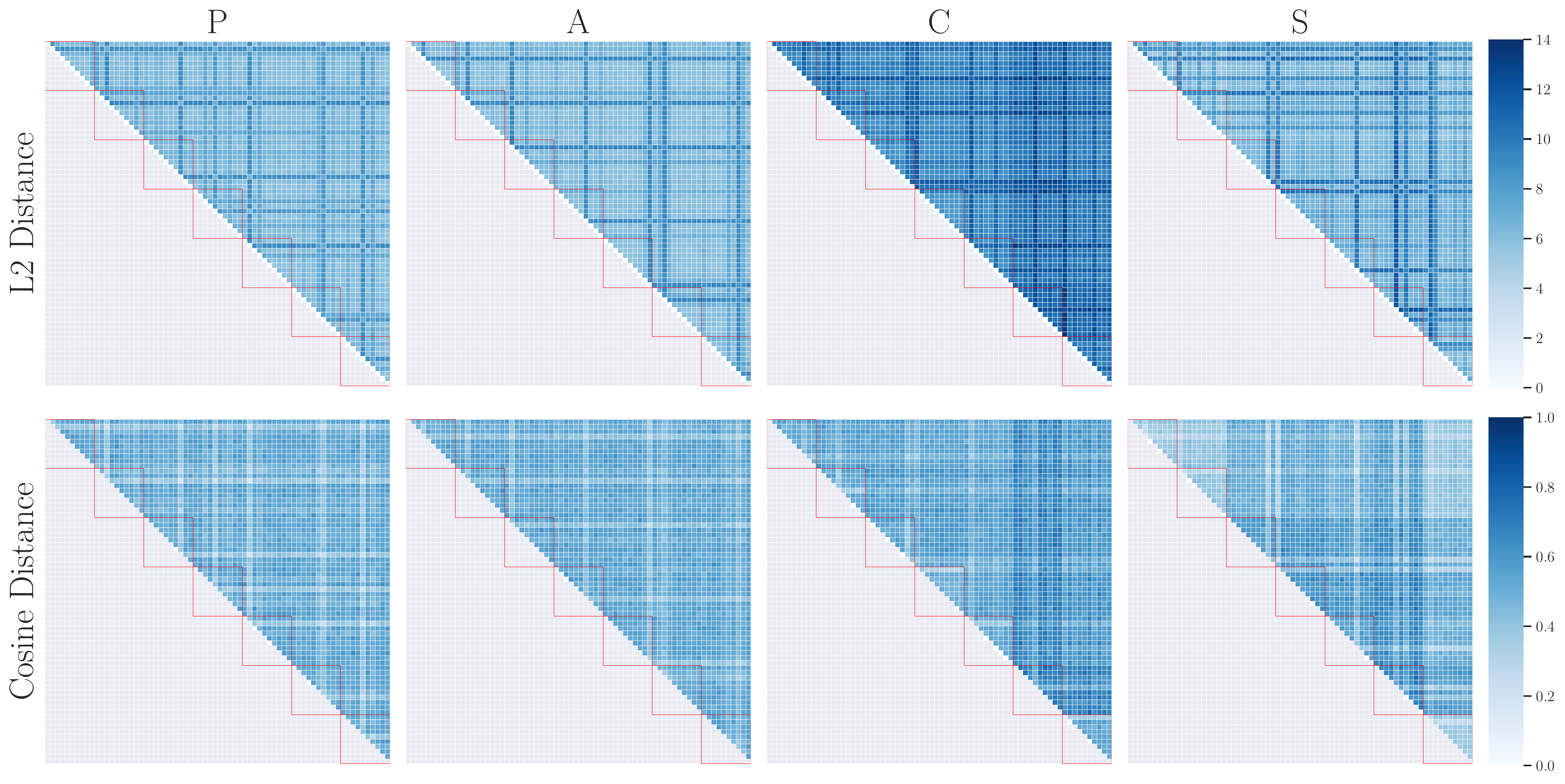}
    \caption[Third data split pairwise self-challenging prototype distances with $w_{c,j} = -1.0$] {Pairwise learned prototype $\ell_2$-distance (top) and cosine-distance $\cdistance$ (bottom) of the best-performing model with negative weight $w_{c,j} = -1.0\; \forall j: \prot \notin \prots_c$ for each testing domain. Red squares denote prototype class correspondence for the $7$ different classes in the PACS dataset. Self-challenging is applied and colormap bounds are adjusted per metric for visualization purposes. Third data split.}
\end{figure}


\begin{figure}[ht]
    \centering
    \includegraphics[width=\textwidth]{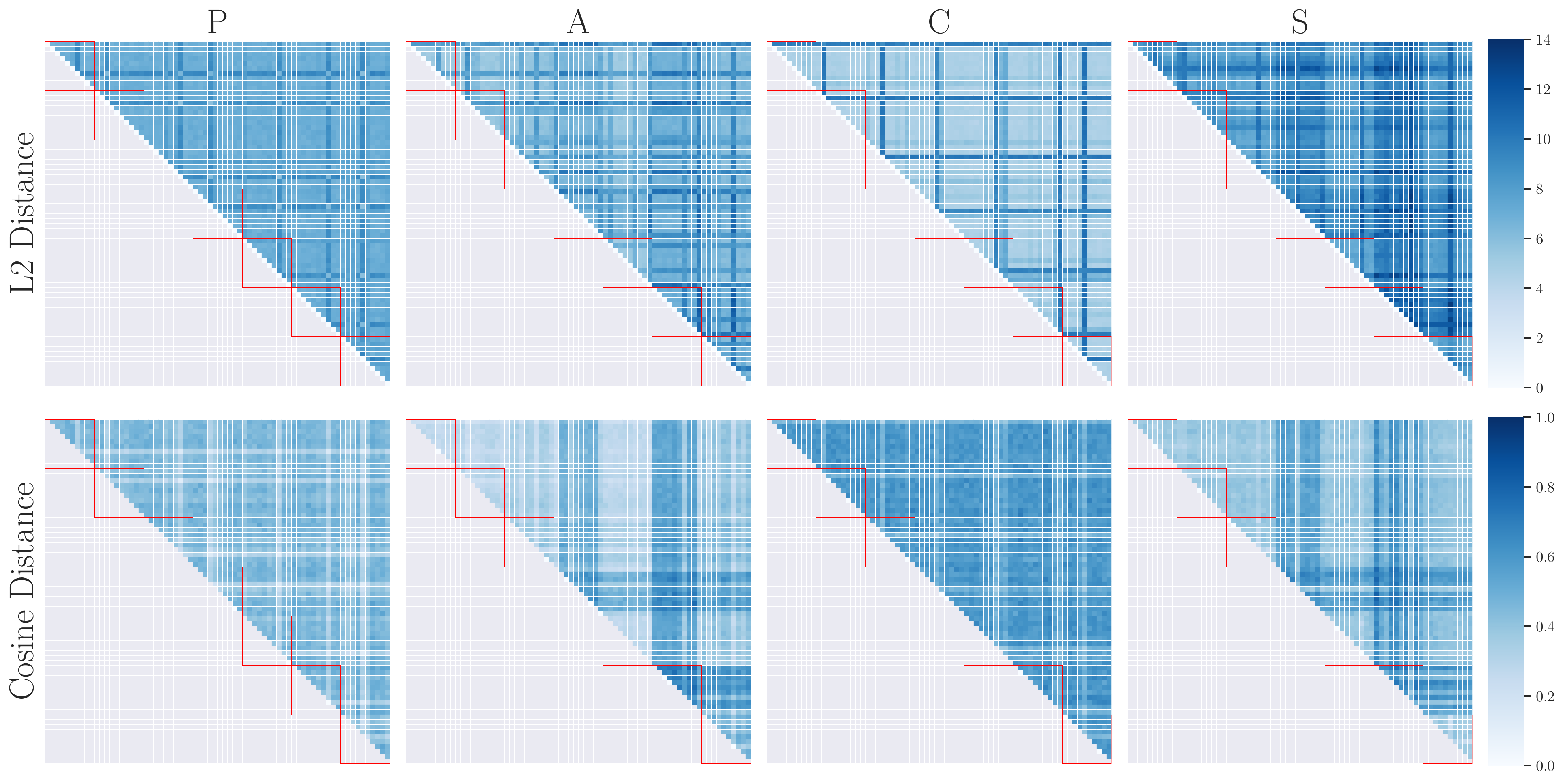}
    \caption[First data split pairwise prototype distances with $w_{c,j} = 0.0$] {Pairwise learned prototype $\ell_2$-distance (top) and cosine-distance $\cdistance$ (bottom) of the best-performing model with negative weight $w_{c,j} = 0.0\; \forall j: \prot \notin \prots_c$ for each testing domain. Red squares denote prototype class correspondence for the $7$ different classes in the PACS dataset. No self-challenging is applied and colormap bounds are adjusted per metric for visualization purposes. First data split.}
    \label{fig:pw_distance_0.0_trial0}
\end{figure}

\begin{figure}[ht]
    \centering
    \includegraphics[width=\textwidth]{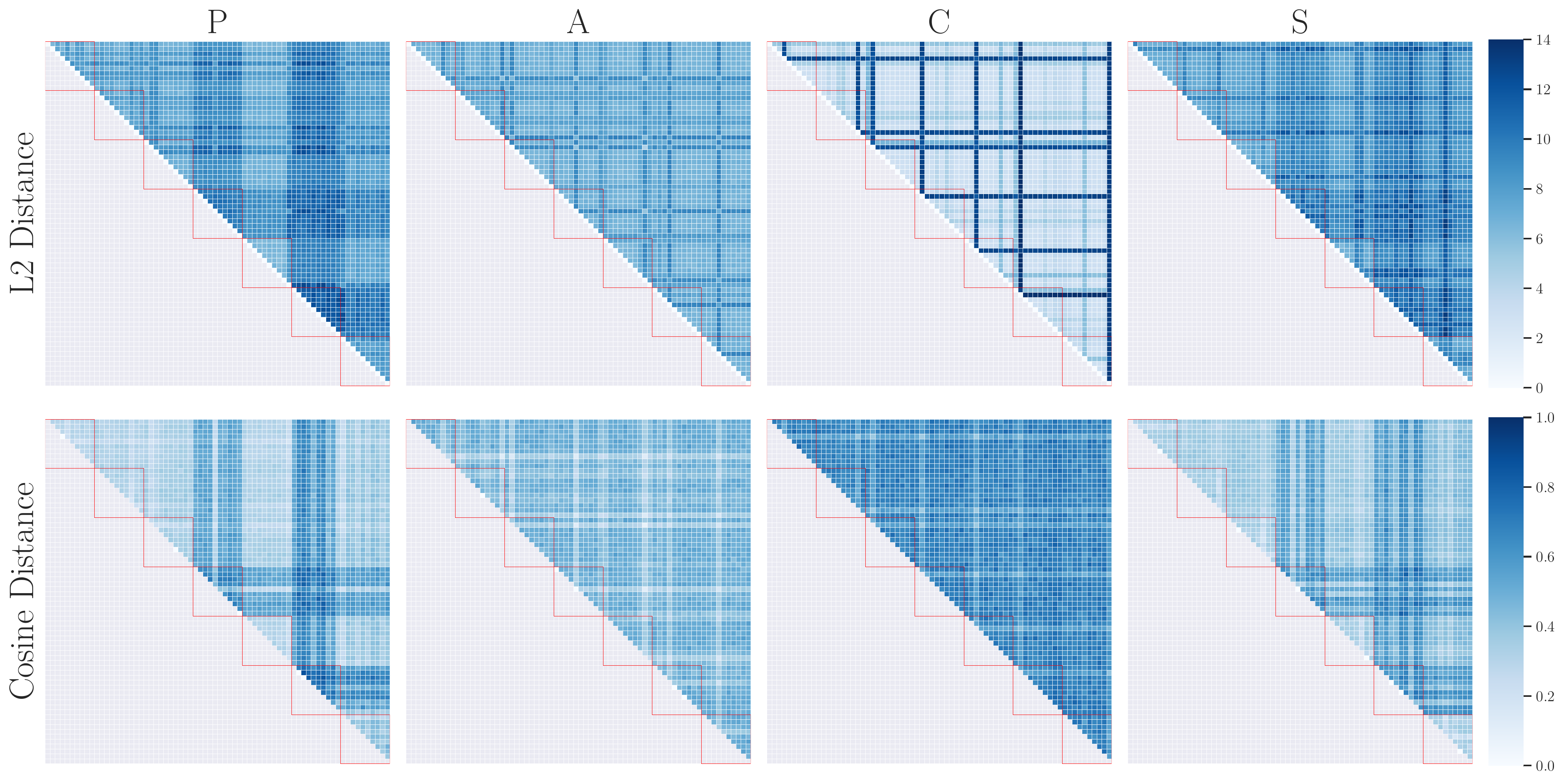}
    \caption[Second data split pairwise prototype distances with $w_{c,j} = 0.0$] {Pairwise learned prototype $\ell_2$-distance (top) and cosine-distance $\cdistance$ (bottom) of the best-performing model with negative weight $w_{c,j} = 0.0\; \forall j: \prot \notin \prots_c$ for each testing domain. Red squares denote prototype class correspondence for the $7$ different classes in the PACS dataset. No self-challenging is applied and colormap bounds are adjusted per metric for visualization purposes. Second data split.}
    \label{fig:pw_distance_0.0_trial1}
\end{figure}

\begin{figure}[ht]
    \centering
    \includegraphics[width=\textwidth]{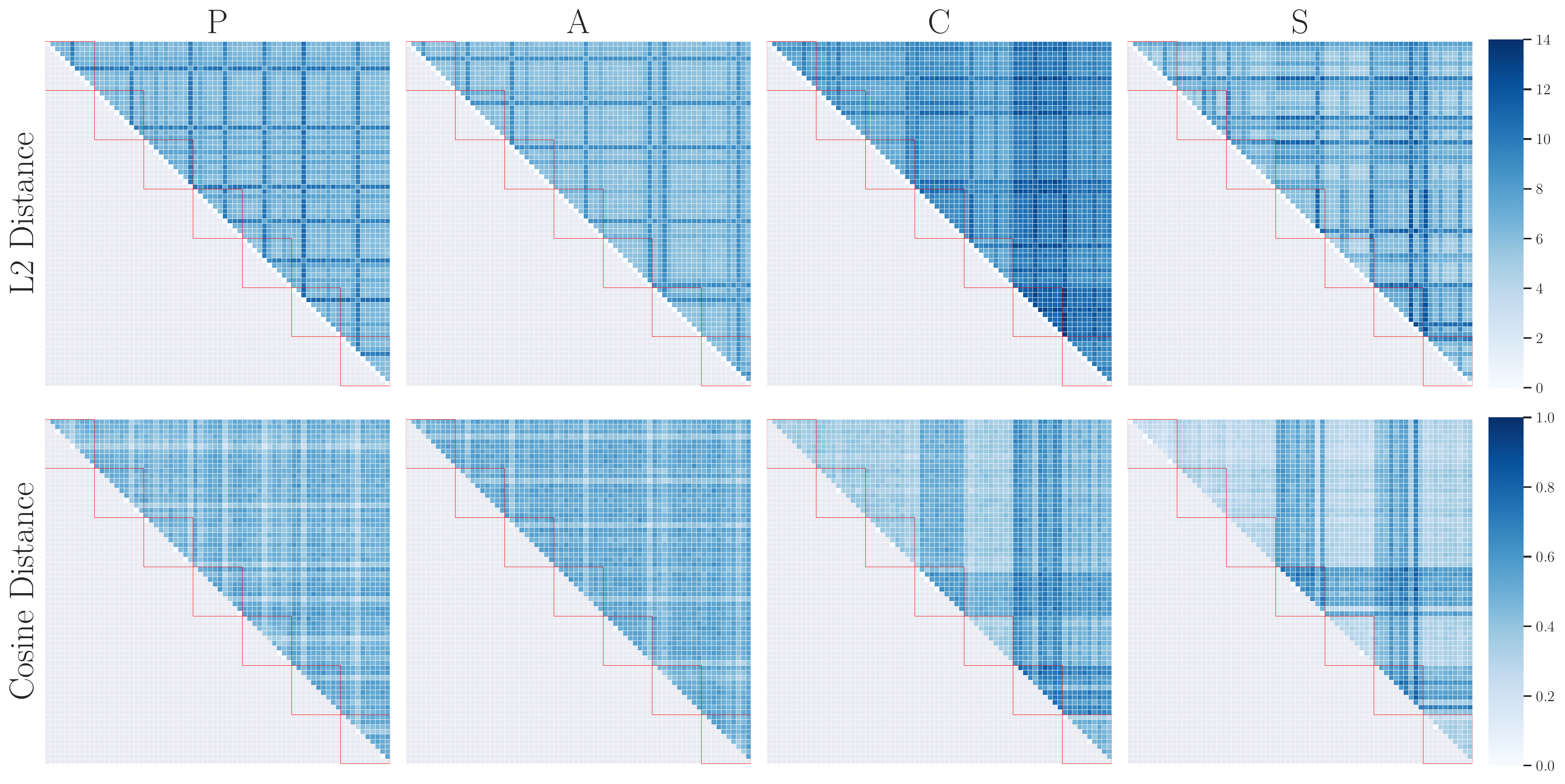}
    \caption[Third data split pairwise prototype distances with $w_{c,j} = 0.0$] {Pairwise learned prototype $\ell_2$-distance (top) and cosine-distance $\cdistance$ (bottom) of the best-performing model with negative weight $w_{c,j} = 0.0\; \forall j: \prot \notin \prots_c$ for each testing domain. Red squares denote prototype class correspondence for the $7$ different classes in the PACS dataset. No self-challenging is applied and colormap bounds are adjusted per metric for visualization purposes. Third data split.}
    \label{fig:pw_distance_0.0_trial2}
\end{figure}

\begin{figure}[ht]
    \centering
    \includegraphics[width=\textwidth]{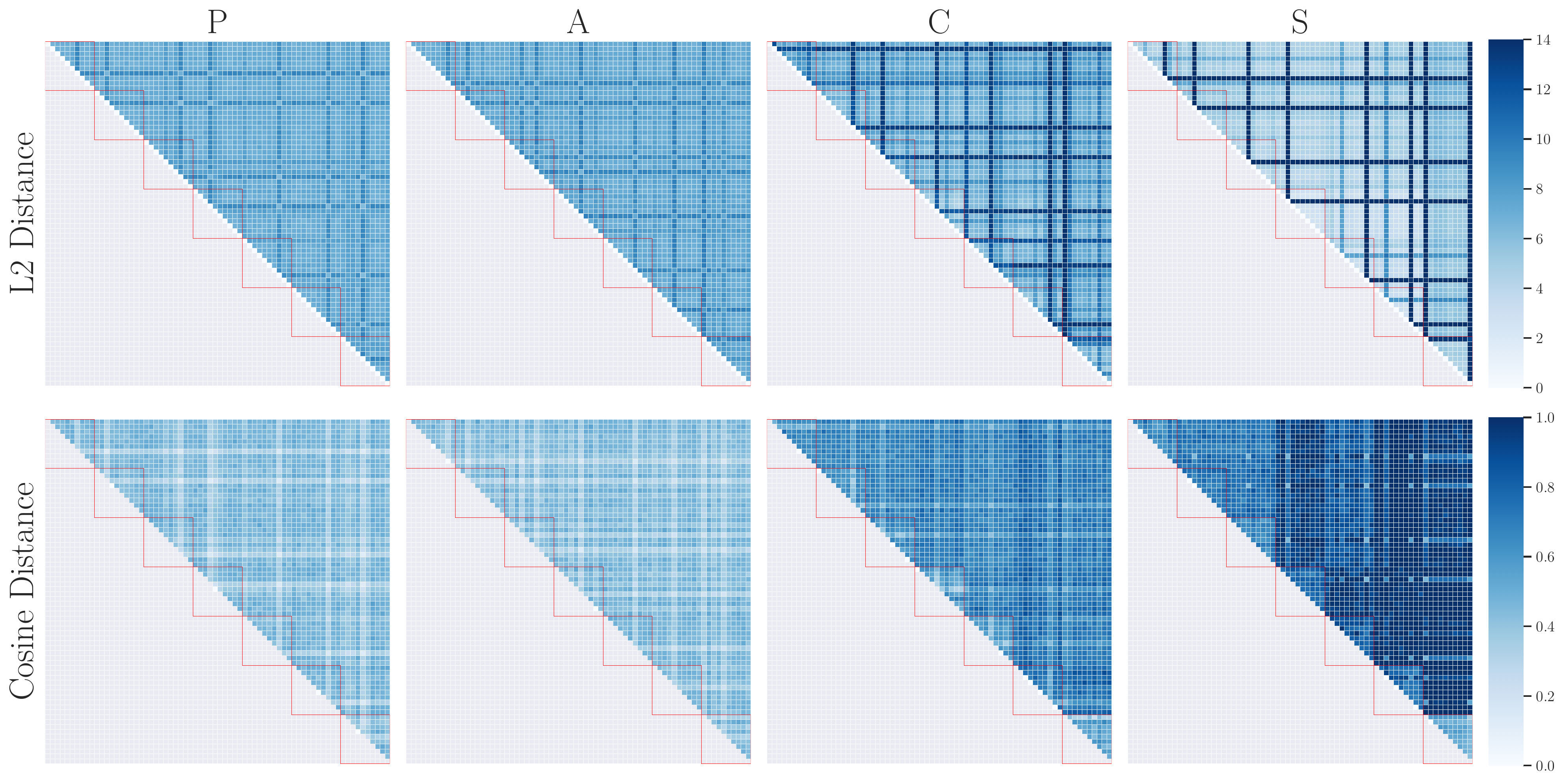}
    \caption[First data split pairwise self-challenging prototype distances with $w_{c,j} = 0.0$] {Pairwise learned prototype $\ell_2$-distance (top) and cosine-distance $\cdistance$ (bottom) of the best-performing model with negative weight $w_{c,j} = 0.0\; \forall j: \prot \notin \prots_c$ for each testing domain. Red squares denote prototype class correspondence for the $7$ different classes in the PACS dataset. Self-challenging is applied and colormap bounds are adjusted per metric for visualization purposes. First data split.}
    \label{fig:pw_distance_0.0_trial0-sc}
\end{figure}

\begin{figure}[ht]
    \centering
    \includegraphics[width=\textwidth]{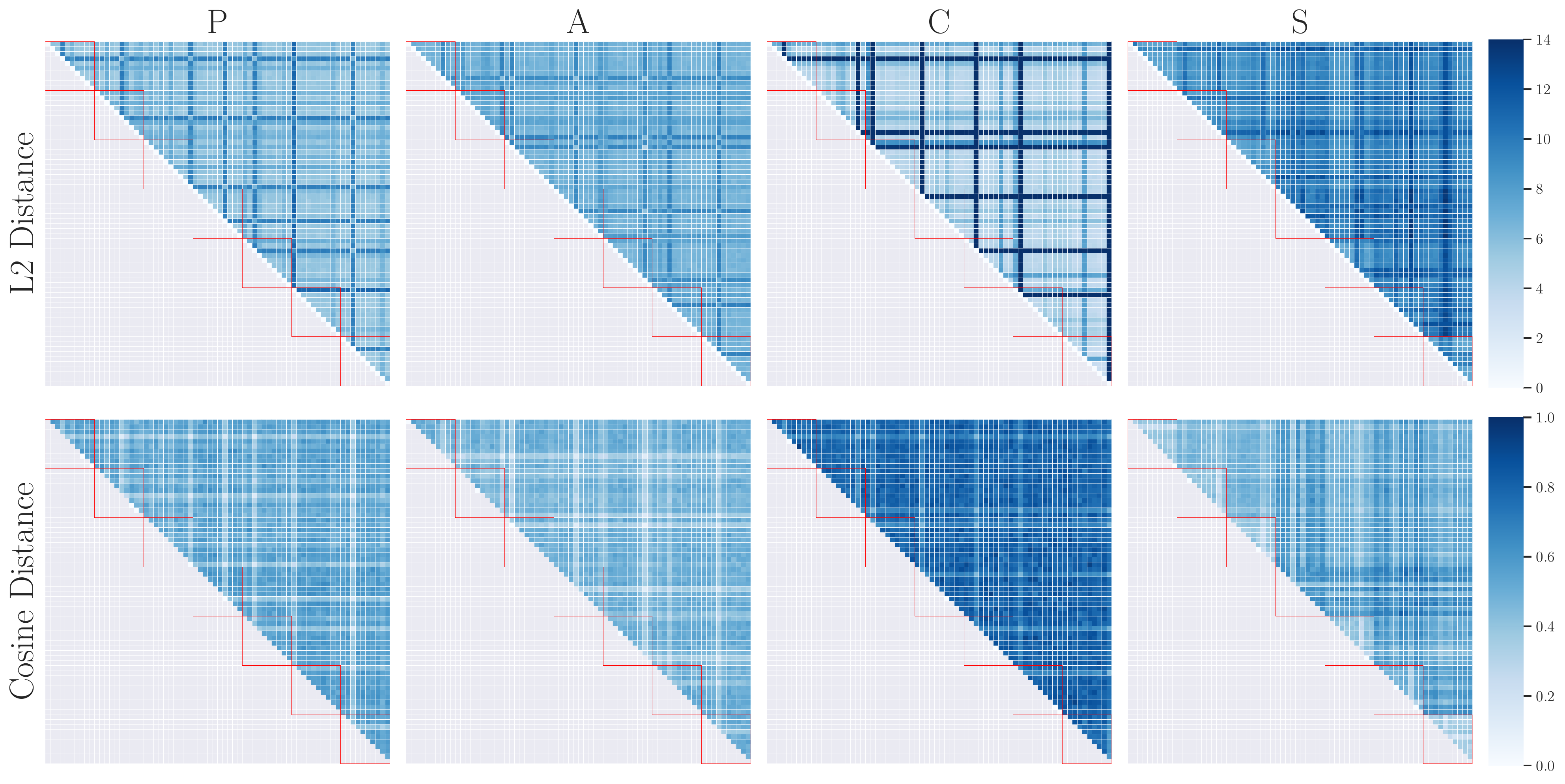}
    \caption[Second data split pairwise self-challenging prototype distances with $w_{c,j} = 0.0$] {Pairwise learned prototype $\ell_2$-distance (top) and cosine-distance $\cdistance$ (bottom) of the best-performing model with negative weight $w_{c,j} = 0.0\; \forall j: \prot \notin \prots_c$ for each testing domain. Red squares denote prototype class correspondence for the $7$ different classes in the PACS dataset. Self-challenging is applied and colormap bounds are adjusted per metric for visualization purposes. Second data split.}
    \label{fig:pw_distance_0.0_trial1-sc}
\end{figure}

\begin{figure}[ht]
    \centering
    \includegraphics[width=\textwidth]{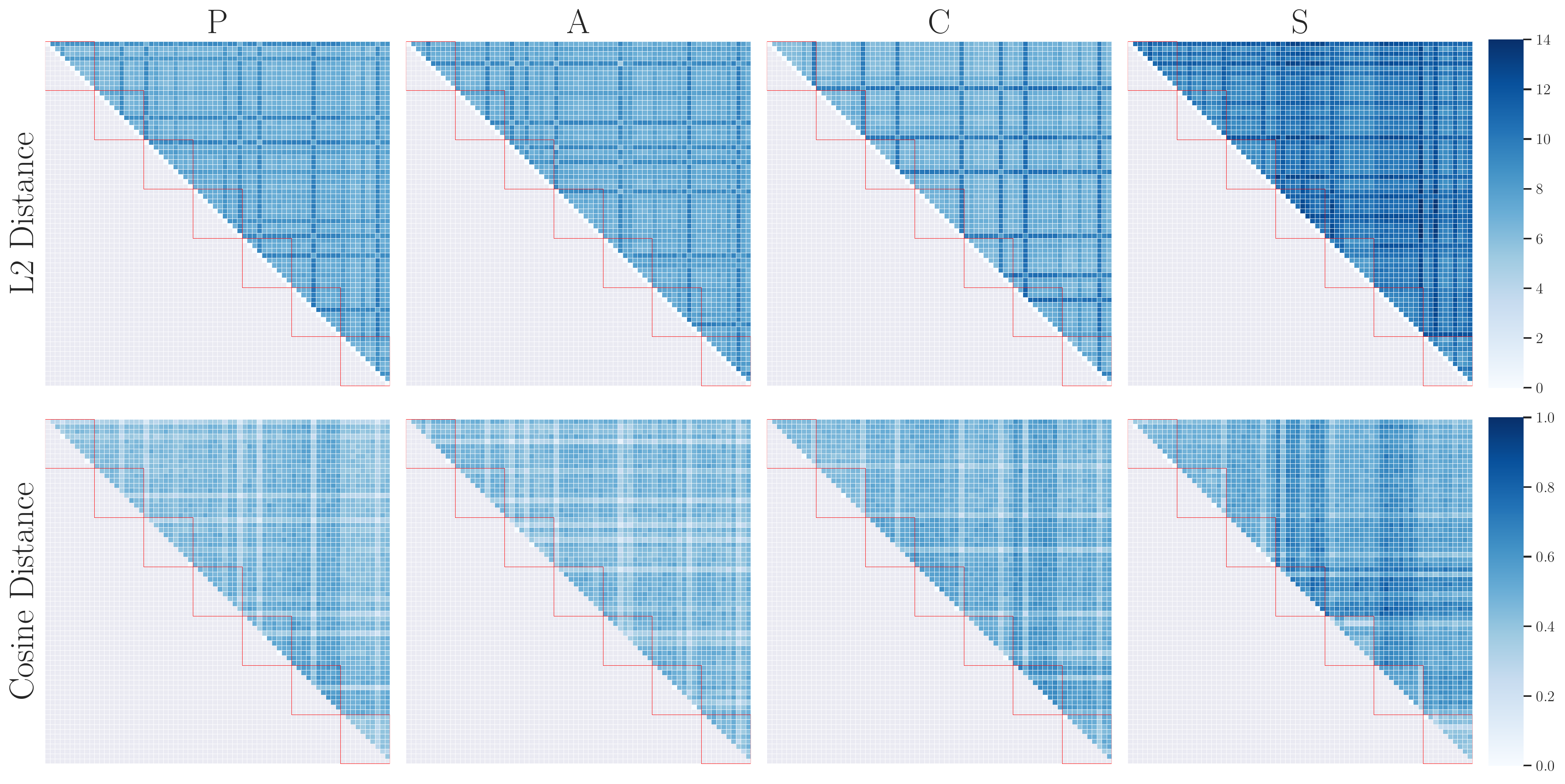}
    \caption[Third data split pairwise self-challenging prototype distances with $w_{c,j} = 0.0$] {Pairwise learned prototype $\ell_2$-distance (top) and cosine-distance $\cdistance$ (bottom) of the best-performing model with negative weight $w_{c,j} = 0.0\; \forall j: \prot \notin \prots_c$ for each testing domain. Red squares denote prototype class correspondence for the $7$ different classes in the PACS dataset. Self-challenging is applied and colormap bounds are adjusted per metric for visualization purposes. Third data split.}
    \label{fig:pw_distance_0.0_trial2-sc}
\end{figure}

\include{Appendices/AppendixC}


\end{document}